# Ecological Cycle Optimizer: A novel nature-inspired metaheuristic algorithm for non-convex global optimization

Boyu Ma[a,b,*,†], Jiaxiao Shi[a,b,†], Yiming Ji[a] and Zhengpu Wang[a]

[a]*State Key Laboratory of Robotics and Systems, Harbin Institute of Technology, Harbin, 150001, China*
[b]*School of Mechanical and Aerospace Engineering, Nanyang Technological University, Singapore, 639798, Singapore*

ABSTRACT

This article proposes the Ecological Cycle Optimizer (ECO), a novel metaheuristic algorithm inspired by energy flow and material cycling within ecosystems. ECO draws an analogy between the dynamic process of solving optimization problems and ecological cycling. Unique update strategies are designed for the producer, consumer and decomposer, aiming to enhance the balance between *exploration* and *exploitation* processes. Through these strategies, ECO is able to approach the global optimum, simulating the evolution of an ecological system toward its optimal state of stability and balance. Moreover, a parameter sensitivity analysis is conducted on 23 classic optimization functions to determine a suitable default configuration for ECO. Furthermore, 30 competitive metaheuristic algorithms are selected to form an algorithm pool, and comprehensive experiments are conducted on the IEEE CEC-2014 and CEC-2017 test suites. Among these, five top-performing algorithms, namely ARO, CFOA, CSA, WSO, and INFO, are chosen for an in-depth comparison with the ECO on the IEEE CEC-2020 test suite, verifying the ECO's exceptional optimization performance. Finally, in order to validate the practical applicability of ECO in complex real-world engineering problems, five state-of-the-art algorithms, including FDB-AGDE, FDB-SFS, LRFDB-COA, L-SHADE, and NSM-SFS are selected for comparative experiments on five engineering problems from the CEC-2020-RW test suite, demonstrating that ECO achieves competitive performance against advanced engineering-oriented algorithms. The ECO project page is available at https://jxxsteven7.github.io/ECO-Optimizer/.

## 1. Introduction

Complex constrained optimization problems in the real world are often characterized by high dimensionality, nonlinearity, and non-convexity. Optimization, as a mathematical tool, plays a crucial role across various disciplines, aiming to identify global optima within constrained spaces. Traditional gradient-based methods, however, frequently encounter local optima when addressing such problems, limiting their global optimization capabilities [1, 2]. In contrast, metaheuristic (MH) algorithms offer advantages such as simplicity in mathematical modeling, ease of implementation, black-box adaptability, and a high degree of randomness, making them effective at escaping local optima to find global solutions. As a result, MH algorithms have been widely applied to large-scale complex constrained optimization problems in diverse fields.

In the past two decades, MH algorithms have attracted significant attention within the academic community, driven by rapid advancements in the field. The inspiration for MH algorithms often derives from natural phenomena, animal behaviors, physical laws, and human activities. Researchers model these patterns mathematically to develop intelligent optimizers for solving complex problems. Previous MH algorithms are primarily inspired by *evolution*, *nature*, *physics*, *human behavior*, and *mathematics*, along with their related improved algorithms.

The implementation process of MH algorithms includes population initialization and the iterative search process, the latter of which consists of three steps: guide selection, convergence search, and population update. The convergence search step of numerous MH algorithms employ different strategies, resulting in diverse performance outcomes. Fundamentally, these convergence search models decompose the optimization process into two key phases: *exploration* and *exploitation*, which are essential to the search-based problem-solving paradigm [3]. *Exploration* allows the algorithm to broadly investigate the search space, identifying regions where a global optimum may exist, whereas *exploitation* focuses on intensively refining the search within those promising regions to locate the global optimum.

[†]These authors contributed equally to this work.
[*]Corresponding author: boyu.ma@ntu.edu.sg (B. Ma)

Although the expected performance of MH algorithms may be comparable across constrained optimization problems, the balance between *exploration* and *exploitation* varies among different algorithms. Excessive *exploration* can result in a slower convergence rate, while overemphasis on *exploitation* risks the algorithm becoming trapped in local optima. These differences in balancing the two phases contribute to the varying performance of algorithms across different problems, which aligns with the No Free Lunch (NFL) theorem [4]. This theorem asserts that no single algorithm is universally optimal for all constrained optimization problems. Consequently, researchers have proposed numerous MH algorithms inspired by diverse concepts, making this an active area of research. In this article, we propose a novel nature-inspired MH optimizer, aimed at demonstrating superior performance compared to previous state-of-the-art (SOTA) MH algorithms across a broader range of complex constrained optimization problems.

Drawing inspiration from energy flow and material cycling within ecosystems, we propose **Ecological Cycle Optimizer** (**ECO**). The dynamic optimization process of ECO is analogous to an ecological cycle, where the search for the global optimum resembles an ecosystem achieving balance and stability. We design specific iterative update strategies for different ecosystem roles, namely producers, consumers, and decomposers. For consumers, distinct predation strategies are developed for herbivores, carnivores, and omnivores, while decomposer strategies incorporate optimal, local random, and global random decomposition approaches. By enhancing the balance between *exploration* and *exploitation*, ECO addresses the limitations of previous MH algorithms, such as slow convergence, low precision, and susceptibility to local optima, thereby achieving superior performance across a broader range of complex non-convex optimization problems. Furthermore, the 23 classic optimization functions, IEEE CEC-2014, CEC-2017, CEC-2020 test suites, along with five classic real-world mechanical design engineering problems from CEC-2020-RW, are used to validate the superior performance of ECO compared to various SOTA MH algorithms.

The remainder of this article is organized as follows. The literature review is outlined in Section 2. Section 3 provides a detailed description of the inspiration and mathematical model underlying the proposed ECO. Section 4 introduces the experimental design, parameter settings, and evaluation criteria. In Section 5, a parameter sensitivity analysis is conducted to determine a suitable default configuration for ECO. Section 6 evaluates the optimization performance potential of ECO on benchmark functions and constructs a comparison pool of 30 SOTA MH algorithms to highlight its comparative advantages. Section 7 further validates the engineering applicability of ECO on real-world engineering optimization problems. Finally, Section 8 concludes the article with some final remarks.

The main contributions of this article are summarized as follows:

(1) Inspired by energy flow and material cycling within ecosystems, a novel nature-inspired metaheuristic algorithm, ECO, is proposed for complex non-convex global optimization;

(2) ECO designs distinct update strategies for producers, consumers with herbivore, carnivore, and omnivore predation behaviors, and decomposers with optimal, local random, and global random decomposition mechanisms, which collaboratively improve the *exploration–exploitation* balance during the search process;

(3) By constructing a comparison pool of 30 SOTA MH algorithms for benchmarking on the CEC-2014 and CEC-2017 suites, and further comparing ECO with the five top-performing algorithms on the CEC-2020 suite, ECO demonstrates competitive performance and robustness;

(4) ECO is further validated on five real-world engineering problems from the CEC-2020-RW suite, achieving superior performance against five highly competitive SOTA algorithms and thereby demonstrating its strong potential for tackling complex engineering optimization scenarios.

## 2. Literature review

This section reviews representative recent MH algorithms from different inspiration categories. Existing MH algorithms can be broadly categorized into five categories inspired by *evolution*, *nature*, *physics*, *human behavior*, and *mathematics*. Despite their different inspirations, these algorithms generally aim to achieve an effective balance between *exploration* and *exploitation* in complex non-convex search spaces.

**Evolution-inspired MH algorithms** draw on Darwinian principles of natural selection and evolve populations of candidate solutions through selection, crossover, and mutation. Genetic Algorithm (GA) [5] established a foundational paradigm for evolutionary optimization through chromosome encoding, fitness-based selection, crossover, and mutation, while Differential Evolution (DE) [6] solves continuous optimization problems using difference-vector-based mutation and simple vector operations.

**Nature-inspired MH algorithms** mimic the collective behaviors of biological populations and the interactions within natural ecosystems, forming one of the most active branches of MH research. Particle Swarm Optimization (PSO) [7] guides the search by modeling the information exchange between individual and collective experience in the foraging behavior of bird flocks, while Grey Wolf Optimizer (GWO) [8] incorporates the social hierarchy and cooperative hunting mechanism of grey wolves to guide population search and balance *exploration* and *exploitation*. In recent years, Artificial Rabbits Optimization (ARO) [9], White Shark Optimizer (WSO) [10], and other optimizers have further transformed biological behaviors such as foraging, hiding, hunting, migration, and reproduction into new search operators, continuously expanding the modeling space of nature-inspired optimization.

**Physics-inspired MH algorithms** translate physical laws and phenomena into search dynamics and guide the movement of candidate solutions by simulating the evolution of physical systems. Simulated Annealing (SA) [11] models the energy minimization process of metal annealing and escapes local optima by probabilistically accepting worse solutions. Gravitational Search Algorithm (GSA) [12] simulates the attraction among individuals based on Newton's law of universal gravitation, with better-performing individuals assigned larger masses and thus exerting stronger attraction. More recently, the RIME optimization algorithm (RIME) [13] and Smell Agent Optimization (SAO) [14] have introduced search mechanisms based on rime-ice growth and smell-molecule diffusion, respectively, further enriching the methodological diversity of physics-inspired optimization.

**Human-inspired MH algorithms** extract search mechanisms from human social collaboration, learning, competition, and cognitive behaviors. Harmony Search (HS) [15] emulates the musical improvisation process in which musicians seek a harmonious combination, generating new solutions through harmony memory consideration, pitch adjustment, and random selection. Teaching–Learning-Based Optimization (TLBO) [16] updates candidate solutions through teacher-guided learning and peer interaction in two successive search phases. Among the many recent developments in this category, Catch Fish Optimization Algorithm (CFOA) [17] and Information Acquisition Optimizer (IAO) [18] derive their search mechanisms from traditional fishing behavior, and human information acquisition and decision-making processes, respectively.

**Mathematics-inspired MH algorithms** directly use mathematical functions, numerical iterations, or analytical expressions to construct search operators. Stochastic Fractal Search (SFS) [19] employs fractal diffusion and stochastic updating to maintain population diversity while refining promising regions. Further developments include weIghted meaN oF vectOrs (INFO) [20], RUNge Kutta optimizer (RUN) [21], and other methods that transform mathematical tools such as weighted averaging and numerical integration into search mechanisms.

Although the above MH algorithms differ in their inspirations and update mechanisms, their core objective is generally to achieve an effective balance between *exploration* and *exploitation* in complex non-convex search spaces. However, existing optimizers still face several common challenges. First, the balance between *exploration* and *exploitation* is typically regulated through update equations, switching rules, or scheduling mechanisms. If these mechanisms are not well aligned with the characteristics of the problem landscape, the algorithm may exhibit premature convergence or inefficient search. Second, beyond common settings such as population size and the maximum number of iterations, many algorithms introduce additional mechanism-specific internal parameters. Their performance may therefore be sensitive to parameter configurations, increasing uncertainty across different tasks. Finally, many newly proposed MH algorithms mainly rely on numerical experiments to demonstrate their performance, while theoretical convergence analyses remain relatively limited.

Motivated by the above challenges, ECO formulates its search process based on energy flow and material cycling within ecosystems. Specifically, ECO organizes the population into producers, consumers, and decomposers, whose collaborative updates jointly support global *exploration*, local *exploitation*, and population renewal. A sensitivity analysis over a broad set of benchmark functions identifies a robust internal parameter configuration with the best overall performance and stability, which is recommended to be kept fixed when applying ECO without problem-specific tuning. In addition, the convergence behavior of ECO is further analyzed under standard stochastic search assumptions.

## 3. Ecological Cycle Optimizer (ECO)

This section introduces a novel nature-inspired MH algorithm, referred to as ECO. First, the inspiration for ECO is described, followed by a detailed presentation of its mathematical model and solution process. Subsequently, its convergence properties and computational complexity are discussed.

### 3.1. Inspiration

An ecosystem is a unified whole formed by the interaction between biological communities (plants, animals, microorganisms, *etc.*), and inorganic environment (soil, water, climate, *etc.*), including producers, consumers, decomposers, and non-living matter and energy [22]. Producers, typically plants, are the cornerstone of the ecosystem. They convert carbon dioxide and water into organic matter through photosynthesis using solar energy, and provide food and oxygen for consumers, marking the beginning of energy flow within the ecosystem. Consumers, primarily including herbivores, carnivores, and omnivores, occupy different trophic levels of the food chain and facilitate energy flow and material cycling through predation within the ecosystem. Decomposers break down the remains of plants and animals into inorganic substances, which are reabsorbed by producers, thus contributing to the material cycle.

Energy flow and material cycling are the primary functions of an ecosystem. Matter serves as the carrier of energy, allowing energy to flow unidirectionally along the food chain, progressively decreasing across trophic levels. Energy, as the driving force, enables the continuous cycling of matter between biological communities and the inorganic environment. Through energy flow and material cycling, the components of the ecosystem are closely interconnected, forming a relatively balanced ecological cycle that collectively maintains the balance and stability of the ecosystem.

The inspiration for the ECO comes from the aforementioned natural phenomena, drawing an analogy between dynamic optimization and ecological cycles. Mathematical models are established for the iterative update strategies of producers, consumers, and decomposers, thereby simulating energy flow and material cycling. As the number of iterations increases, the ECO progressively searches for the theoretical global optimal solution, simulating the ecosystem as it reaches an optimal state of balance and stability.

### 3.2. Mathematical model

This section establishes the ECO's mathematical model, simulating the iterative update strategies of producers, consumers, and decomposers.

First, the following four rules are formulated for the ECO algorithm:

(1) The consumers in the population include only herbivores, carnivores, and omnivores, with trophic levels specified as: herbivore < carnivore < omnivore;

(2) Herbivores feed on producers, carnivores feed on herbivores, and omnivores can prey on producers, herbivores, and carnivores. It is specified that all consumers cannot prey on organisms at the same trophic level;

(3) After each iteration cycle, producers, herbivores, carnivores, and omnivores are all decomposed by decomposers;

(4) The fitness of individuals in the population is inversely proportional to their energy; that is, higher energy indicates lower fitness;

(5) After each individual in the population is updated iteratively, it is retained if its fitness improves, and eliminated if its fitness worsens, while the individual's historical best is still preserved, thereby simulating the survival of the fittest within the ecosystem.

Based on these rules, Fig. 1 visually presents a schematic of an ecosystem.

For the ECO algorithm, population initialization is performed by random sampling within the constrained space. During the iterative search phase, the roulette wheel selection-based multi-elite guiding strategy is used to guide selection, while the fitness value-based method is employed for population updates to simulate the survival of the fittest in nature. For the most critical convergence search operators that influence the ECO's *exploration* and *exploitation* phases, unique strategies for producers, consumers and decomposers are proposed and detailed in this section.

#### *3.2.1. Population initialization*

Before the algorithm begins, the population needs to be initialized. The population size is defined as $N_{\mathrm{pop}}$, with the number of producers being $N_{\mathrm{Pro}}$. Among consumers, the numbers of herbivores, carnivores, and omnivores are $N_{\mathrm{Her}}$, $N_{\mathrm{Car}}$, and $N_{\mathrm{Omn}}$, respectively. Therefore, $N_{\mathrm{pop}} = N_{\mathrm{Pro}} + N_{\mathrm{Her}} + N_{\mathrm{Car}} + N_{\mathrm{Omn}}$. Let $P_{\mathrm{Pro}}$, $P_{\mathrm{Her}}$, $P_{\mathrm{Car}}$, and $P_{\mathrm{Omn}}$ denote the population compositions of producers, herbivores, carnivores, and omnivores, respectively, where $P_{\mathrm{Pro}} + P_{\mathrm{Her}} + P_{\mathrm{Car}} + P_{\mathrm{Omn}} = 1$. The corresponding population sizes are calculated as $N_{\mathrm{Pro}} = N_{\mathrm{pop}} \times P_{\mathrm{Pro}}$, $N_{\mathrm{Her}} = N_{\mathrm{pop}} \times P_{\mathrm{Her}}$, $N_{\mathrm{Car}} = N_{\mathrm{pop}} \times P_{\mathrm{Car}}$, and $N_{\mathrm{Omn}} = N_{\mathrm{pop}} \times P_{\mathrm{Omn}}$, respectively. In implementation, these values are rounded to integer population sizes. The objective function of the optimization problem is $f(\cdot)$, with a dimension of $D$. The upper and lower bounds of the constrained space are $\boldsymbol{U}_{\mathrm{b}} \in \mathbb{R}^D$ and $\boldsymbol{L}_{\mathrm{b}} \in \mathbb{R}^D$, respectively. The current number of iterations is $k$, and the maximum number of iterations is $k_{\max}$, hence $1 \leqslant k \leqslant k_{\max}$.

Randomly initialize population individual $\boldsymbol{X}_i \in \mathbb{R}^D$ within the constrained space:

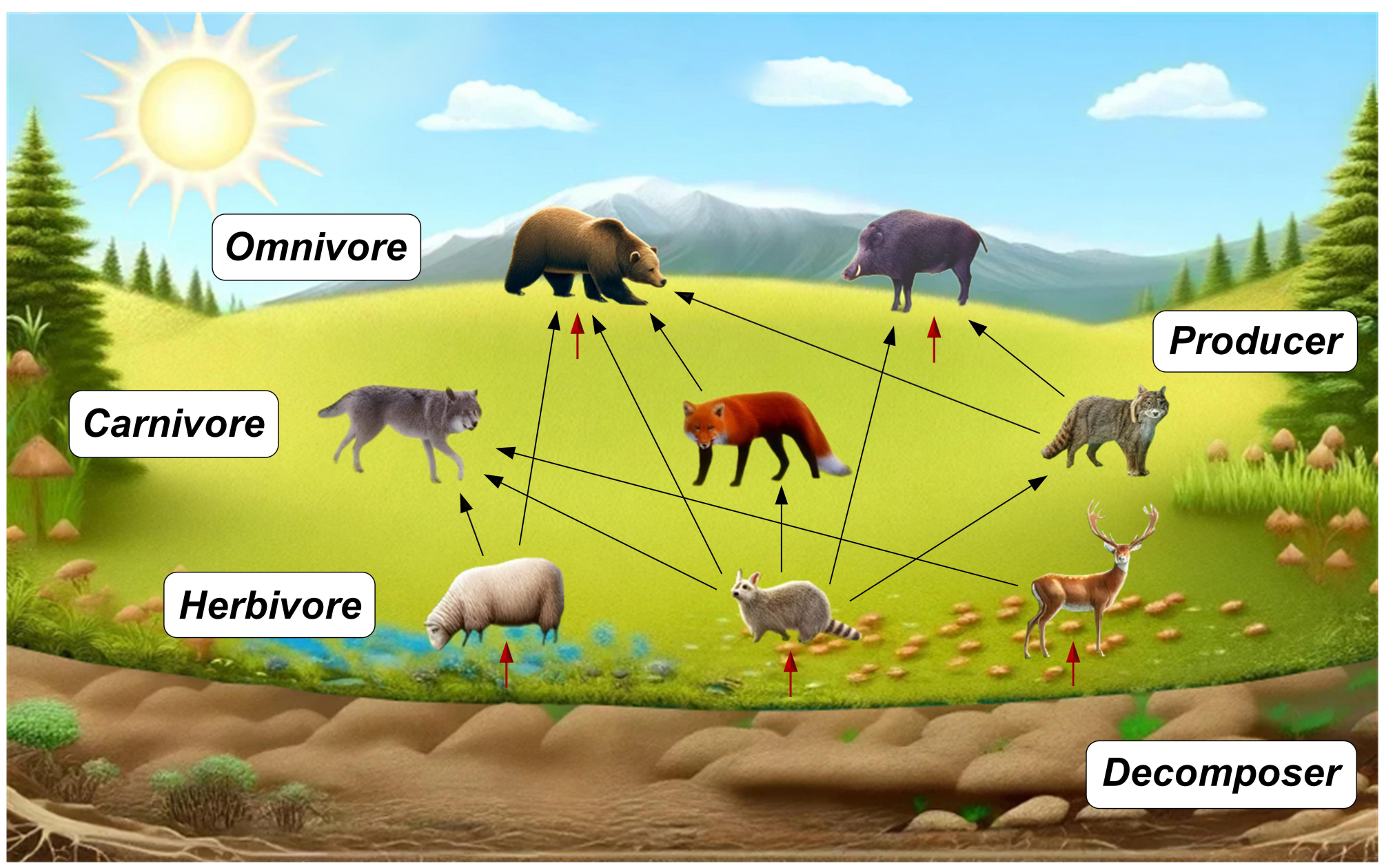

**Figure 1:** Schematic of an ecosystem.

$$\boldsymbol{X}_i = \boldsymbol{L}_\mathrm{b} + \mathrm{rand}(\boldsymbol{U}_\mathrm{b} - \boldsymbol{L}_\mathrm{b}), \quad i = 1, 2, ..., N_\mathrm{pop}, \tag{1}$$

where rand $\in \mathbb{R}$ is a random number in the range [0, 1].

Therefore, producers $\boldsymbol{X}_\mathrm{Pro} \in \mathbb{R}^{N_\mathrm{Pro} \times D}$, herbivores $\boldsymbol{X}_\mathrm{Her} \in \mathbb{R}^{N_\mathrm{Her} \times D}$, carnivores $\boldsymbol{X}_\mathrm{Car} \in \mathbb{R}^{N_\mathrm{Car} \times D}$, and omnivores $\boldsymbol{X}_\mathrm{Omn} \in \mathbb{R}^{N_\mathrm{Omn} \times D}$ are initialized as follows:

$$\boldsymbol{X}_\mathrm{Pro} = \left[\boldsymbol{X}_i\right]_{i=1}^{N_\mathrm{Pro}}, \ \boldsymbol{X}_\mathrm{Her} = \left[\boldsymbol{X}_i\right]_{i=N_\mathrm{Pro}+1}^{N_\mathrm{Pro}+N_\mathrm{Her}}, \ \boldsymbol{X}_\mathrm{Car} = \left[\boldsymbol{X}_i\right]_{i=N_\mathrm{Pro}+N_\mathrm{Her}+1}^{N_\mathrm{Pro}+N_\mathrm{Her}+N_\mathrm{Car}}, \ \boldsymbol{X}_\mathrm{Omn} = \left[\boldsymbol{X}_i\right]_{i=N_\mathrm{Pro}+N_\mathrm{Her}+N_\mathrm{Car}+1}^{N_\mathrm{pop}}, \tag{2}$$

where $\boldsymbol{X}_\mathrm{Pro}$, $\boldsymbol{X}_\mathrm{Her}$, $\boldsymbol{X}_\mathrm{Car}$, and $\boldsymbol{X}_\mathrm{Omn}$ are formed by vertically stacking $\boldsymbol{X}_i$ over the indicated index ranges.

During each loop iteration, after the producer and consumer update according to their respective strategy, the individuals in the population are decomposed one by one based on the decomposition strategy to generate inorganic substances. In ECO, these inorganic substances are regarded as decomposers $\boldsymbol{X}_\mathrm{Dec} \in \mathbb{R}^{N_\mathrm{pop} \times D}$, so there is no need to initialize the decomposers.

Additionally, when updating the population individuals, if their position exceeds the bounds of the constrained space, they are re-initialized according to (1).

### 3.2.2. Producer strategy

During material cycling within the ecosystem, producers can reabsorb the inorganic substances generated by decomposers from the decomposition of plant and animal remains to meet their nutritional needs.

In ECO, the individuals formed after the decomposition of producers and consumers simulate the inorganic substances produced by decomposers, while the current producer individuals are regarded as the existing nutrients. During each iteration, the producers select $N_\mathrm{Pro}$ units with the highest energy (*i.e.*, lowest fitness) from the existing nutrients and inorganic substances to absorb nutrients, considering this process as the updating strategy for producers, expressed as follows:

$$\boldsymbol{X}_\mathrm{nut}(k+1) = \mathrm{Sort}\begin{bmatrix}\boldsymbol{X}_\mathrm{Pro}(k) \\ \boldsymbol{X}_\mathrm{Dec}(k)\end{bmatrix}, \tag{3}$$

$$\boldsymbol{X}_\mathrm{Pro}(k+1) = \left[\boldsymbol{X}_\mathrm{nut}(k+1)\right]_{1:N_\mathrm{Pro}}, \tag{4}$$

where $\boldsymbol{X}_\mathrm{nut} \in \mathbb{R}^{(N_\mathrm{Pro}+N_\mathrm{Dec}) \times D}$ is the nutrients required by producers, sorted by energy level, Sort $[\cdot]$ is the sorting function that reorders the elements according to certain rules, and $[\cdot]_{1:N_\mathrm{Pro}}$ selects the elements from index 1 to $N_\mathrm{Pro}$.

The sorting rule of the Sort $[\cdot]$ function is based on the fitness values $f\left[\boldsymbol{X}_\mathrm{Pro}(k)\right]$ and $f\left[\boldsymbol{X}_\mathrm{Dec}(k)\right]$ in ascending order, meaning that the elements of $\boldsymbol{X}_\mathrm{Pro}(k)$ and $\boldsymbol{X}_\mathrm{Dec}(k)$ are rearranged by fitness values from lowest to highest.

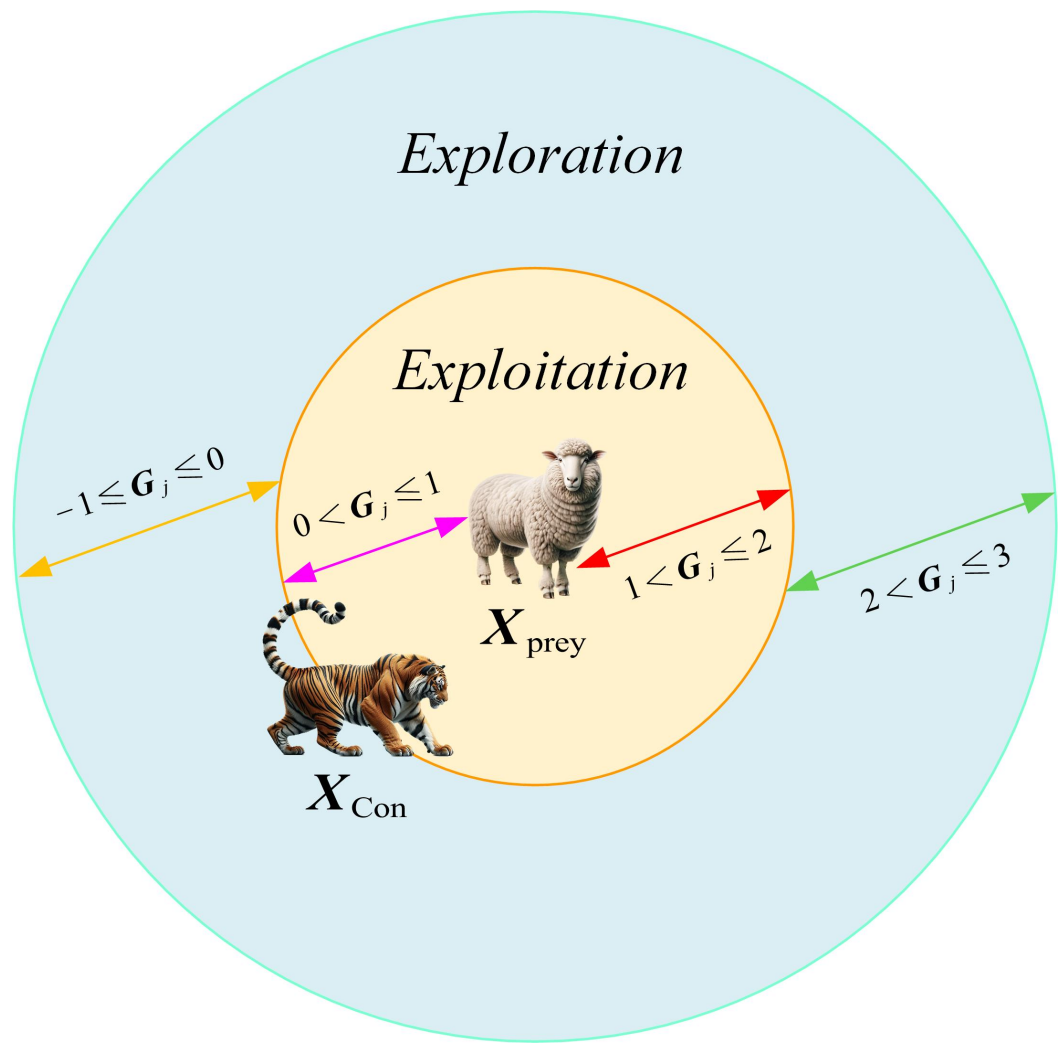

Figure 2: Consumer-prey predatory dynamics.

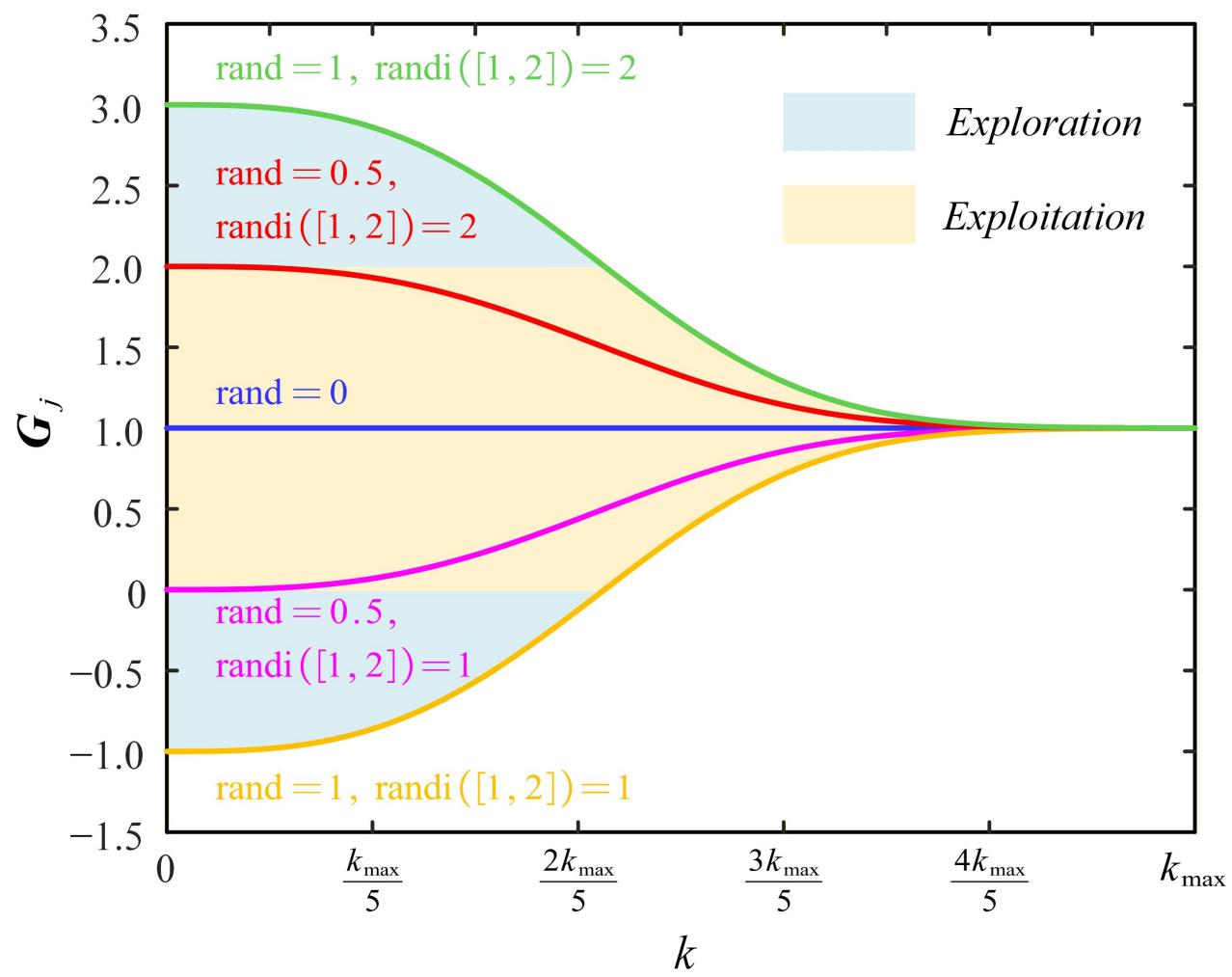

Figure 3: Variation of predation factor $G_j(k)$ over iterations.

### 3.2.3. Consumer strategy

Consumers promote the energy flow and material cycling within the ecosystem by preying on organisms with lower trophic levels than themselves. During the predation process, consumers often lurk in the shadows, constantly monitoring the movements of their prey and waiting for the opportune moment to strike. Therefore, the mathematical model for the predation behavior of the consumers is designed as follows:

$$\boldsymbol{X}_{\text{Con}}(k+1) = \boldsymbol{X}_{\text{Con}}(k) + \boldsymbol{G}(k) \otimes \left\{ \text{rand} \left[ \boldsymbol{X}_{\text{prey}}(k+1) - \boldsymbol{X}_{\text{Con}}(k) \right] \right\}, \tag{5}$$

where the operator $\otimes$ denotes element-wise multiplication between two vectors. Here, $\boldsymbol{X}_{\text{Con}} \in \mathbb{R}^D$ represents the position of the consumer individual, $\boldsymbol{X}_{\text{prey}} \in \mathbb{R}^D$ represents the position of the prey, and $\boldsymbol{G} \in \mathbb{R}^D$ denotes the predation factor vector corresponding to the consumer.

Fig. 2 depicts the schematic of the consumer preying on the prey. There are multiple factors in nature that jointly constrain the predation behavior of consumers. For instance, obstacles present along the predation path may hinder the predator's movement and pursuit; additionally, the prey's camouflage, defense mechanisms, and ability to flee further reduce the success rate of predation. Therefore, for the predation model (5), random numbers within the interval [0, 1] are employed to simulate these complex stochastic factors, adding random weights to the predator's movement vector toward the prey. This enhances the randomness in the ECO's optimization process, thereby contributing to the prevention of the algorithm becoming trapped in local optima.

(1) **Predation factor vector function $\boldsymbol{G}$**

Each dimension $\boldsymbol{G}_j$ of the consumer's predation factor vector function $\boldsymbol{G}$ is calculated according to:

$$\boldsymbol{G}_j(k) = 1 + 2 \text{ rand } e^{-9\left(\frac{k}{k_{\max}}\right)^3} (-1)^{\text{randi}([1,2])}, \quad j = 1, 2, ..., D, \tag{6}$$

where randi ([1, 2]) randomly takes an integer value of 1 or 2.

The graph of $\boldsymbol{G}_j(k)$ is shown in Fig. 3. The incorporation of the random numbers rand and randi ([1, 2]) in $\boldsymbol{G}_j(k)$ allows it to take values within the interval $[-1, 3]$, enabling ECO to exhibit different behaviors at various phases of the iteration process. By combining Fig. 2 and Fig. 3, we further analyze the impact of the predation factor $\boldsymbol{G}$ on the performance of the ECO.

In the initial phase of the iteration, $\boldsymbol{G}_j$ can randomly take values within the interval $[-1, 0] \cup (2, 3]$, prompting ECO to exhibit a greater tendency toward *exploration*. The algorithm extensively searches the regions where the global optimum may exist, reducing the risk of becoming trapped in local optima. Such behavior is akin to a situation in which the current prey fails to satisfy the consumer's energy needs, and the consumer has not yet found the prey that maximizes its energy intake (*i.e.*, the prey with the lowest fitness). To conserve energy, the consumer abandons the current target and searches a broader region for more energy-rich prey.

As the number of iterations $k$ increases, $\boldsymbol{G}_j$ adaptively adjusts and gradually reduces its range of random values, allowing ECO to transition from *exploration* to *exploitation*. When $\boldsymbol{G}_j \in (0, 2]$, ECO enters the *exploitation* phase, intensively searching for the global optimum within a multidimensional hypersphere of random radius $\left\|\boldsymbol{X}_{\text{prey}} - \boldsymbol{X}_{\text{Con}}\right\|$. Such behavior is akin to a situation where the consumer launches an attack on the prey within a circular region centered on the prey, with a radius of $\left\|\boldsymbol{X}_{\text{prey}} - \boldsymbol{X}_{\text{Con}}\right\|$.

Due to $\lim_{k\to k_{\max}} \boldsymbol{G}_j = 1$, in the later phase of the iteration, ECO continuously performs *exploitation* within the neighborhood of the global optimum, while refining and uncovering the highly precise global optimum $\boldsymbol{X}_{\text{best}}$.

(2) **Herbivore predation strategy**

In ECO, each herbivore is only allowed to prey on three producers, guided by the top three elite individuals. The predation principle applies the roulette wheel selection [23, 24] to the producer population, with selection determined probabilistically based on their energy levels. The higher the energy of a producer (*i.e.*, the smaller the fitness), the greater the probability of being preyed upon. Define the $i$th herbivore as $\boldsymbol{X}_{\text{Her}\,(i)} \in \mathbb{R}^D$ ($i = 1, 2, ..., N_{\text{Her}}$), and the three producers it preys on as $\boldsymbol{X}_{\text{Pro}\,(\text{A})} \in \mathbb{R}^D$, $\boldsymbol{X}_{\text{Pro}\,(\text{B})} \in \mathbb{R}^D$, and $\boldsymbol{X}_{\text{Pro}\,(\text{C})} \in \mathbb{R}^D$. Based on the consumer predation model (5), the herbivore's predation strategy is designed as follows:

$$\begin{aligned}\boldsymbol{X}_{\text{Her}\,(i)}(k+1) = \boldsymbol{X}_{\text{Her}\,(i)}(k) + \boldsymbol{G}(k) \otimes \big\{&\text{rand}\left[\boldsymbol{X}_{\text{Pro}\,(A)}(k+1) - \boldsymbol{X}_{\text{Her}\,(i)}(k)\right] \\ &+ \text{rand}\left[\boldsymbol{X}_{\text{Pro}\,(B)}(k+1) - \boldsymbol{X}_{\text{Her}\,(i)}(k)\right] + \text{rand}\left[\boldsymbol{X}_{\text{Pro}\,(C)}(k+1) - \boldsymbol{X}_{\text{Her}\,(i)}(k)\right]\big\}, \quad i = 1, 2, ..., N_{\text{Her}}.\end{aligned} \tag{7}$$

(3) **Carnivore predation strategy**

ECO specifies that each carnivore preys on three herbivores using the roulette wheel selection rule. Define the $i$th carnivore as $\boldsymbol{X}_{\text{Car}\,(i)} \in \mathbb{R}^D$ ($i = 1, 2, ..., N_{\text{Car}}$), with $\boldsymbol{X}_{\text{Her}\,(\text{A})} \in \mathbb{R}^D$, $\boldsymbol{X}_{\text{Her}\,(\text{B})} \in \mathbb{R}^D$, and $\boldsymbol{X}_{\text{Her}\,(\text{C})} \in \mathbb{R}^D$ representing the three preyed herbivores. According to (5), the predation strategy of the carnivore is expressed as follows:

$$\begin{aligned}\boldsymbol{X}_{\text{Car}\,(i)}(k+1) = \boldsymbol{X}_{\text{Car}\,(i)}(k) + \boldsymbol{G}(k) \otimes \big\{&\text{rand}\left[\boldsymbol{X}_{\text{Her}\,(A)}(k+1) - \boldsymbol{X}_{\text{Car}\,(i)}(k)\right] \\ &+ \text{rand}\left[\boldsymbol{X}_{\text{Her}\,(B)}(k+1) - \boldsymbol{X}_{\text{Car}\,(i)}(k)\right] + \text{rand}\left[\boldsymbol{X}_{\text{Her}\,(C)}(k+1) - \boldsymbol{X}_{\text{Car}\,(i)}(k)\right]\big\}, \quad i = 1, 2, ..., N_{\text{Car}}.\end{aligned} \tag{8}$$

(4) **Omnivore predation strategy**

The omnivore randomly preys on one producer, one herbivore, and two carnivores based on the roulette wheel selection, which constitutes a four-elite guiding approach. Define the $i$th omnivore as $\boldsymbol{X}_{\text{Omn}\,(i)} \in \mathbb{R}^D$ ($i = 1, 2, ..., N_{\text{Omn}}$), with $\boldsymbol{X}_{\text{Pro}\,(\text{D})} \in \mathbb{R}^D$ and $\boldsymbol{X}_{\text{Her}\,(\text{D})} \in \mathbb{R}^D$ representing the preyed producer and herbivore, respectively, and $\boldsymbol{X}_{\text{Car}\,(\text{A})} \in \mathbb{R}^D$ and $\boldsymbol{X}_{\text{Car}\,(\text{B})} \in \mathbb{R}^D$ as the two preyed carnivores. In accordance with (5), the predation strategy of the omnivore is presented as follows:

$$\begin{aligned}\boldsymbol{X}_{\text{Omn}\,(i)}(k+1) = \boldsymbol{X}_{\text{Omn}\,(i)}(k) + \boldsymbol{G}(k) \otimes \big\{&\text{rand}\left[\boldsymbol{X}_{\text{Pro}\,(D)}(k+1) - \boldsymbol{X}_{\text{Omn}\,(i)}(k)\right] + \text{rand}\left[\boldsymbol{X}_{\text{Her}\,(D)}(k+1) - \boldsymbol{X}_{\text{Omn}\,(i)}(k)\right] \\ &+ \text{rand}\left[\boldsymbol{X}_{\text{Car}\,(A)}(k+1) - \boldsymbol{X}_{\text{Omn}\,(i)}(k)\right] + \text{rand}\left[\boldsymbol{X}_{\text{Car}\,(B)}(k+1) - \boldsymbol{X}_{\text{Omn}\,(i)}(k)\right]\big\}, \quad i = 1, 2, ..., N_{\text{Omn}}.\end{aligned} \tag{9}$$

(5) **Roulette wheel selection for elite individuals**

ECO adopts the roulette wheel selection-based multi-elite guiding strategy, which increases the probability of consumers preying on higher-energy (*i.e.*, lower-fitness) prey to meet their energy intake requirements. This behavior leads to local *exploitation* by the algorithm in the neighborhood of the optimal solution. Simultaneously, it avoids the drawback of prioritizing the selection of all elite individuals, which could lead the algorithm into local optima and premature convergence, thereby enhancing the global *exploration* capability of ECO.

In the candidate pool for the roulette wheel selection, individuals with lower fitness are considered superior and have a higher probability of being selected. Taking the producer population as an example, the number of individuals participating in the roulette wheel selection is $N_{\text{Pro}}$. Define the $i$th producer as $\boldsymbol{X}_{\text{Pro}\,(i)} \in \mathbb{R}^D$ ($i = 1, 2, ..., N_{\text{Pro}}$), and its probability of being selected is given by:

$$P_{\text{ECO}}\left[\boldsymbol{X}_{\text{Pro}\,(i)}\right] = \frac{1\Big/ f\left[\boldsymbol{X}_{\text{Pro}\,(i)}\right]}{\sum_{j=1}^{N_{\text{Pro}}} 1\Big/ f\left[\boldsymbol{X}_{\text{Pro}\,(j)}\right]}. \tag{10}$$

The sum of the probabilities of the first $i$ individuals represents the cumulative probability corresponding to $\boldsymbol{X}_{\text{Pro}\,(i)}$, expressed as follows:

$$q_{\text{ECO}}\left[\boldsymbol{X}_{\text{Pro}\,(i)}\right] = \sum_{j=1}^{i} P_{\text{ECO}}\left[\boldsymbol{X}_{\text{Pro}\,(j)}\right]. \tag{11}$$

A random number rand is generated within the interval [0, 1]. If rand $\leqslant q_{\mathrm{ECO}}\left[\boldsymbol{X}_{\mathrm{Pro}\,(1)}\right]$, the first individual is selected. Otherwise, the process iterates until $q_{\mathrm{ECO}}\left[\boldsymbol{X}_{\mathrm{Pro}\,(i-1)}\right] < \mathrm{rand} \leqslant q_{\mathrm{ECO}}\left[\boldsymbol{X}_{\mathrm{Pro}\,(i)}\right]$ ($i = 2, 3, ..., N_{\mathrm{Pro}}$), selecting the $i$th individual. To select multiple elite individuals, this process is repeated the corresponding number of times.

The pseudocode of the roulette wheel selection for elite individuals is shown in Algorithm 1.

#### 3.2.4. Decomposer strategy

Decomposers constitute a critical part of the ecosystem, facilitating the breakdown of plant and animal remains, as well as excretions, into inorganic substances, thus promoting ecological cycling and ensuring the stability of the ecosystem. In each iteration of ECO, decomposers are required to decompose producers and consumers sequentially.

**Algorithm 1:** Pseudocode of the roulette wheel selection for elite individuals.

```
Input: Population individual fitness f[X_Pro(i)] (i = 1, 2, ..., N_Pro) and the number of selected individuals (i.e., the
       number of roulette wheel selections)
Output: All selected individuals
1  Probability P_ECO[X_Pro(i)] of each individual being selected by the roulette wheel selection according to (10);
2  Cumulative probability q_ECO[X_Pro(i)] corresponding to each individual calculated according to (11);
3  for each roulette wheel selection do
4  |   Generate a random number rand within [0, 1];
5  |   if rand ⩽ q_ECO[X_Pro(i)] then
6  |   |   Select the first individual;
7  |   else
8  |   |   while no individual selected do
9  |   |   |   if q_ECO[X_Pro(i−1)] < rand ⩽ q_ECO[X_Pro(i)] (i = 2, 3, ..., N_Pro) then
10 |   |   |   |   Select the ith individual;
11 |   |   |   |   break;
12 |   |   |   end
13 |   |   end
14 |   end
15 |   Record the selected individual;
16 end
17 return all selected individuals
```

Accordingly, this section presents a mathematical model for the decomposition process and introduces three probability-based decomposition strategies to balance the process of *exploration* and *exploitation* in ECO.

For each individual in the population, the probability of undergoing optimal decomposition is controlled by $p_{\mathrm{opt}}$, while the remaining probability is assigned to random decomposition. In the case of random decomposition, $p_{\mathrm{loc}}$ controls the probability of performing local random decomposition, and $1 - p_{\mathrm{loc}}$ controls the probability of performing global random decomposition.

(1) **Optimal decomposition**

When any individual $\boldsymbol{X}_i$ $\left(i \in \left[1, 2, ..., N_{\mathrm{pop}}\right]\right)$ in the population is selected with probability $p_{\mathrm{opt}}$ for optimal decomposition, the individual $\boldsymbol{X}_{\mathrm{bestk}} \in \mathbb{R}^D$ with the highest energy (*i.e.*, lowest fitness) in the current iteration is chosen. The selected individual is decomposed toward the neighborhood of $\boldsymbol{X}_{\mathrm{bestk}}$, promoting the decomposer to provide the most nutrient-rich inorganic matter to the producers, thereby enhancing material cycling. This process will significantly enhance the *exploitation* capability of ECO, improving its convergence speed and solution accuracy.

The multidimensional neighborhood $\boldsymbol{X}_{\mathrm{nei}} \in \mathbb{R}^D$ of $\boldsymbol{X}_{\mathrm{bestk}}$ is defined as follows:

$$\boldsymbol{X}_{\mathrm{nei},\,j}\,(k+1) = \mathrm{rand}\,\boldsymbol{X}_{\mathrm{bestk},\,j}\,(k+1)\,, \quad j = 1, 2, ..., D, \tag{12}$$

where $\boldsymbol{X}_{\mathrm{nei},\,j} \in \mathbb{R}$ and $\boldsymbol{X}_{\mathrm{bestk},\,j} \in \mathbb{R}$ are $j$th dimensional elements of the neighborhood $\boldsymbol{X}_{\mathrm{nei}}$ and the best individual $\boldsymbol{X}_{\mathrm{bestk}}$ in the current iteration, respectively.

The population individual $\boldsymbol{X}_i$ is decomposed along the direction of $\boldsymbol{X}_{\mathrm{nei}} - \boldsymbol{X}_i$ to a random position within a distance $\pm 0.2 \left\|\boldsymbol{X}_{\mathrm{nei}} - \boldsymbol{X}_i\right\|$ from the center $\boldsymbol{X}_{\mathrm{nei}}$, and the inorganic matter generated at this point is regarded as the $i$th decomposer $\boldsymbol{X}_{\mathrm{Dec}\,(i)} \in \mathbb{R}^D$. The optimal decomposition strategy is designed as follows:

$$\boldsymbol{X}_{\mathrm{Dec}\,(i)}\,(k+1) = \boldsymbol{X}_{\mathrm{nei}}\,(k+1) + (0.4\ \mathrm{rand} - 0.2)\left[\boldsymbol{X}_{\mathrm{nei}}\,(k+1) - \boldsymbol{X}_i\,(k+1)\right], \quad i \in \left[1, 2, ..., N_{\mathrm{pop}}\right]. \tag{13}$$

(2) **Local random decomposition**

The remaining population individuals undergo random decomposition, with probability $p_{\mathrm{loc}}$ of being further selected for local random decomposition. In this process, individual $\boldsymbol{X}_i$ is radially decomposed to a random position within a multidimensional hypersphere centered at itself, with an *exploration* radius of $\left\|\boldsymbol{X}_{\mathrm{bestk}}-\boldsymbol{X}_i\right\|$. The range of the radial distance is $\left[0, \left\|\boldsymbol{X}_{\mathrm{bestk}}-\boldsymbol{X}_i\right\|\right]$.

Define a random vector $\boldsymbol{V}_{\mathrm{rand}} \in \mathbb{R}^D$, with each of its elements as:

$$\boldsymbol{V}_{\mathrm{rand},j} = 2\,\mathrm{rand} - 1, \quad j = 1, 2, \ldots, D. \tag{14}$$

The local random decomposition strategy is designed as follows:

$$\boldsymbol{X}_{\mathrm{Dec}(i)}(k+1) = \boldsymbol{X}_i(k+1) + \mathrm{rand}\left\|\boldsymbol{X}_{\mathrm{bestk}}(k+1) - \boldsymbol{X}_i(k+1)\right\| \frac{\boldsymbol{V}_{\mathrm{rand}}}{\left\|\boldsymbol{V}_{\mathrm{rand}}\right\|}, \quad i \in \left[1, 2, \ldots, N_{\mathrm{pop}}\right]. \tag{15}$$

In the initial phase of the iteration, local random decomposition enables population individuals to perform radial *exploration* within a larger hypersphere, facilitating the search for regions that may contain the global optimum. As iterations increase, most population individuals converge near $\boldsymbol{X}_{\mathrm{bestk}}$ in the later phase of the algorithm, leading to a gradual decrease in the maximum *exploration* radius $\left\|\boldsymbol{X}_{\mathrm{bestk}}-\boldsymbol{X}_i\right\|$. Individuals continue to search for the optimal solution around $\boldsymbol{X}_{\mathrm{bestk}}$ with a certain probability, further enhancing the algorithm's *exploitation* capability. Therefore, the local random decomposition strategy effectively balances the *exploration* and *exploitation* processes of the ECO.

(3) **Global random decomposition**

The rest of the population is selected for global random decomposition. In this process, individual $\boldsymbol{X}_i$ is decomposed to a random position through a random walk, enhancing the algorithm's global *exploration* capability. The maximum step size for the random walk is set to $2/3 \min\left(\boldsymbol{L}_{\mathrm{b}} - \boldsymbol{U}_{\mathrm{b}}\right)$ across all dimensions of the constrained space, in order to prevent the random walk from exceeding the boundaries of the constrained space. The step size of the random walk is denoted as $\boldsymbol{w}_{\mathrm{rand}} \in \mathbb{R}^D$, with each dimension calculated as follows:

$$\boldsymbol{w}_{\mathrm{rand},j} = \frac{2}{3}\,\mathrm{rand}\, H(k) \min\left(\boldsymbol{L}_{\mathrm{b}} - \boldsymbol{U}_{\mathrm{b}}\right), \quad j = 1, 2, \ldots, D, \tag{16}$$

where $H(\cdot)$ is a global random walk coefficient function, designed as follows:

$$H(k) = \cos(\mathrm{rand}\,\pi)\left(1 - \frac{k}{1.5k_{\max}}\right)^{\left(\frac{5k}{k_{\max}}\right)}. \tag{17}$$

Define a weight $\mathrm{rand}_{wei} \in \mathbb{R}$ for the random walk as a random number within the interval $[0, 1]$, and apply a random weight to individual $\boldsymbol{X}_i$ and the step size $\boldsymbol{w}_{\mathrm{rand}}$ of the random walk, thereby increasing the randomness of the *exploration*. Thus, the global random decomposition strategy is designed as follows:

$$\boldsymbol{X}_{\mathrm{Dec}(i)}(k+1) = \mathrm{rand}_{wei}\,\boldsymbol{X}_i(k+1) + \left(1 - \mathrm{rand}_{wei}\right)\boldsymbol{w}_{\mathrm{rand}}, \quad i \in \left[1, 2, \ldots, N_{\mathrm{pop}}\right]. \tag{18}$$

With the introduction of rand in function $H$, it takes random values within the interval $[-1, 1]$, and $\lim_{k \to k_{\max}} H = 0$. In the initial phase of the iteration, the value range of $H$ is relatively large, leading to a larger random walk step size $\boldsymbol{w}_{\mathrm{rand}}$, enabling population individuals to perform extensive *exploration* within the constrained space. With increasing iterations, the value range of $H$ adaptively decreases, and the random walk step size decreases accordingly. In the later phase of the iteration, as convergence stabilizes, ECO demonstrates a strong capability to escape local optima, particularly for objective functions where many local optima are densely distributed around the global optimum.

## 3.3. ECO workflow

The pseudocode and flowchart of the ECO are presented in Algorithm 2 and Fig. 4, respectively.

Many previous metaheuristic algorithms rely on multiple parameters that significantly influence their performance and are highly sensitive to parameter settings. To achieve an optimal balance between *exploration* and *exploitation* for superior solution performance, these algorithms often require extensive parameter tuning.

In contrast, the mathematical model and solution process of the ECO indicate that its implementation is more convenient, with the population size $N_{\mathrm{pop}}$ and the maximum number of iterations $k_{\max}$ serving as the primary external control parameters. Meanwhile, the population compositions and decomposition probabilities are regarded as internal parameters. A sensitivity analysis over a broad set of benchmark functions identifies a robust default internal parameter configuration with strong overall performance and stability, which is then kept fixed across all subsequent experiments and engineering applications without problem-specific tuning. The unique strategies of production, consumption, and decomposition in ECO allow the algorithm to dynamically adjust the processes of *exploration* and *exploitation*,

**Algorithm 2:** Pseudocode of the Ecological Cycle Optimizer (ECO).

**Input:** Population size $N_{\text{pop}}$, the maximum number of iterations $k_{\max}$, objective function $f(\cdot)$, dimension $D$, and the upper bound $\boldsymbol{U}_{\text{b}}$ and lower bound $\boldsymbol{L}_{\text{b}}$ of the constrained space

**Output:** The optimal solution $\boldsymbol{X}_{\text{best}}$ and optimal fitness $f\left(\boldsymbol{X}_{\text{best}}\right)$

1 Calculate the numbers of producers $N_{\text{Pro}}$, herbivores $N_{\text{Her}}$, carnivores $N_{\text{Car}}$, and omnivores $N_{\text{Omn}}$ proportionally;
2 Initialize $\boldsymbol{X}_{\text{Pro}}$, $\boldsymbol{X}_{\text{Her}}$, $\boldsymbol{X}_{\text{Car}}$, and $\boldsymbol{X}_{\text{Omn}}$ sequentially according to (1) and (2);
3 **for** *each iteration* $k$ $(k = 1, 2, ..., k_{\max})$ **do**
4     **if** $k \neq 1$ **then**
5         Producer update rule: apply production strategy (3) and (4) to update $\boldsymbol{X}_{\text{Pro}}$;
6     **end**
7     Calculate the predation factor vector $\boldsymbol{G}$ of the consumer according to (6);
8     **for** *each herbivore* $\boldsymbol{X}_{\text{Her}\,(i)}$ $(i = 1, 2, ..., N_{\text{Her}})$ **do**
9         Select three producers $\boldsymbol{X}_{\text{Pro}\,(\text{A})}$, $\boldsymbol{X}_{\text{Pro}\,(\text{B})}$, and $\boldsymbol{X}_{\text{Pro}\,(\text{C})}$ using roulette wheel selection Algorithm 1;
10         Herbivore update rule: apply predation strategy (7) to update $\boldsymbol{X}_{\text{Her}\,(i)}$;
11         Perform function evaluation of $f\left(\boldsymbol{X}_{\text{Her}\,(i)}\right)$;
12     **end**
13     **for** *each carnivore* $\boldsymbol{X}_{\text{Car}\,(i)}$ $(i = 1, 2, ..., N_{\text{Car}})$ **do**
14         Select three herbivores $\boldsymbol{X}_{\text{Her}\,(\text{A})}$, $\boldsymbol{X}_{\text{Her}\,(\text{B})}$, and $\boldsymbol{X}_{\text{Her}\,(\text{C})}$ using roulette wheel selection Algorithm 1;
15         Carnivore update rule: apply predation strategy (8) to update $\boldsymbol{X}_{\text{Car}\,(i)}$;
16         Perform function evaluation of $f\left(\boldsymbol{X}_{\text{Car}\,(i)}\right)$;
17     **end**
18     **for** *each omnivore* $\boldsymbol{X}_{\text{Omn}\,(i)}$ $(i = 1, 2, ..., N_{\text{Omn}})$ **do**
19         Select one producer $\boldsymbol{X}_{\text{Pro}\,(\text{D})}$, one herbivore $\boldsymbol{X}_{\text{Her}\,(\text{D})}$, and two carnivores $\boldsymbol{X}_{\text{Car}\,(\text{A})}$ and $\boldsymbol{X}_{\text{Car}\,(\text{B})}$ using roulette wheel selection Algorithm 1;
20         Omnivore update rule: apply predation strategy (9) to update $\boldsymbol{X}_{\text{Omn}\,(i)}$;
21         Perform function evaluation of $f\left(\boldsymbol{X}_{\text{Omn}\,(i)}\right)$;
22     **end**
23     **for** *each decomposer* $\boldsymbol{X}_{\text{Dec}\,(i)}$ $(i = 1, 2, ..., N_{\text{pop}})$ **do**
24         **if** rand $< p_{\text{opt}}$ **then**
25             Optimal decomposition rule: apply (13) to perform decomposition of $\boldsymbol{X}_i$ $\left(i \in \left[1, 2, ..., N_{\text{pop}}\right]\right)$;
26         **else**
27             **if** rand $< p_{\text{loc}}$ **then**
28                 Local random decomposition rule: apply (15) to perform decomposition of $\boldsymbol{X}_i$ $\left(i \in \left[1, 2, ..., N_{\text{pop}}\right]\right)$;
29             **else**
30                 Global random decomposition rule: apply (18) to perform decomposition of $\boldsymbol{X}_i$ $\left(i \in \left[1, 2, ..., N_{\text{pop}}\right]\right)$;
31             **end**
32         **end**
33         Perform function evaluation of $f\left(\boldsymbol{X}_{\text{Dec}\,(i)}\right)$;
34     **end**
35     Update the historical optimal population individual $\boldsymbol{X}_{\text{best}}$ and the global optimum $f\left(\boldsymbol{X}_{\text{best}}\right)$;
36 **end**
37 **return** $\boldsymbol{X}_{\text{best}}$ *and* $f\left(\boldsymbol{X}_{\text{best}}\right)$

enhancing its stability and resulting in extremely fast convergence speed, high solution accuracy, and strong ability to escape local optima. Therefore, ECO demonstrates outstanding solution performance when addressing more complex non-convex optimization problems.

**Remark 1.** *According to the parameter sensitivity analysis in Section 5, the recommended default population proportions for ECO are set as* $P_{\text{Pro}} = 0.2$, $P_{\text{Her}} = 0.3$, $P_{\text{Car}} = 0.3$, *and* $P_{\text{Omn}} = 0.2$, *while the default decomposition probabilities are set as* $p_{\text{opt}} = 0.6$ *and* $p_{\text{loc}} = 0.6$. *This default configuration is adopted in all subsequent comparative experiments and engineering applications, and we recommend keeping it fixed when applying ECO without problem-specific parameter tuning.*

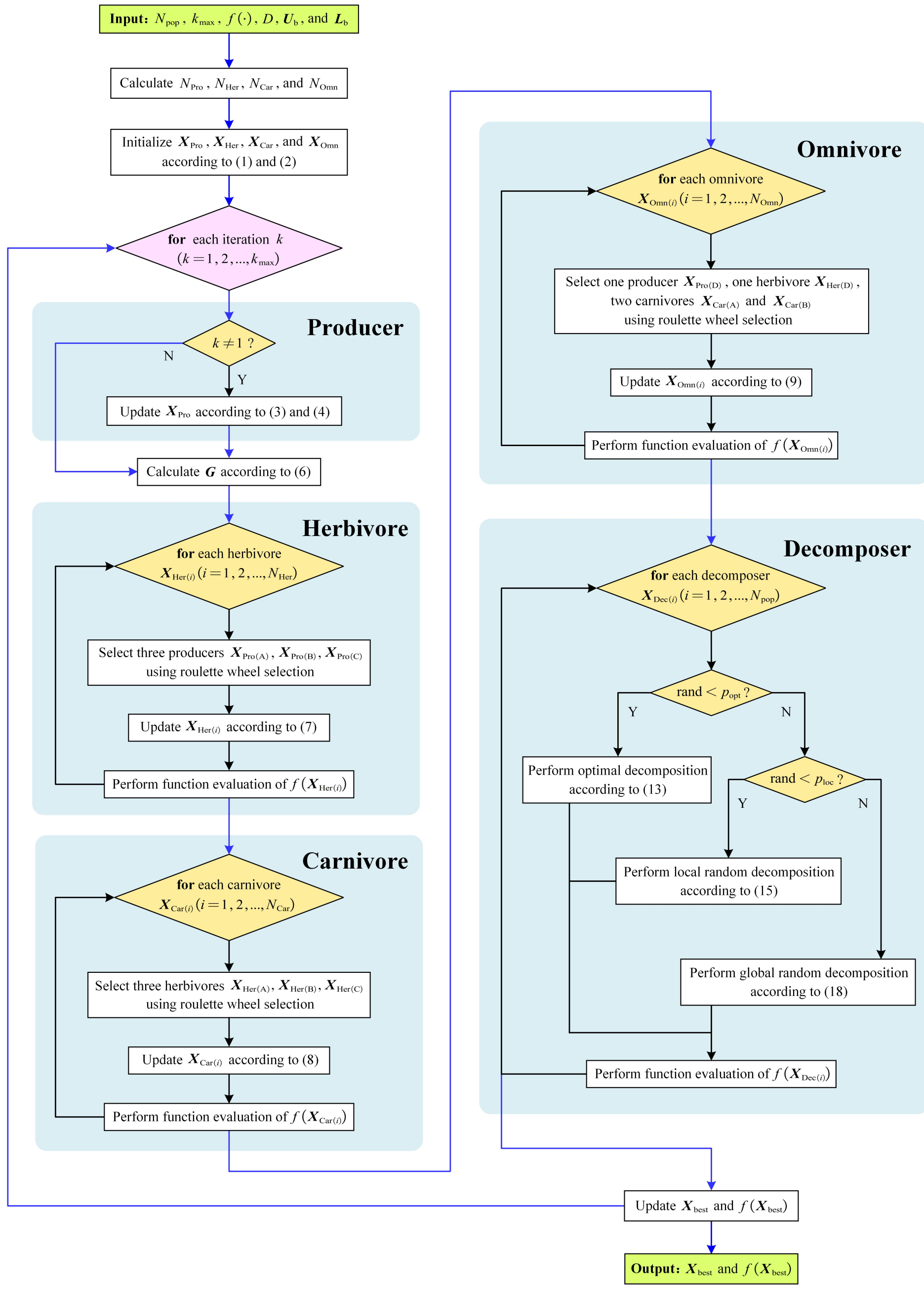

**Figure 4:** Flowchart of Ecological Cycle Optimizer (ECO).

## 3.4. Convergence analysis

Since ECO is a stochastic population-based optimizer, its convergence property is analyzed in terms of the historical optimal population individual maintained by the algorithm, following the standard convergence-analysis framework for stochastic search and evolutionary algorithms [25].

The constrained space is denoted by $\Omega = \left\{ \boldsymbol{x} \in \mathbb{R}^D \mid \boldsymbol{L}_{\text{b}} \leq \boldsymbol{x} \leq \boldsymbol{U}_{\text{b}} \right\}$, and $\nu(\cdot)$ denotes the $D$-dimensional volume, *i.e.*, the Lebesgue measure, of any measurable subset of $\Omega$. $\boldsymbol{X}_{\text{best}}(k)$ is used to denote the historical optimal population individual after the $k$th iteration. Although ECO terminates when the maximum number of iterations $k_{\max}$ is reached, the convergence analysis considers the associated stochastic process as the iteration budget tends to infinity. The finite-budget implementation is discussed in Remark 3.

**Assumption 1.** *The constrained space* $\Omega$ *is a nonempty compact hyperrectangle with positive* $D$*-dimensional volume, i.e.,* $\nu(\Omega) > 0$*.*

**Assumption 2.** *The objective function* $f(\cdot) : \Omega \to \mathbb{R}$ *is continuous on* $\Omega$*.*

**Assumption 3.** *ECO satisfies a positive global reachability condition. Let* $\mathbb{P}(\cdot)$ *denote the probability of an event, and* $\mathbb{P}(\cdot \mid \mathcal{F}_{k-1})$ *the conditional probability given the search history before the kth iteration. Here,* $\mathcal{F}_{k-1}$ *represents all information generated by ECO up to the* $(k-1)$*th iteration, including the current population, historical best solutions, fitness values, and previously generated random variables. Specifically, for any measurable subset* $\mathcal{A} \subseteq \Omega$ *with positive* $D$*-dimensional volume, i.e.,* $\nu(\mathcal{A}) > 0$*, there exists a constant* $\rho_{\mathcal{A}} > 0$*, independent of k, such that*

$$\mathbb{P}\left(C_k \cap \mathcal{A} \neq \emptyset \mid \mathcal{F}_{k-1}\right) \geq \rho_{\mathcal{A}}, \quad k = 1, 2, \ldots, \tag{19}$$

*where* $C_k$ *denotes the set of candidate solutions evaluated at the kth iteration.*

Assumptions 1 and 2 are standard in the convergence analysis of stochastic optimizers [25, 26] and ensure the existence of at least one global optimal solution. Assumption 3 is a standard positive global reachability condition for stochastic search algorithms [25]. It should be understood as a theoretical condition under which the asymptotic convergence of ECO can be established.

For ECO, this condition is related to the stochastic *exploration* mechanisms, including the random population initialization, the boundary re-initialization rule, and the global random decomposition strategy. Specifically, the random initialization and boundary re-initialization operations sample candidate solutions uniformly over the constrained space whenever they are invoked, while the global random decomposition rule introduces random global perturbations that contribute to global *exploration* during the search process. These stochastic exploration mechanisms provide a practical basis for the positive-reachability assumption. Nevertheless, the assumption is explicitly stated to clarify that the convergence result is conditional, without introducing any extra restart mechanism into ECO.

Let $\boldsymbol{x}^* \in \Omega$ be a global optimal solution with global optimal value $f^* = f(\boldsymbol{x}^*)$. For any $\varepsilon > 0$, define the $\varepsilon$-optimal region as

$$\Omega_\varepsilon = \{\boldsymbol{x} \in \Omega \mid f(\boldsymbol{x}) < f^* + \varepsilon\}. \tag{20}$$

Since $f$ is continuous on the compact set $\Omega$, $\Omega_\varepsilon$ has positive $D$-dimensional volume, *i.e.*, $\nu(\Omega_\varepsilon) > 0$, for every $\varepsilon > 0$.

**Lemma 1.** *The sequence* $\{f(\boldsymbol{X}_{\text{best}}(k))\}$ *generated by ECO is monotonically non-increasing and bounded below.*

*Proof.* According to the elitist preservation mechanism of ECO, an updated individual is retained only if its fitness is improved, and the historical best solution is preserved. Therefore, the historical optimal population individual is never replaced by a worse one. For a minimization problem, this gives

$$f(\boldsymbol{X}_{\text{best}}(k+1)) \leq f(\boldsymbol{X}_{\text{best}}(k)), \quad k = 1, 2, \ldots. \tag{21}$$

Since $f^*$ is the global minimum over $\Omega$, we also have

$$f(\boldsymbol{X}_{\text{best}}(k)) \geq f^*. \tag{22}$$

Thus, $\{f(\boldsymbol{X}_{\text{best}}(k))\}$ is monotonically non-increasing and bounded below. □

**Theorem 1.** *Under Assumptions 1–3, the historical best fitness generated by ECO converges almost surely to the global optimal value, i.e.,*

$$\mathbb{P}\left(\lim_{k\to\infty} f(\boldsymbol{X}_{\text{best}}(k)) = f^*\right) = 1. \tag{23}$$

*Proof.* For any fixed $\varepsilon > 0$, the set $\Omega_\varepsilon$ has positive $D$-dimensional volume. By Assumption 3 with $\mathcal{A} = \Omega_\varepsilon$, there exists a constant $\rho_\varepsilon := \rho_{\Omega_\varepsilon} > 0$, independent of $k$, such that

$$\mathbb{P}\left(C_k \cap \Omega_\varepsilon \neq \emptyset \mid \mathcal{F}_{k-1}\right) \geq \rho_\varepsilon, \quad k = 1, 2, \ldots. \tag{24}$$

Let $B_K^\varepsilon$ denote the event that ECO evaluates no candidate solution in $\Omega_\varepsilon$ during the first $K$ iterations. Conditioning on the search history before the $K$th iteration, the probability of missing $\Omega_\varepsilon$ at the $K$th iteration is no larger than $1 - \rho_\varepsilon$. Therefore,

$$\mathbb{P}(B_K^\varepsilon) \leq (1 - \rho_\varepsilon)\mathbb{P}(B_{K-1}^\varepsilon). \tag{25}$$

Applying this inequality recursively gives

$$\mathbb{P}(B_K^\varepsilon) \leq (1 - \rho_\varepsilon)^K. \tag{26}$$

Taking $K \to \infty$ gives

$$\mathbb{P}\left(\bigcap_{k=1}^{\infty}\{C_k \cap \Omega_\varepsilon = \emptyset\}\right) = \lim_{K\to\infty} \mathbb{P}(B_K^\varepsilon) = 0. \tag{27}$$

Therefore, with probability one, ECO eventually evaluates at least one candidate solution in $\Omega_\varepsilon$.

Once such a candidate solution is evaluated, Lemma 1 ensures that the historical best fitness will not deteriorate afterwards. Hence, for any fixed $\varepsilon > 0$, with probability one, there exists a finite iteration index $K_\varepsilon$ such that

$$f(\boldsymbol{X}_{\text{best}}(k)) < f^* + \varepsilon, \qquad \forall k \geq K_\varepsilon. \tag{28}$$

Now take the countable sequence $\varepsilon_\ell = 1/\ell$, $\ell = 1, 2, \ldots$. The above result holds with probability one for every $\varepsilon_\ell$. Since the countable intersection of probability-one events still has probability one, we have, with probability one, that for every $\ell$ there exists a finite $K_\ell$ such that

$$f(\boldsymbol{X}_{\text{best}}(k)) < f^* + \frac{1}{\ell}, \qquad \forall k \geq K_\ell. \tag{29}$$

Together with $f(\boldsymbol{X}_{\text{best}}(k)) \geq f^*$, this implies

$$\lim_{k\to\infty} f(\boldsymbol{X}_{\text{best}}(k)) = f^* \tag{30}$$

with probability one. This completes the proof. □

**Remark 2.** *Theorem 1 establishes the almost-sure global convergence of ECO in terms of the historical best fitness. It does not require every population individual to converge to a global optimal solution. Instead, it proves that, under the stated assumptions, the fitness of the historical optimal population individual maintained by ECO converges to the global optimal value.*

**Remark 3.** *For a practical implementation with a finite maximum iteration number $k_{\max}$, Theorem 1 should be interpreted as an asymptotic result. It indicates that, when the evaluation budget is sufficiently large and the positive global reachability condition is satisfied, the probability that ECO attains a solution with an objective value arbitrarily close to the global optimal value approaches one.*

## 3.5. Computational complexity analysis

Computational complexity analysis is essential for evaluating the efficiency and scalability of optimization algorithms when solving problems of different scales and dimensions. It offers a theoretical perspective that reveals how the computational cost of an algorithm grows with the increase of algorithm parameters. For the ECO algorithm, the computational complexity is primarily related to the population size $N_{\text{pop}}$, the problem dimension $D$, and the maximum number of iterations $k_{\max}$. This section analyzes the time and space complexity of the ECO algorithm.

### *3.5.1. Time complexity*

Time complexity reflects the computational burden required to execute the algorithm. During the ECO process, the computational complexity of population initialization is $\mathcal{O}\left(N_{\text{pop}}D\right)$. In each iteration, the computational complexity of calculating the predation factor vector function $\boldsymbol{G}$ is $\mathcal{O}(D)$, and that of the roulette wheel selection is $\mathcal{O}\left(N_{\text{pop}}\right)$. Updating the population individuals has a computational complexity of $\mathcal{O}\left(N_{\text{pop}}D\right)$, while the complexity of function evaluation is $\mathcal{O}\left(N_{\text{pop}}F(D)\right)$, where $F(D)$ denotes the time complexity of the objective function in the optimization problem. Therefore, the overall dominant time complexity of the ECO algorithm is given as follows:

$$\mathcal{O}_{\text{time}}(\text{ECO}) \approx \mathcal{O}\left(N_{\text{pop}}D + k_{\max}\left[D + N_{\text{pop}} + N_{\text{pop}}D + N_{\text{pop}}F(D)\right]\right) \approx \mathcal{O}\left(k_{\max}N_{\text{pop}}\left[D + F(D)\right]\right). \tag{31}$$

If the objective function is computationally simple, the time complexity of ECO reduces to $\mathcal{O}(k_{\max}N_{\text{pop}}D)$. When the objective-function evaluation is expensive, namely when $F(D)$ is much larger than $D$, the total computational cost is dominated by the evaluation term, and the time complexity can be approximated as $\mathcal{O}(k_{\max}N_{\text{pop}}F(D))$. Therefore, ECO has relatively lightweight internal update rules, but its practical efficiency may still be limited by expensive objective-function evaluations, which is a common issue for population-based metaheuristic algorithms. Since these evaluations are independent across different population individuals, parallel implementation may help reduce the practical wall-clock time when computational resources are available.

### *3.5.2. Space complexity*

Space complexity measures the memory resources needed to store data structures and intermediate results during computation. ECO mainly requires storing several variables, including the predation factor vector function $\boldsymbol{G}$, which has a space complexity of $\mathcal{O}(D)$; population individuals with space complexity $\mathcal{O}\left(N_{\text{pop}}D\right)$; personal best positions and fitness values requiring $\mathcal{O}\left(N_{\text{pop}}D\right)$ and $\mathcal{O}\left(N_{\text{pop}}\right)$; and the global optimal solution $\boldsymbol{X}_{\text{best}}$ and its optimal fitness $f\left(\boldsymbol{X}_{\text{best}}\right)$, requiring $\mathcal{O}(D)$ and $\mathcal{O}(1)$, respectively. Therefore, the overall space complexity of the ECO algorithm is:

$$\mathcal{O}_{\text{space}}(\text{ECO}) \approx \mathcal{O}(D + N_{\text{pop}}D + N_{\text{pop}}D + N_{\text{pop}} + D + 1) \approx \mathcal{O}\left(N_{\text{pop}}D\right). \tag{32}$$

Since the dominant term is $\mathcal{O}\left(N_{\text{pop}}D\right)$, other terms can be neglected in asymptotic analysis.

## 4. Experimental overview

This section introduces the experimental design and the analytical methods used to evaluate ECO. Specifically, the benchmark test suites, parameter settings, performance metrics, and statistical comparison procedures are described to establish a fair and reproducible evaluation framework for assessing the effectiveness and robustness of ECO.

### 4.1. Experimental design

To conduct a fair and reproducible evaluation, standardized optimization functions are selected as benchmarks. Before the comparative experiments, the internal parameters of ECO are first examined through sensitivity analysis. Subsequently, classic benchmark functions and IEEE CEC test suites are adopted to evaluate the optimization performance of ECO. The benchmark test suites used in these experiments are summarized in Table 1, with the detailed definitions and evaluation rules provided in their corresponding technical reports. Accordingly, the experiments are organized into three components, each with a clearly defined objective and methodological framework:

(1) **Parameter sensitivity analysis**
First, the internal parameters of ECO are analyzed on 23 classic optimization functions. The population composition and decomposition probabilities are investigated separately, and the recommended default configuration is determined according to the ranking results.

(2) **Performance potential**
Second, the performance potential of ECO in solving optimization problems is evaluated. CEC-2014, CEC-2017, and CEC-2020 are selected as benchmark test suites. 29 SOTA MH algorithms are combined with ECO to form an algorithm pool containing 30 MH algorithms for comparative evaluation.

(3) **Engineering applications**
Finally, the applicability of ECO to real-world engineering problems is investigated. CEC-2020-RW test suite is adopted, which consists of constrained engineering problems derived from practical scenarios. Five representative problems and five widely used powerful MH algorithms are selected for comparison with ECO.

### 4.2. Performance metrics

This subsection introduces the evaluation criteria and statistical methods used to analyze the experimental results. Since each experiment is conducted independently and repeatedly, the results have to be viewed as a group of stochastic samples rather than a single deterministic value. Therefore, **statistical analysis** is firstly required to draw reliable conclusions [27]. In addition, **non-parametric testing** methods are commonly adopted for MH algorithms because the distributions of optimization results are generally unknown thus may not satisfy the assumptions required by parametric tests [28, 29]. This include the **pairwise comparison** between ECO and each competing MH algorithm on individual functions, and **multiple-comparison** among all algorithms across an entire test suite [27].

### *4.2.1. Statistical analysis*

For each test suite, every MH algorithm is independently executed $N$ times on each optimization function. The minimum (Min), average (Ave), and standard deviation (Std) values of the obtained best fitness values are recorded. In addition, box plots are generated to visually illustrate the distribution of the solution values. For minimization functions, a smaller Ave value usually indicates better performance. However, because MH algorithms are stochastic, such differences should be supported by statistical significance tests. Nonparametric tests are therefore introduced to determine whether the observed superiority is reliable rather than caused by random variation [28].

**Table 1**
Description of benchmark test suites.

| Experimental objective | Benchmarks |
|---|---|
| Parameter sensitivity analysis | 23 classic optimization functions[1] |
| Performance potential | CEC-2014[2] |
| | CEC-2017[3] |
| | CEC-2020[4] |
| Engineering applications | CEC-2020-RW[5] |

*Notes:* [1]https://ieeexplore.ieee.org/document/771163 (accessed: July 2026)
[2]https://github.com/P-N-Suganthan/CEC2014 (accessed: July 2026)
[3]https://github.com/P-N-Suganthan/CEC2017-BoundContrained (accessed: July 2026)
[4]https://github.com/P-N-Suganthan/2020-Bound-Constrained-Opt-Benchmark (accessed: July 2026)
[5]https://github.com/P-N-Suganthan/2020-RW-Constrained-Optimisation (accessed: July 2026)

**Table 2**
Similarities and difference of Wilcoxon tests.

| Content | Similarities |
|---|---|
| $H_0$ | There is no significant difference between the two groups of data<br>Both groups originate from populations with identical distributions |
| Output | $p$-value |
| **Content** | **Difference** |
| Input | Wilcoxon signed-rank test: paired samples from 2 groups<br>Wilcoxon rank-sum test: 2 groups of data |

#### *4.2.2. Pairwise comparison*

To compare ECO with each competing MH algorithm on individual optimization functions, the **Wilcoxon test** is adopted to compare ECO with each competing algorithm on individual functions. It includes the **Wilcoxon signed-rank test** for paired samples and the **Wilcoxon rank-sum test** for independent samples [30]. Both tests examine the null hypothesis $H_0$ that the two groups have no significant distributional difference, and their main output is the $p$-value. Their similarities and differences are summarized in Table 2.

Because the repeated results of two MH algorithms on the same function are not inherently paired, this study uses the **Wilcoxon rank-sum test**. The obtained $p$-value is compared with the significance level $\alpha$, and the Ave values are then used to determine the comparison result. If $p$-value $< \alpha$, $H_0$ is rejected. The algorithm with the smaller Ave is regarded as superior, denoted by "+", while the other isdenoted by "−". Otherwise, $H_0$ cannot be rejected, and the two algorithms are denoted by "=".

#### *4.2.3. Multiple comparisons*

To evaluate the overall performance across an entire test suite, this study also conducts multiple comparisons among all algorithms. Specifically, all algorithms are ranked on each function according to Ave. The ranks are then averaged over all functions to obtain the global ranking, following the core principle of the **Friedman mean rank** method.

Using Wilcoxon rank-sum tests for global ranking introduces three critical issues. First, for $m$ MH algorithms and $n$ optimization functions, $nm(m-1)/2$ pairwise tests are required, which increases the computational cost. Second, the "=" equivalence classification may lead to ranking indeterminacy. Third, from a statistical perspective, performing a large number of pairwise comparisons without correction may increase the probability of committing one or more Type I errors, which is known as the loss of control over the Family-Wise Error Rate (FWER) [28].

Therefore, the **Friedman test** [31] is adopted to address these limitations. The null hypothesis $H_0$ of the Friedman test is that no significant difference exists among the $m$ groups of data; in this study, this means that no significant difference exists among the performances of the $m$ MH algorithms. Similar to the Wilcoxon test, the Friedman test outputs a statistical significance $p$-value. If $p$-value $< \alpha$, $H_0$ is rejected, and the overall ranking can then be determined according to the ascending Friedman mean rank.

## 5. Parameter sensitivity analysis of ECO

This section investigates the sensitivity of ECO to its internal parameters and determines a suitable default configuration for the subsequent comparative experiments.

### 5.1. Internal parameters of ECO

The first group of parameters controls the population composition of ECO. $P_{\text{Pro}}$, $P_{\text{Her}}$, $P_{\text{Car}}$, and $P_{\text{Omn}}$ denote the proportions of producers, herbivores, carnivores, and omnivores, respectively. These proportions satisfy $P_{\text{Pro}} + P_{\text{Her}} + P_{\text{Car}} + P_{\text{Omn}} = 1$.

A higher producer proportion may strengthen the exploitation-oriented nutrient absorption process, whereas a higher consumer proportion may increase the diversity of predation-guided search trajectories.

The second group of parameters controls the decomposition mechanism. Two probability parameters, $p_{\text{opt}}$ and $p_{\text{loc}}$, are used to determine whether a decomposer performs optimal decomposition, local random decomposition, or global random decomposition. Specifically, $p_{opt}$ directly controls the probability of optimal decomposition, while $p_{loc}$ adjusts the relative preference for local random decomposition when random decomposition is selected.

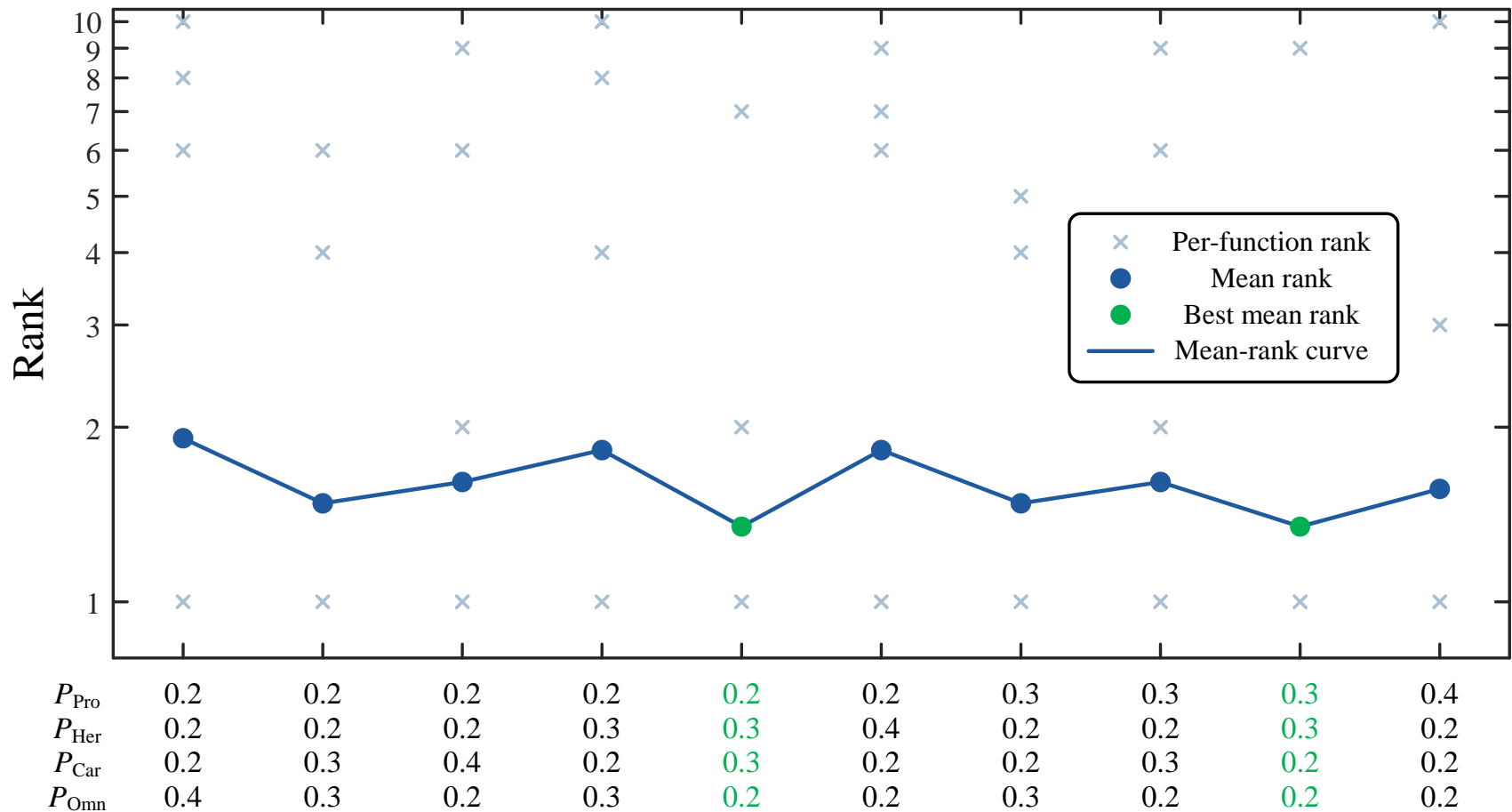

**Figure 5:** Population proportion sensitivity analysis of ECO on 23 classic optimization functions.

## 5.2. Experimental settings

On the 23 classic benchmark functions, the problem dimension is set to $D = 30$, and the termination criterion is MaxFEs $= 10,000 \times D$. For each parameter candidate, ECO is independently executed $N = 10$ times on each benchmark function. The final best fitness values obtained from these independent runs are used for statistical analysis.

To reduce the interaction between population composition and decomposition probabilities, the sensitivity analysis is divided into two stages. In the first stage, $p_{opt} = p_{loc} = 0.5$ is set, and the population composition is varied. In an ecosystem, producers, herbivores, carnivores, and omnivores interact with each other, and an excessively dominant or excessively weak component may disturb the balance between material cycling, predation-driven search, and decomposition. Therefore, each ecological component is constrained within $\{0.2, 0.3, 0.4\}$, so that the four proportions remain relatively balanced. Under this constraint, 10 feasible population configurations are generated as references to the proportion structure of the ECO ecosystem.

In the second stage, the best parameter setting found in the first stage will be adopted, and the probability parameters are investigated. Both $p_{opt}$ and $p_{loc}$ are selected from $\{0.3, 0.4, 0.5, 0.6, 0.7\}$, yielding 25 candidate probability configurations. This grid covers conservative, balanced, and random-decomposition-oriented settings without expanding the search space excessively.

For each benchmark function, all parameter candidates are ranked according to the Ave, Std and Min value of the 10 independent runs. After ranking all candidates on each of the 23 benchmark functions, the Friedman mean rank of each candidate is calculated as the overall sensitivity indicator.

## 5.3. Sensitivity results and parameter selection

This subsection interprets the sensitivity outcomes and identifies the final ECO parameter setting. In Figs. 5 and 6, the horizontal axis represents the candidate parameter configurations, while the vertical axis denotes the rank obtained on the 23 benchmark functions. The "×" markers indicate the ranks achieved by each parameter configuration on individual functions. Many markers overlap at the lowest rank because several configurations obtain tied first ranks on multiple functions, and the green marker highlights the configuration with the smallest mean rank. Therefore, the green-marked configurations are selected as the recommended ECO parameters.

The population sensitivity results are shown in Fig. 5. As can be observed, $P_{Pro} = 0.2$, $P_{Her} = 0.3$, $P_{Car} = 0.3$, and $P_{Omn} = 0.2$, as well as $P_{Pro} = 0.3$, $P_{Her} = 0.3$, $P_{Car} = 0.2$, and $P_{Omn} = 0.2$, achieve the best mean rank among the 10 population configurations. These results indicate that ECO benefits from a balanced ecological structure, as the different ecological roles are mutually coupled and collaboratively contribute to the dynamic search process. Considering both the overall optimization performance and the ecological interpretation, we recommend $P_{Pro} = 0.2$, $P_{Her} = 0.3$, $P_{Car} = 0.3$, and $P_{Omn} = 0.2$ as the default population proportions for ECO. This setting is adopted in all later optimization experiments.

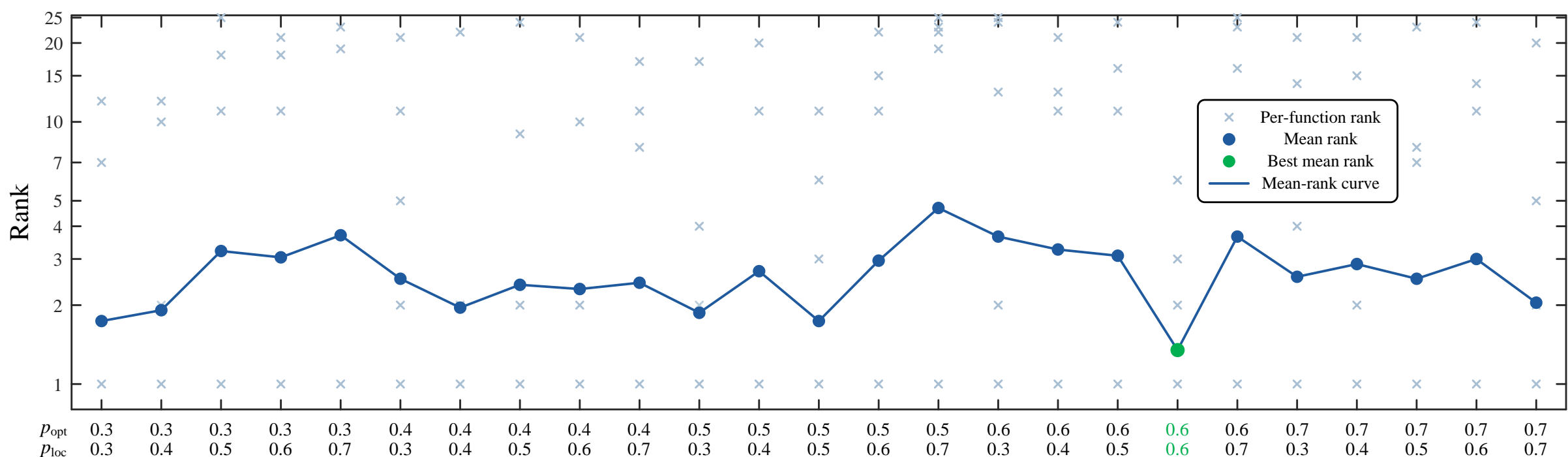

**Figure 6:** Decomposition probability sensitivity analysis of ECO on 23 classic optimization functions.

The probability sensitivity results are depicted in Fig. 6. According to the mean-rank curve, $p_{\text{opt}} = 0.6$ and $p_{\text{loc}} = 0.6$ provide the best overall performance among the 25 decomposition-probability configurations. Together with the population proportions $P_{\text{Pro}} = 0.2$, $P_{\text{Her}} = 0.3$, $P_{\text{Car}} = 0.3$, and $P_{\text{Omn}} = 0.2$, these parameters constitute the default internal parameter of ECO and remain unchanged in all subsequent optimization tasks. We also recommend using this configuration when applying ECO to other optimization problems, without problem-specific tuning.

## 6. Comparison of ECO with state-of-the-art metaheuristic algorithms

This section compares ECO with competing SOTA MH algorithms to further evaluate its performance potential under the selected default parameter configuration. First, a candidate pool comprising 30 SOTA MH algorithms, including ECO, is established, and comparative experiments are conducted on the CEC-2014 and CEC-2017 test suites. Subsequently, the top five competing algorithms are selected for further comparison with ECO on the CEC-2020 test suite, where solution accuracy, convergence speed, and qualitative algorithmic characteristics are analyzed.

### 6.1. Competing algorithms and parameter settings

The parameter configurations of the 30 MH algorithms in the candidate pool are summarized in Table 3. The candidate pool follows the order used in the numerical result tables and includes ECO together with 29 competing SOTA MH algorithms. To ensure that all algorithms in the candidate pool perform at their reported capacity, thereby enhancing the credibility of experimental results, the parameters of these algorithms are configured according to the settings reported in their original publications, with the exception of the population size, which is uniformly set to 30.

### 6.2. 30 state-of-the-art metaheuristic algorithms competition

This subsection evaluates ECO against the full 30-algorithm pool on CEC-2014 and CEC-2017. Following by the technical report of CEC-2014 and CEC-2017, the search termination criterion is set to MaxFEs $= 10,000 \times D$ and the number of independent runs $N = 51$. Solutions of MH algorithms including Min, Ave, and Std on every optimization function are recorded. $D = 10$ is adopted in this section, as this dimensionality aligns with practical optimization scenarios. MH algorithms are ranked 1 to 30 on each optimization function based on the following priority: smaller Ave, followed by Min and Std. Then, the Friedman test is adopted and Friedman mean rank across all optimization functions is calculated, and global performance rankings are generated accordingly.

After the independent repeated trials, the numerical results of all algorithms are shown in Table 14 and Table 15. Only Ave values are listed in these two tables due to the large volume of data. To identify the most competitive MH algorithm, the numerical results are first subjected to the Friedman test, yielding $p$-value $< \alpha$. Consequently, all the MH algorithms are globally ranked according to their Friedman mean ranks, with the results summarized in Table 4.

ECO is observed to achieve the first position among all MH algorithms on the CEC-2014 and CEC-2017 test suites. Besides ECO, the top five competitive MH algorithms are ARO [9], CFOA [17], CSA [36], WSO [10], and INFO [20]. These six algorithms are further analyzed comprehensively in the subsequent section.

### 6.3. Performance of 6 winning metaheuristic algorithms on CEC-2020

In this section, ECO is systematically compared with ARO, CFOA, CSA, WSO, and INFO on CEC-2020 test suite. Following by the technical report, the experimental parameters are set as follows: the search termination criterion

**Table 3**
Parameter settings of the 30 state-of-the-art metaheuristic algorithms in the comparison pool.

| MH | Parameters settings | Ref. | Year |
|---|---|---|---|
| ECO | Population proportions $P_{\text{Pro}} = 0.2$, $P_{\text{Her}} = 0.3$, $P_{\text{Car}} = 0.3$, and $P_{\text{Omn}} = 0.2$, decomposition probabilities $p_{\text{opt}} = 0.6$ and $p_{\text{loc}} = 0.6$ | This article | 2026 |
| AO | Exploitation adjustment parameters $\alpha = 0.1$ and $\delta = 0.1$ | [32] | 2021 |
| ARO | Hiding parameter $H$ decreases from 1 to $1/T$ ($T$ is the maximal number of iterations) | [9] | 2022 |
| AVOA | Measurement parameters $L_1 = 0.8$ and $L_2 = 0.2$, fixed parameter $w = 2.5$, strategy selection parameters $P_1 = 0.6$, $P_2 = 0.4$, and $P_3 = 0.6$ | [33] | 2021 |
| BFO | Distribution radius $R = 0$, random function $U(1) = 1$, initial step-jump parameter $J(1) = 1$ | [34] | 2024 |
| BKA | Attack behavior control parameter $p = 0.9$ | [35] | 2024 |
| CFOA | Move offset $r_3$ changes from either $-1$ or 1 toward 0, parameter $r_4 = 1, 2$, or 3 | [17] | 2024 |
| CSA | Constants $p_1 = 0.25$, $p_2 = 1.5$, $\rho = 1$, $c_1 = 1.75$, $c_2 = 1.75$, $\alpha = 3.5$, $\beta = 3$, and $\gamma = 1$, prey perception probability $P_p = 0.1$ | [36] | 2021 |
| CTCM | Tribes size $m = 20$, experience factor $c_1 = 2$, obey factor $c_2 = 1$, escape factor $c_3 = 0.1$ | [37] | 2025 |
| DBO | Deflection coefficient $k = 0.1$, natural coefficient $\alpha = -1$ or 1, probability value $\lambda = 0.1$, parameters $b = 0.3$ and $S = 0.5$ | [38] | 2023 |
| DMOA | The number of babysitters $bs = 3$, alpha female's vocalization $peep = 2$ | [39] | 2022 |
| ESOA | Step factors $step_a = 0.1$ and $step_b = 0.5$ | [40] | 2022 |
| FATA | Reflectance $\alpha = 0.2$, the first-half refraction ratio $Para_1$ changes from either $-2$ or 2 toward 0, the second-half refraction ratio $Para_2$ changes from 0 toward either $-150$ or 150 | [41] | 2024 |
| FLA | The number of water particles evaporate $N_e = 5$ | [42] | 2024 |
| GPC | Gravity $g = 9.8$, ramp angle $\theta = 10$, minimum friction $\mu_{k_\text{min}} = 1$, maximum friction $\mu_{k_\text{max}} = 10$, substitution probability = 0.5 | [43] | 2021 |
| HBA | Honey badger foraging ability parameter $\beta = 6$, constant $C = 2$ | [44] | 2022 |
| HGS | Hunger threshold $LH = 100$, parameter $l = 0.03$ | [45] | 2021 |
| IAO | Quality factor $\Phi$ increases from $-0.99$ to 0 | [18] | 2024 |
| INFO | Constants $c = 2$ and $d = 4$ | [20] | 2022 |
| PEOA | Constant $\beta = 1.5$, cluster size factor $\rho = 0.04$, the minimum possible population size $S_{\text{min}} = 5$, probability $p = 0.33$, memory size $H = 0.2$ | [46] | 2022 |
| PKO | Beating factor $BF = 8$, predatory efficiency of the pied kingfisher $PE_{\text{max}} = 0.5$ and $PE_{\text{min}} = 0$ | [47] | 2024 |
| RDCO | Strength coefficient $\alpha = 0.3$, base Hill coefficient $n_{\text{base}} = 2.0$, base dissociation constant $K_{\text{base}} = 0.8$ | [48] | 2026 |
| RIME | Step function parameter $w = 5$ | [13] | 2023 |
| RUN | Control parameters $a = 20$ and $b = 12$ | [21] | 2021 |
| SAO | Olfaction capacity $olf = 0.75$, step length constant $SL = 0.9$ | [14] | 2021 |
| SMA | Parameter $v_c$ changes from either $-1$ or 1 toward 0, parameter $z = 0.03$ | [49] | 2020 |
| SO | Constants $c_1 = 0.5$, $c_2 = 0.05$, and $c_3 = 2$ | [50] | 2022 |
| SSA | Safety threshold $ST = 0.8$ | [51] | 2020 |
| TJO | Control parameter $a$ increases from 0.5 to 1.5, $c$ decreases from 1.5 to 0.5 | [52] | 2026 |
| WSO | Maximum undulation frequency $f_{\text{max}} = 0.75$, minimum undulation frequency $f_{\text{max}} = 0.07$, acceleration coefficient $\tau = 4.125$, positive constants $a_0 = 6.25$, $a_1 = 100$, and $a_2 = 0.0005$ | [10] | 2022 |

Note: All parameters remain consistent with those used in the original publications.

**Table 4**
Friedman mean rank of ECO and competing metaheuristic algorithms on CEC-2014 and CEC-2017.

| MH | Friedman | | Mean | Rank | MH | Friedman | | Mean | Rank | MH | Friedman | | Mean | Rank |
|---|---|---|---|---|---|---|---|---|---|---|---|---|---|---|
| | CEC14 | CEC17 | | | | CEC14 | CEC17 | | | | CEC14 | CEC17 | | |
| **ECO** | **3** | **2.38** | **2.69** | **1** | IAO | 13.80 | 12.17 | 13 | 11 | BFO | 17.10 | 18.69 | 17.88 | 21 |
| **ARO** | **7.40** | **7.34** | **7.37** | **2** | SO | 12.83 | 14.31 | 13.56 | 12 | CTCM | 19.60 | 19.76 | 19.68 | 22 |
| **CFOA** | **8.90** | **6.55** | **7.75** | **3** | SAO | 14.53 | 14.55 | 14.54 | 13 | SSA | 17.97 | 22.07 | 19.98 | 23 |
| **CSA** | **10.70** | **7.72** | **9.24** | **4** | PKO | 16.10 | 13.28 | 14.71 | 14 | AVOA | 18.77 | 21.28 | 20 | 24 |
| **WSO** | **12.37** | **9.31** | **10.86** | **5** | HBA | 14.30 | 16.52 | 15.39 | 15 | ESOA | 21.27 | 20.76 | 21.02 | 25 |
| **INFO** | **10.73** | **13.28** | **11.98** | **6** | DMOA | 16.93 | 13.86 | 15.42 | 16 | FATA | 21.43 | 20.86 | 21.15 | 26 |
| RIME | 11.90 | 12.55 | 12.22 | 7 | RUN | 15.10 | 16.69 | 15.88 | 17 | BKA | 20.53 | 21.93 | 21.22 | 27 |
| TJO | 14.37 | 11.17 | 12.80 | 8 | PEOA | 16.10 | 15.90 | 16 | 18 | DBO | 21.80 | 21.14 | 21.47 | 28 |
| RDCO | 14.47 | 11.14 | 12.83 | 9 | HGS | 14.87 | 18.90 | 16.85 | 19 | GPC | 20.50 | 22.48 | 21.47 | 29 |
| SMA | 11.83 | 14.17 | 12.98 | 10 | FLA | 16.73 | 18.90 | 17.80 | 20 | AO | 23.53 | 24 | 23.76 | 30 |

MaxFEs $= 10,000 \times D$, where $D = 10/30/50/100$, and the number of independent runs $N = 30$. Solutions of MH algorithms including Min, Ave, and Std on every optimization functions will be recorded. A comparative analysis will be conducted to evaluate their solution accuracy and convergence rates. Additionally, a qualitative assessment of ECO will be performed to characterize its algorithmic behavior and optimization mechanisms.

### *6.3.1. Statistical analysis*

The average solution values of the six algorithms are shown in Table 16, with their Friedman mean ranks and global rankings provided in Table 5. ECO demonstrates statistically significant superiority across all tested dimensions and optimization functions, achieving a Friedman rank of 1 in all dimensional scenarios. Notably, its performance improvement becomes more pronounced as dimensionality increases. This behavior is consistent with the characteristics of CEC-2020. The suite contains shifted and rotated unimodal/basic functions as well as more difficult hybrid and composition functions, where nonseparability, multimodality, and heterogeneous sub-landscapes make it harder to maintain both global *exploration* and local *exploitation*.

**Table 5**
Friedman mean rank on CEC-2020.

| MH | ECO | ARO | CFOA | CSA | WSO | INFO |
|---|---|---|---|---|---|---|
| $D = 10$ | **1.7** | 3.6 | 3.7 | 3.9 | 3.3 | 4.8 |
| $D = 30$ | **1.5** | 4.5 | 3.7 | 3.4 | 4.6 | 3.3 |
| $D = 50$ | **1.2** | 4.0 | 3.3 | 3.3 | 5.1 | 4.1 |
| $D = 100$ | **1.1** | 4.1 | 3.2 | 3.8 | 5.0 | 3.8 |
| Mean / Rank | **1.4 / 1** | 4.1 / 5 | 3.5 / 2 | 3.6 / 3 | 4.5 / 6 | 4.0 / 4 |

**Table 6**
Wilcoxon rank-sum test results on CEC-2020.

| ECO vs. | ARO | CFOA | CSA | WSO | INFO |
|---|---|---|---|---|---|
| $D = 10$ | 5/4/1 | 8/1/1 | 7/3/0 | 4/4/2 | 8/1/1 |
| $D = 30$ | 10/0/0 | 8/2/0 | 7/3/0 | 9/1/0 | 6/1/3 |
| $D = 50$ | 10/0/0 | 7/3/0 | 7/2/1 | 9/1/0 | 7/3/0 |
| $D = 100$ | 10/0/0 | 9/1/0 | 9/0/1 | 9/1/0 | 9/1/0 |

ECO is well suited to such problems, as its performance advantage mainly arises from the cooperation among different ecological roles throughout the ecological cycle. Producer selection continuously preserves and intensifies promising low-fitness regions, consumer predation generates multiple movement patterns around promising and alternative regions at different ecological levels, and decomposition introduces optimal, local-random, and global-random search patterns. Through their complementary update mechanisms, these ecological roles jointly drive population evolution, maintain search diversity, and reinforce promising search directions, thereby reducing the risk of premature convergence on multimodal and rotated landscapes while retaining rapid intensification near high-quality regions. This coordinated search process provides a reasonable explanation for the high solution accuracy achieved by ECO on several CEC-2020 functions.

The results of the Wilcoxon rank-sum tests are shown in Table 6. ECO demonstrates a clear advantage in all dimensions. ARO, CFOA, WSO, and INFO can still defeat ECO on a few low-dimensional functions, which indicates that ECO is not always the fastest choice on relatively simple or less deceptive landscapes. On such problems, a more direct update rule may occasionally exploit the search space with less population renewal. However, as the dimension and landscape complexity increase, these competitors show weaker scalability, whereas ECO benefits from the balanced cooperation of population cycling, predation-based movement, and decomposition-based renewal. CSA remains competitive on several high-dimensional functions, but ECO shows stronger robustness across CEC-2020.

The dimensional scalability of ECO and the competing algorithms is shown in Fig. 15. The solution value distributions are depicted in Fig. 12, while the ranking variations across CEC-2020 are presented in Fig. 13. It can be observed that the vast majority of MH algorithms show considerable fluctuations, whereas ECO demonstrates excellent stability. Moreover, the distribution range of ECO is significantly narrower than those of the competing algorithms, with only a few outliers observed, further confirming the consistency and robustness of its optimization performance.

### *6.3.2. Convergence analysis*

The convergence speed and the trajectory of the best-found solutions during each function evaluation is shown in Fig. 14. It reveal three distinct optimization phases in ECO's performance. During the early exploration stage, ECO demonstrates significantly faster convergence than competing algorithms, indicating superior exploratory capabilities. This is followed by a transitional mid-phase where ECO's convergence rate exhibits a deliberate deceleration, allowing competing methods to temporarily gain performance advantages. However, in the final exploitation stage, ECO regains its rapid convergence characteristics, ultimately achieving optimal solutions across most test functions and dimensional configurations. This triphasic behavior, characterized by strong initial exploration, adaptive mid-stage refinement, and powerful final exploitation, enables ECO to consistently outperform alternative approaches while maintaining robust performance across diverse problem landscapes.

### *6.3.3. Qualitative analysis*

A comprehensive qualitative analysis of ECO's solution process is performed, as illustrated in Fig. 16. The visualization framework systematically presents four key analytical components for each test function: three-dimensional topological plots (with two-dimensional projections for clarity) depict the function landscape, followed by spatial distributions of searched regions using representative particle subsets. Solution trajectories are characterized through three complementary metrics: (1) the best-found solution curve; (2) population-wide average solution curve; and (3) dynamic *exploration-exploitation* balance profile. Finally, value trajectories of the optimal particle in the primary dimensions provide evolutionary signatures of convergence behavior.

The evaluation framework examined six representative functions covering four CEC-2020 categories, revealing three fundamental behavioral patterns. First, comprehensive global search capability is evidenced by particle distributions spanning the entire search space, with concentrated exploration near global optima regions—a manifestation of ECO's equilibrium between *exploration* and *exploitation*. Second, distinct phase transition characteristics emerge: initial iterations exhibit marked divergence between average and elite particle values (reflecting exploratory dominance), while terminal phases demonstrate unanimous value convergence across all 30 particles (confirming precise exploitation). Third, the quantitative *exploration-exploitation* balance metric provides empirical validation of this dynamic transition process, exhibiting characteristic sigmoidal progression from exploration-dominant to exploitation-focused search states.

#### 6.3.4. Running time comparison analysis

Finally, the computational efficiency of the six winning MH algorithms is examined to compare their processing speeds. As all algorithms are executed $N = 30$ times across all dimensions of each optimization function, the average computational runtime is presented in Table 17.

Analysis reveals that CSA and WSO exhibit the highest computational efficiency, with ECO performing comparably close behind. Conversely, INFO demonstrates the slowest processing speed, exhibiting an average runtime approximately triple that of the other algorithms. Furthermore, it is observed that the relative computational efficiency of all MH algorithms remains largely consistent across varying dimensions. However, a simple linear relationship between problem dimensionality and solving time is not identified for these algorithms. This characteristic is intrinsically linked to their unique operational mechanisms.

Overall, ECO demonstrates remarkably high solution accuracy without sacrificing much computational runtime, showcasing commendable overall performance.

## 7. Engineering validation of ECO

In this section, the performance of ECO in solving real-world engineering problems is verified. The engineering problems in the CEC-2020-RW test suite include complex objective functions and constraints, which is suitable benchmarks for evaluating engineering-oriented optimization capability. Five classic mechanical design problems are selected as benchmarks: (1) RC15: weight minimization of a speed reducer; (2) RC17: tension/compression spring design (case-1); (3) RC19: welded beam design, (4) RC20: three-bar truss design; and (5) RC31: gear train design.

According to the technical report, the experimental parameters are set as follows: the number of independent runs $N = 25$ and the search termination criterion are defined as:

$$\text{MaxFEs} = \begin{cases} 1 \times 10^5, & D \leq 10 \\ 2 \times 10^5, & 10 < D \leq 30 \\ 4 \times 10^5, & 30 < D \leq 50 \\ 8 \times 10^5, & 50 < D \leq 150 \\ 10^6, & 150 < D \end{cases} . \tag{33}$$

For each optimization function, the Min, Ave, and Std of MH algorithms, as well as the coordinates of the search particles corresponding to the optimization objectives of the engineering problems, will be recorded.

### 7.1. Competing algorithms and parameter settings

To enhance the persuasiveness of the experimental results, we select five SOTA MH algorithms that have been widely applied in engineering fields. They are FDB-AGDE [53], FDB-SFS [54], LRFDB-COA [55], L-SHADE [56], and NSM-SFS [57]. All parameters are configured according to the optimal settings reported in the respective source publications, as indicated in Table 7. Notably, both NSM-SFS and FDB-AGDE employ three distinct search mechanisms, denoted as case-1, case-2, and case-3. Empirical testing on the five engineering problems selected for this study determines that NSM-SFS (case-1) and FDB-AGDE (case-3) deliver superior performance.

### 7.2. Weight minimization of a speed reducer

The weight minimization of a speed reducer problem is a complex nonlinear optimization project in mechanical braking systems, aimed at minimizing the total weight of a speed reducer while adhering to constraints related to gear teeth bending stress, shaft stress, surface stress, and transverse shaft deflection. The structure of the speed reducer is

**Table 7**
Competing metaheuristic algorithms and parameter settings.

| MH | Parameters settings | Ref. | Year |
|---|---|---|---|
| FDB-AGDE | Population size $P = 50$, parameter $p = 0.1$ | [53] | 2021 |
| FDB-SFS | Start point $N_{\mathrm{P}} = 50$, maximum diffusion number $q = 1$ | [54] | 2021 |
| LRFDB-COA | Number of coyotes $= N_{\mathrm{P}} \times N_{\mathrm{c}}$, number of packs $N_{\mathrm{P}} = 20$, number of coyotes per pack $N_{\mathrm{c}} = 5$ | [55] | 2021 |
| L-SHADE | Initial population size $N^{\mathrm{init}} = 18 \times D$, parameters $r^{\mathrm{arc}} = 2.6$ and $p = 0.11$, historical memory size $H = 6$ | [56] | 2014 |
| NSM-SFS | Start point $N_{\mathrm{P}} = 50$, maximum diffusion number $q = 1$ | [57] | 2023 |

Note: All parameters remain consistent with those used in the original publications.

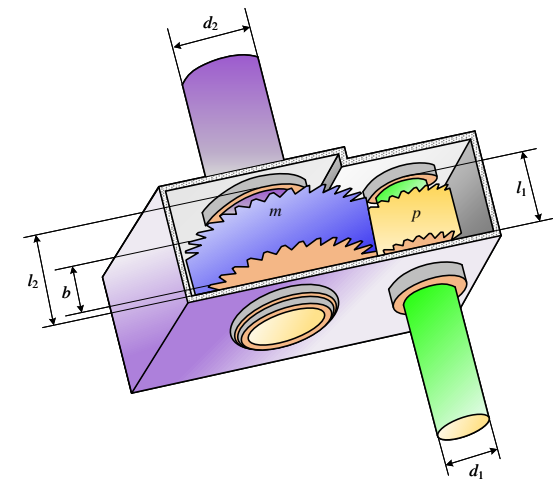

**Figure 7**
Weight minimization of a speed reducer.

**Table 8**
Numerical results on RC15 (speed reducer).

| | ECO | FDB-AGDE | FDB-SFS | LRFDB-COA | L-SHADE | NSM-SFS |
|---|---|---|---|---|---|---|
| $x_1$ | **3.50000000** | 3.50000000 | **3.50000000** | **3.50000000** | **3.50000000** | **3.50000000** |
| $x_2$ | **0.70000000** | 0.70000000 | **0.70000000** | **0.70000000** | **0.70000000** | **0.70000000** |
| $x_3$ | **17.0000000** | 17.0000000 | **17.0000000** | **17.0000000** | **17.0000000** | **17.0000000** |
| $x_4$ | **7.30000000** | 7.30000000 | **7.30000000** | **7.30000000** | **7.30000000** | **7.30000000** |
| $x_5$ | **7.71531991** | 7.71531991 | **7.71531991** | **7.71531991** | **7.71531991** | **7.71531991** |
| $x_6$ | **3.35054095** | 3.35054095 | **3.35054095** | **3.35054095** | **3.35054095** | **3.35054095** |
| $x_7$ | **5.28665446** | 5.28665446 | **5.28665446** | **5.28665446** | **5.28665446** | **5.28665446** |
| Min | **2994.42447** | 2994.42447 | **2994.42447** | **2994.42447** | **2994.42447** | **2994.42447** |
| Ave | **2994.42447** | 2994.42447 | **2994.42447** | **2994.42447** | **2994.42447** | **2994.42447** |
| Std | **9.2825E-13** | 1.0229E-08 | **9.2825E-13** | **9.2825E-13** | **9.2825E-13** | **9.2825E-13** |

illustrated in Fig. 7. This problem involves seven decision variables: surface width ($b = x_1$), gear module ($m = x_2$), pinion teeth count ($p = x_3$), lengths of the 1st and 2nd shafts between bearings ($l_1 = x_4,\ l_2 = x_5$), and diameters of the 1st and 2nd shafts ($d_1 = x_6,\ d_2 = x_7$), as follows:

$$\boldsymbol{X} = \left[x_1, x_2, x_3, x_4, x_5, x_6, x_7\right] = \left[b, m, p, l_1, l_2, d_1, d_2\right]. \tag{34}$$

The objective is to minimize:

$$f(\boldsymbol{X}) = 0.7854 x_1 x_2^2 \left(3.3333 x_3^2 + 14.9334 x_3 - 43.0934\right) - 1.508 x_1 \left(x_6^2 + x_7^2\right) + 7.4777 \left(x_6^3 + x_7^3\right) + 0.7854 \left(x_4 x_6^2 + x_5 x_7^2\right), \tag{35}$$

subject to:

$$\begin{aligned}
&g_1(\mathbf{X}) = \frac{27}{x_1 x_2^2 x_3} - 1 \le 0,\ g_2(\mathbf{X}) = \frac{397.5}{x_1 x_2^2 x_3^2} - 1 \le 0,\ g_3(\mathbf{X}) = \frac{1.93 x_4^3}{x_2 x_3 x_6^4} - 1 \le 0,\ g_4(\mathbf{X}) = \frac{1.93 x_5^3}{x_2 x_3 x_7^4} - 1 \le 0,\\
&g_5(\mathbf{X}) = \frac{x_2 x_3}{40} - 1 \le 0,\quad g_6(\mathbf{X}) = \frac{5 x_2}{x_1} - 1 \le 0,\quad g_7(\mathbf{X}) = \frac{x_1}{12 x_2} - 1 \le 0,\quad g_8(\mathbf{X}) = \frac{1.5 x_6 + 1.9}{x_4} - 1 \le 0,\\
&g_9(\mathbf{X}) = \frac{\sqrt{\left(\frac{745 x_5}{x_2 x_3}\right)^2 + 1.69 \times 10^7}}{110 x_6^3} - 1 \le 0,\quad g_{10}(\mathbf{X}) = \frac{\sqrt{\left(\frac{745 x_5}{x_2 x_3}\right)^2 + 1.57 \times 10^5}}{85 x_7^3} - 1 \le 0,\quad g_{11}(\mathbf{X}) = \frac{1.1 x_7 + 1.9}{x_5} - 1 \le 0.
\end{aligned} \tag{36}$$

Variable ranges are defined as follows:

$$2.6 \le x_1 \le 3.6,\quad 0.7 \le x_2 \le 0.8,\quad 17 \le x_3 \le 28,\quad 7.3 \le x_4 \le 8.3,\quad 7.3 \le x_5 \le 8.3,\quad 2.9 \le x_6 \le 3.9,\quad 5.0 \le x_7 \le 5.5. \tag{37}$$

As shown in Table 8, ECO, NSM-SFS, FDB-SFS, L-SHADE, and LRFDB-COA achieve the best performance on this problem. In contrast, FDB-AGDE obtains the same Min and Ave values but shows a larger Std, indicating weaker numerical stability. Comprehensive numerical validation confirms that all constraints are satisfied.

## 7.3. Tension/compression spring design

The tension/compression spring design problem aims to minimize the weight of a spring subject to several engineering constraints. The spring depicted in Fig. 8 must meet the requirements on shear stress, surge frequency, and deflection to ensure optimal performance. The design variables include the wire diameter ($d = x_1$), the mean coil diameter ($D = x_2$), and the number of active coils ($N = x_3$), which must be precisely estimated to achieve the optimal design configuration. Thus,

$$\boldsymbol{X} = \left[x_1, x_2, x_3\right] = \left[d, D, N\right]. \tag{38}$$

The objective is to minimize:

$$f(\boldsymbol{X}) = (x_3 + 2) x_2 x_1^2, \tag{39}$$

subject to:

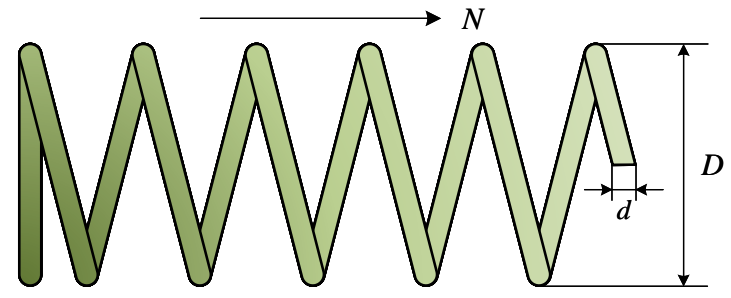

**Figure 8**
Tension/compression spring design.

**Table 9**
Numerical results on RC17 (spring design).

| MHS | **ECO** | FDB-AGDE | FDB-SFS | LRFDB-COA | L-SHADE | NSM-SFS |
|---|---|---|---|---|---|---|
| $x_1$ | **0.05169231** | 0.05168906 | 0.05168901 | 0.05172036 | 0.05168907 | 0.05168904 |
| $x_2$ | **0.35679602** | 0.35671777 | 0.35671643 | 0.35747109 | 0.35671784 | 0.35671715 |
| $x_3$ | **11.2843781** | 11.2889638 | 11.2890426 | 11.2449453 | 11.2889601 | 11.2890006 |
| Min | **0.01266523** | 0.01266523 | 0.01266523 | 0.01266524 | 0.01266523 | 0.01266523 |
| Ave | **0.01266525** | 0.01266527 | 0.01266533 | 0.01266538 | 0.01266950 | 0.01266525 |
| Std | **3.8839E-09** | 2.2047E-06 | 9.4667E-06 | 1.0027E-06 | 6.5781E-05 | 1.8822E-08 |

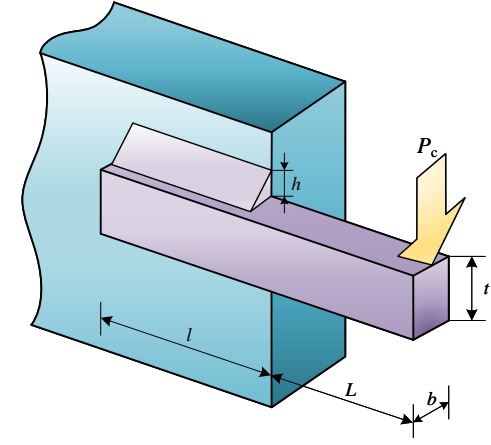

**Figure 9**
Welded beam design.

**Table 10**
Numerical results on RC19 (welded beam design).

| | **ECO** | **FDB-AGDE** | **FDB-SFS** | **LRFDB-COA** | **L-SHADE** | **NSM-SFS** |
|---|---|---|---|---|---|---|
| $x_1$ | **0.20572964** | **0.20572964** | **0.20572964** | **0.20572964** | **0.20572964** | **0.20572964** |
| $x_2$ | **3.25312004** | **3.25312004** | **3.25312004** | **3.25312004** | **3.25312004** | **3.25312004** |
| $x_3$ | **9.03662391** | **9.03662391** | **9.03662391** | **9.03662391** | **9.03662391** | **9.03662391** |
| $x_4$ | **0.20572964** | **0.20572964** | **0.20572964** | **0.20572964** | **0.20572964** | **0.20572964** |
| Min | **1.69524716** | **1.69524716** | **1.69524716** | **1.69524716** | **1.69524716** | **1.69524716** |
| Ave | **1.69524716** | **1.69524716** | **1.69524716** | **1.69524716** | **1.69524716** | **1.69524716** |
| Std | **2.2662E-16** | **2.2662E-16** | **2.2662E-16** | **2.2662E-16** | **2.2662E-16** | **2.2662E-16** |

$$
\begin{aligned}
& g_1(\boldsymbol{X}) = 1 - \frac{x_3^3}{71785x_1^4} \leq 0, \quad g_2(\boldsymbol{X}) = \frac{4x_2^2 - x_1x_2}{12566(x_2x_1^3 - x_1^4)} + \frac{1}{5108x_1^2} \leq 0, \\
& g_3(\boldsymbol{X}) = 1 - \frac{140.45x_1}{x_2x_3^2} \leq 0, \quad g_4(\boldsymbol{X}) = \frac{x_1 + x_2}{1.5} - 1 \leq 0.
\end{aligned} \tag{40}
$$

Variable ranges are defined as follows:

$$
0.05 \leq x_1 \leq 2.00, \quad 0.25 \leq x_2 \leq 1.30, \quad 2.00 \leq x_3 \leq 15.0. \tag{41}
$$

As presented in Table 9, ECO achieves the best overall performance among the evaluated algorithms according to Ave, Min, and Std. NSM-SFS is highly competitive, while FDB-SFS, FDB-AGDE, L-SHADE, and LRFDB-COA exhibit larger Ave or Std values. Comprehensive numerical validation confirms that all constraints are satisfied.

## 7.4. Welded beam design

The welded beam design problem, as illustrated in Fig. 9, aims to minimize the total manufacturing cost of a welded beam by optimizing four design variables: height ($h = x_1$), length ($l = x_2$), thickness ($t = x_3$), and width ($b = x_4$), subject to seven constraints. The decision variable vector is:

$$
\boldsymbol{X} = \left[x_1, x_2, x_3, x_4\right] = \left[h, l, t, b\right]. \tag{42}
$$

The objective is to minimize:

$$
f(\boldsymbol{X}) = 1.1047x_1x_2 + 0.04811x_3x_4(14.0 + x_2), \tag{43}
$$

subject to:

$$
\begin{aligned}
& g_1(\boldsymbol{X}) = \tau(\boldsymbol{X}) - \tau_{\max} \leq 0, \ g_2(\boldsymbol{X}) = \sigma(\boldsymbol{X}) - \sigma_{\max} \leq 0, \ g_3(\boldsymbol{X}) = \delta(\boldsymbol{X}) - \delta_{\max} \leq 0, \ g_4(\boldsymbol{X}) = x_1 - x_4 \leq 0, \\
& g_5(\boldsymbol{X}) = P - P_c(\boldsymbol{X}) \leq 0, \quad g_6(\boldsymbol{X}) = 0.125 - x_1 \leq 0, \quad g_7(\boldsymbol{X}) = 1.1047x_1x_2 + 0.04811x_3x_4(14.0 + x_2) - 5.0 \leq 0.
\end{aligned} \tag{44}
$$

where

$$
\tau(\boldsymbol{X}) = \sqrt{(t')^2 + 2t't''\frac{x_2}{2R} + (t'')^2}, \quad t' = \frac{P}{\sqrt{2x_1x_2}}, \quad t'' = \frac{MR}{J}, \quad M = P\left(L + \frac{x_2}{2}\right), \quad R = \sqrt{\frac{x_2^2}{4} + \left(\frac{x_1 + x_3}{2}\right)^2},
$$

$$
J = 2\left(\sqrt{2x_1x_2}\left[\frac{x_2^2}{4} + \left(\frac{x_1 + x_3}{2}\right)^2\right]\right), \quad \sigma(\boldsymbol{X}) = \frac{6PL}{x_4x_3^2}, \quad \delta(\boldsymbol{X}) = \frac{6PL^3}{Ex_3x_4}, \quad P_c(\boldsymbol{X}) = \frac{4.0134E\sqrt{\frac{x_3^2x_4^2}{36}}}{L^2}\left(1 - \frac{x_3}{2L}\sqrt{\frac{E}{4G}}\right),
$$

$$
P = 6000, \quad L = 14, \quad E = 30 \times 10^6, \quad G = 12 \times 10^6, \quad \tau_{\max} = 13600, \quad \sigma_{\max} = 30000, \quad \delta_{\max} = 0.25.
$$

Variable ranges are defined as follows:

$$
0.1 \leq x_3, x_2 \leq 10, \quad 0.1 \leq x_4 \leq 2, \quad 0.125 \leq x_1 \leq 2. \tag{45}
$$

The optimization results for the welded beam design problem across all MHS algorithms are presented in Table 10. All algorithms yield identical results, with the solutions satisfying the design constraints and demonstrating robust performance. Comprehensive numerical validation confirms these findings.

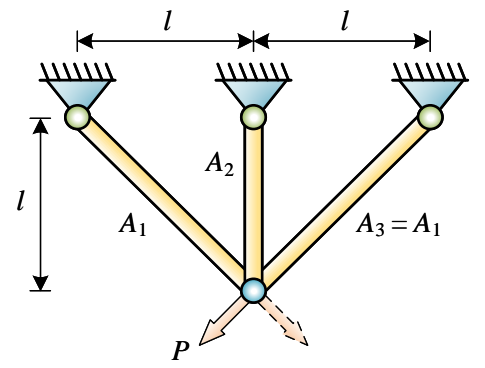

**Figure 10**
Three bar truss design.

**Table 11**
Numerical results on RC20 (three-bar truss design).

| | **ECO** | **FDB-AGDE** | **FDB-SFS** | LRFDB-COA | **L-SHADE** | **NSM-SFS** |
|---|---|---|---|---|---|---|
| $x_1$ | **0.78867513** | **0.78867514** | **0.78867514** | 0.78867507 | **0.78867513** | **0.78867513** |
| $x_2$ | **0.40824830** | **0.40824829** | **0.40824829** | 0.40824848 | **0.40824829** | **0.40824830** |
| Min | **263.895843** | **263.895843** | **263.895843** | 263.895843 | **263.895843** | **263.895843** |
| Ave | **263.895843** | **263.895843** | **263.895843** | 263.895843 | **263.895843** | **263.895843** |
| Std | **5.8016E-14** | **5.8016E-14** | **5.8016E-14** | 3.1422E-11 | **5.8016E-14** | **5.8016E-14** |

**Figure 11**
Gear train design.

**Table 12**
Numerical results on RC31 (gear train design).

| | **ECO** | FDB-AGDE | FDB-SFS | LRFDB-COA | L-SHADE | NSM-SFS |
|---|---|---|---|---|---|---|
| $x_1$ | **49.3000403** | 49.4918951 | 49.3824442 | 48.6638793 | 49.3698757 | 49.2889111 |
| $x_2$ | **19.3605917** | 16.0983856 | 15.7965736 | 18.5973568 | 16.0521244 | 16.0674738 |
| $x_3$ | **15.8481360** | 19.3860636 | 19.3517569 | 16.2893830 | 19.2014224 | 18.9821250 |
| $x_4$ | **42.8673784** | 42.6301489 | 43.4353539 | 43.2077863 | 42.5424910 | 43.3920358 |
| Min | **2.7009E-12** | 2.7010E-12 | 2.7017E-12 | 2.7009E-12 | 2.7015E-12 | 2.7010E-12 |
| Ave | **2.7009E-12** | 1.8669E-11 | 3.7588E-10 | 1.4927E-11 | 2.0184E-11 | 1.5408E-11 |
| Std | **0.0000E+00** | 2.1711E-10 | 4.7952E-10 | 1.0189E-11 | 2.4235E-10 | 1.0324E-11 |

## 7.5. Three-bar truss design

The three-bar truss design problem aims to determine the optimal values of three variables, which include the areas of three bars ($A_1 = A_3$, $A_2$), and minimize the weight of the truss while meeting specific constraints. The general structure of the truss is depicted in Fig. 10. The design variables are defined as follows:

$$\boldsymbol{X} = \left[x_1, x_2\right] = \left[A_1, A_2\right]. \tag{46}$$

The objective is to minimize:

$$f(\boldsymbol{X}) = \left(2\sqrt{2}x_1 + x_2\right) l, \tag{47}$$

subject to:

$$g_1(\boldsymbol{X}) = \frac{\sqrt{2}x_1 + x_2}{\sqrt{2}x_1^2 + 2x_1x_2} P - \sigma \leq 0, \quad g_2(\boldsymbol{X}) = \frac{x_2}{\sqrt{2}x_1^2 + 2x_1x_2} P - \sigma \leq 0, \quad g_3(\boldsymbol{X}) = \frac{1}{\sqrt{2}x_2 + x_1} P - \sigma \leq 0. \tag{48}$$

where

$$l = 100\,\text{cm}, \quad P = 2\,\text{kN/cm}^2, \quad \sigma = 2\,\text{kN/cm}^2.$$

The variable ranges are defined as follows:

$$0 \leq x_1, x_2 \leq 1. \tag{49}$$

As shown in Table 11, ECO, NSM-SFS, FDB-SFS, FDB-AGDE, and L-SHADE achieve the best performance for this problem. Although LRFDB-COA reaches the same Min and Ave values, its Std is larger, indicating lower stability across independent runs. Comprehensive numerical validation confirms that all constraints are satisfied.

## 7.6. Gear train design

The gear train design problem aims to minimize the cost associated with the gear ratio of the gear train depicted in Fig. 11. The problem involves four integer variables: $T_a$, $T_b$, $T_d$, and $T_f$, which represent the number of teeth on four different gearwheels. The gear ratio is defined as $T_bT_d/(T_aT_f)$. The design variables are defined as follows:

$$\boldsymbol{X} = \left[x_1, x_2, x_3, x_4\right] = \left[T_a, T_b, T_d, T_f\right]. \tag{50}$$

The objective is to minimize:

$$f(\boldsymbol{X}) = \left(\frac{1}{6.931} - \frac{T_bT_d}{T_aT_f}\right)^2, \tag{51}$$

subject to:

$$g_i(x_i) = 12 - x_i \leq 0, \quad i = 1, 2, 3, 4, \quad g_i(x_i) = 60 - x_i \leq 0, \quad i = 5, 6, 7, 8. \tag{52}$$

Variable ranges are defined as follows:

$$0.01 \leq x_1, x_2, x_3, x_4 \leq 60. \tag{53}$$

**Table 13**
Friedman mean rank of ECO and competing metaheuristic algorithms on CEC-2020-RW benchmark.

| MH | **ECO** | FDB-AGDE | FDB-SFS | LRFDB-COA | L-SHADE | NSM-SFS |
|---|---|---|---|---|---|---|
| RC15 | 1 | 6 | 1 | 1 | 1 | 1 |
| RC17 | 1 | 3 | 4 | 5 | 6 | 2 |
| RC19 | 1 | 1 | 1 | 1 | 1 | 1 |
| RC20 | 1 | 1 | 1 | 6 | 1 | 1 |
| RC31 | 1 | 4 | 6 | 2 | 5 | 3 |
| Mean / Rank | **1.0 / 1** | 3.0 / 5 | 2.6 / 3 | 3.0 / 5 | 2.8 / 4 | 1.6 / 2 |

As presented in Table 12, ECO achieves the best performance on this problem, obtaining the smallest Ave and Std. LRFDB-COA and NSM-SFS follow ECO, while FDB-SFS, FDB-AGDE, and L-SHADE exhibit larger Ave or Std values. Comprehensive numerical validation confirms that all constraints are satisfied.

### 7.7. Rank of all metaheuristic algorithms on CEC-2020-RW

As shown in Table 13, ECO achieves the best overall performance among the evaluated algorithms on the five engineering problems. NSM-SFS ranks second, followed by FDB-SFS and L-SHADE. FDB-AGDE and LRFDB-COA obtain the same final rank because their Friedman mean ranks are identical.

The ranking results indicate that ECO remains highly competitive against engineering-oriented algorithms with established parameter settings. Its comparatively simple parameterization and strong performance across the five selected engineering problems suggest good generality for solving constrained real-world optimization tasks.

## 8. Conclusion

This article proposes a novel MH algorithm called ECO to tackle complex, non-convex constrained optimization problems characterized by numerous local optima. Inspired by the processes of energy flow and material cycling within ecosystems, ECO uses a mathematical model to simulate the ecological cycle, incorporating tailored update strategies for the producers, consumers, and decomposers. Specifically, consumer update strategies involve targeted predation mechanisms for herbivores, carnivores, and omnivores, while decomposer strategies comprise optimal, local random, and global random decomposition mechanisms.

To comprehensively evaluate the optimization performance of the ECO, parameter sensitivity analysis and two types of comparative experiments are conducted. First, the internal ecological parameters of ECO are investigated on 23 classical benchmark functions. The sensitivity analysis identifies a robust default configuration, where the recommended population proportions are $P_{\text{Pro}} = 0.2$, $P_{\text{Her}} = 0.3$, $P_{\text{Car}} = 0.3$, and $P_{\text{Omn}} = 0.2$, and the decomposition probabilities are $p_{\text{opt}} = 0.6$ and $p_{\text{loc}} = 0.6$. We recommend keeping it fixed when applying ECO without problem-specific parameter tuning.

For the first comparative evaluation, an algorithm pool of 30 competitive MH algorithms is established, and ECO is compared with the competing algorithms on the IEEE CEC-2014 and CEC-2017 test suites. Based on the experimental results, six top-performing algorithms, namely ECO, ARO, CFOA, CSA, WSO, and INFO, are selected for further evaluation on the IEEE CEC-2020 test suite. The numerical results obtained by all MH algorithms are analyzed, and both the Wilcoxon rank-sum test and the Friedman test are conducted. The results show that ECO achieves the best overall optimization performance among them. This algorithm-pool experiment confirms ECO's superiority over a wide range of advanced algorithms introduced in recent years.

For the second comparative evaluation, comprehensive comparative simulations are conducted to validate ECO's applicability for solving complex real-world constrained optimization problems. The benchmark algorithms include five improved SOTA MH algorithms, comprising NSM-SFS, FDB-SFS, FDB-AGDE, L-SHADE, and LRFDB-COA. These algorithms are evaluated on five engineering problems from the IEEE CEC-2020-RW test suite. All numerical results undergo rigorous analysis, complemented by the Friedman statistical test for performance ranking. The results indicate that ECO achieves competitive performance against these SOTA algorithms, thereby demonstrating its strong potential for tackling complex real-world scenarios.

In conclusion, ECO demonstrates rapid convergence, high-precision solutions, and a strong ability to escape local optima, outperforming the vast majority of existing SOTA MH algorithms in terms of optimization performance. Its simple ecological structure, sensitivity-analysis-based default configuration, and superior capability in solving complex optimization problems make it a highly effective tool. Future research will focus on extending ECO to multi-objective optimization tasks and integrating it with more advanced general-purpose constraint-handling techniques for complex constrained optimization problems, thereby further broadening its applicability.

## Code and project resources

The official project page of ECO is available at https://jxxsteven7.github.io/ECO-Optimizer/.
The source code and detailed instructions are available at https://github.com/jxxsteven7/ECO-Optimizer.
The MATLAB implementation is also available at https://www.mathworks.com/matlabcentral/fileexchange/180852-ecological-cycle-optimizer-eco.

## CRediT authorship contribution statement

**Boyu Ma:** Conceptualization, Methodology, Software, Validation, Formal analysis, Investigation, Resources, Data curation, Writing - original draft, Writing - review & editing, Visualization, Supervision, Project administration. **Jiaxiao Shi:** Methodology, Software, Validation, Formal analysis, Investigation, Resources, Data curation, Writing - original draft, Writing - review & editing, Visualization. **Yiming Ji:** Formal analysis, Writing - review & editing, Visualization. **Zhengpu Wang:** Writing - review & editing, Visualization.

# Appendix A. Experimental results on CEC-2014 and CEC-2017

**Table 14**

Numerical results of ECO and competing metaheuristic algorithms on CEC-2014.

| MH | *F*1 | *F*2 | *F*3 | *F*4 | *F*5 | *F*6 | *F*7 | *F*8 | *F*9 | *F*10 | *F*11 | *F*12 | *F*13 | *F*14 | *F*15 | *F*16 | *F*17 | *F*18 | *F*19 | *F*20 | *F*21 | *F*22 | *F*23 | *F*24 | *F*25 | *F*26 | *F*27 | *F*28 | *F*29 | *F*30 |
|---|---|---|---|---|---|---|---|---|---|---|---|---|---|---|---|---|---|---|---|---|---|---|---|---|---|---|---|---|---|---|
| **ECO** | 1.262E+03 | 2.208E+02 | 3.035E+02 | 4.035E+02 | **5.183E+02** | **6.003E+02** | 7.001E+02 | 8.048E+02 | 9.110E+02 | **1.007E+03** | 1.426E+03 | 1.200E+03 | **1.300E+03** | **1.400E+03** | **1.501E+03** | 1.602E+03 | 1.895E+03 | 1.848E+03 | **1.901E+03** | 2.029E+03 | 2.131E+03 | **2.209E+03** | **2.493E+03** | 2.515E+03 | **2.633E+03** | **2.700E+03** | **2.702E+03** | **3.000E+03** | 3.175E+03 | 3.775E+03 |
| ARO | 2.653E+03 | 2.748E+02 | 3.001E+02 | 4.202E+02 | 5.200E+02 | 6.015E+02 | 7.001E+02 | 8.022E+02 | 9.159E+02 | 1.066E+03 | 1.495E+03 | 1.200E+03 | 1.300E+03 | 1.400E+03 | 1.501E+03 | 1.602E+03 | 2.733E+03 | 1.951E+03 | 1.901E+03 | **2.007E+03** | 2.128E+03 | 2.235E+03 | 2.500E+03 | 2.541E+03 | 2.692E+03 | 2.700E+03 | 2.827E+03 | 3.039E+03 | 3.163E+03 | 3.806E+03 |
| CFOA | 9.765E+03 | 1.098E+03 | 8.624E+02 | 4.249E+02 | 5.196E+02 | 6.010E+02 | 7.001E+02 | 8.078E+02 | **9.079E+02** | 1.346E+03 | 1.426E+03 | 1.201E+03 | 1.300E+03 | 1.400E+03 | 1.501E+03 | **1.602E+03** | 2.198E+03 | 2.017E+03 | 1.902E+03 | 2.063E+03 | 2.391E+03 | 2.226E+03 | 2.629E+03 | **2.515E+03** | 2.664E+03 | 2.700E+03 | 2.899E+03 | 3.177E+03 | 3.232E+03 | 3.640E+03 |
| CSA | 3.403E+04 | 7.322E+02 | 7.019E+02 | 4.215E+02 | 5.201E+02 | 6.029E+02 | 7.001E+02 | 8.108E+02 | 9.101E+02 | 1.201E+03 | 1.585E+03 | 1.200E+03 | 1.300E+03 | 1.400E+03 | 1.501E+03 | 1.602E+03 | 2.079E+03 | 1.903E+03 | 1.903E+03 | 2.053E+03 | 2.289E+03 | 2.220E+03 | 2.629E+03 | 2.528E+03 | 2.668E+03 | 2.700E+03 | 2.999E+03 | 3.217E+03 | 2.438E+05 | 3.599E+03 |
| WSO | 8.738E+03 | 6.459E+02 | 3.944E+02 | 4.148E+02 | 5.198E+02 | 6.021E+02 | 7.005E+02 | 8.052E+02 | 9.109E+02 | 1.192E+03 | **1.409E+03** | 1.201E+03 | 1.300E+03 | 1.400E+03 | 1.501E+03 | 1.602E+03 | 1.963E+03 | 1.820E+03 | 1.901E+03 | 2.009E+03 | 2.164E+03 | 2.247E+03 | 2.629E+03 | 2.532E+03 | 2.683E+03 | 2.703E+03 | 3.009E+03 | 3.270E+03 | 1.646E+05 | 3.795E+03 |
| INFO | **1.000E+02** | **2.000E+02** | **3.000E+02** | 4.226E+02 | 5.200E+02 | 6.035E+02 | 7.001E+02 | 8.118E+02 | 9.263E+02 | 1.175E+03 | 1.801E+03 | 1.200E+03 | 1.300E+03 | 1.400E+03 | 1.501E+03 | 1.603E+03 | 2.207E+03 | 1.835E+03 | 1.902E+03 | 2.030E+03 | 2.386E+03 | 2.266E+03 | 2.500E+03 | 2.546E+03 | 2.690E+03 | 2.700E+03 | 2.900E+03 | **3.000E+03** | 1.181E+05 | 3.857E+03 |
| RIME | 9.650E+04 | 3.725E+03 | 3.413E+02 | 4.181E+02 | 5.200E+02 | 6.023E+02 | 7.001E+02 | **8.000E+02** | 9.117E+02 | 1.071E+03 | 1.461E+03 | 1.200E+03 | 1.300E+03 | 1.400E+03 | 1.501E+03 | 1.602E+03 | 6.327E+03 | 1.396E+04 | 1.901E+03 | 2.125E+03 | 2.365E+03 | 2.252E+03 | 2.629E+03 | 2.525E+03 | 2.679E+03 | 2.700E+03 | 2.983E+03 | 3.225E+03 | 6.887E+04 | 3.893E+03 |
| TJO | 1.502E+05 | 1.416E+03 | 1.458E+03 | 4.272E+02 | 5.203E+02 | 6.013E+02 | 7.002E+02 | 8.155E+02 | 9.161E+02 | 1.539E+03 | 1.789E+03 | 1.201E+03 | 1.300E+03 | 1.400E+03 | 1.501E+03 | 1.603E+03 | 2.245E+03 | 1.912E+03 | 1.902E+03 | 2.087E+03 | 2.406E+03 | 2.251E+03 | 2.629E+03 | 2.520E+03 | 2.664E+03 | 2.700E+03 | 2.934E+03 | 3.179E+03 | 3.363E+03 | 3.788E+03 |
| RDCO | 7.391E+04 | 1.210E+03 | 1.270E+03 | 4.354E+02 | 5.204E+02 | 6.006E+02 | **7.001E+02** | 8.149E+02 | 9.170E+02 | 2.141E+03 | 2.229E+03 | 1.201E+03 | 1.300E+03 | 1.400E+03 | 1.502E+03 | 1.602E+03 | 3.235E+03 | 6.479E+03 | 1.903E+03 | 2.117E+03 | 3.334E+03 | 2.233E+03 | 2.629E+03 | 2.517E+03 | 2.645E+03 | 2.700E+03 | 2.702E+03 | 3.268E+03 | 3.341E+03 | 4.413E+03 |
| SMA | 1.121E+05 | 7.136E+03 | 8.527E+02 | 4.209E+02 | 5.201E+02 | 6.040E+02 | 7.003E+02 | 8.005E+02 | 9.127E+02 | 1.164E+03 | 1.631E+03 | 1.200E+03 | 1.300E+03 | 1.400E+03 | 1.501E+03 | 1.602E+03 | 9.632E+03 | 1.753E+04 | 1.902E+03 | 3.164E+03 | 3.036E+03 | 2.223E+03 | 2.500E+03 | 2.530E+03 | 2.683E+03 | 2.700E+03 | 2.887E+03 | **3.000E+03** | 3.535E+03 | 3.651E+03 |
| IAO | 1.904E+04 | 6.268E+07 | 1.090E+03 | 4.135E+02 | 5.202E+02 | 6.037E+02 | 7.044E+02 | 8.203E+02 | 9.211E+02 | 1.470E+03 | 1.689E+03 | 1.200E+03 | 1.300E+03 | 1.400E+03 | 1.508E+03 | 1.603E+03 | **1.846E+03** | **1.813E+03** | 1.902E+03 | 2.019E+03 | **2.119E+03** | 2.225E+03 | 2.500E+03 | 2.592E+03 | 2.698E+03 | 2.700E+03 | 2.829E+03 | **3.000E+03** | 3.135E+03 | 3.610E+03 |
| SO | 2.137E+04 | 1.047E+03 | 9.775E+02 | 4.192E+02 | 5.203E+02 | 6.034E+02 | 7.002E+02 | 8.125E+02 | 9.180E+02 | 1.259E+03 | 1.722E+03 | **1.200E+03** | 1.300E+03 | 1.400E+03 | 1.501E+03 | 1.603E+03 | 4.179E+03 | 3.478E+03 | 1.902E+03 | 2.145E+03 | 2.384E+03 | 2.256E+03 | 2.500E+03 | 2.542E+03 | 2.692E+03 | 2.700E+03 | 2.877E+03 | 3.000E+03 | 5.910E+04 | 3.880E+03 |
| SAO | 2.430E+04 | 4.225E+03 | 2.723E+03 | 4.252E+02 | 5.195E+02 | 6.012E+02 | 7.001E+02 | 8.128E+02 | 9.139E+02 | 1.293E+03 | 1.592E+03 | 1.200E+03 | 1.300E+03 | 1.400E+03 | 1.501E+03 | 1.603E+03 | 5.346E+03 | 1.069E+04 | 1.902E+03 | 7.338E+03 | 7.833E+03 | 2.258E+03 | 2.629E+03 | 2.527E+03 | 2.694E+03 | 2.700E+03 | 3.038E+03 | 3.225E+03 | 4.557E+05 | 3.795E+03 |
| PKO | 1.854E+05 | 7.077E+03 | 2.564E+03 | 4.348E+02 | 5.201E+02 | 6.037E+02 | 7.001E+02 | 8.127E+02 | 9.178E+02 | 1.270E+03 | 1.889E+03 | 1.200E+03 | 1.300E+03 | 1.400E+03 | 1.501E+03 | 1.603E+03 | 1.733E+04 | 5.263E+03 | 1.901E+03 | 3.557E+03 | 2.852E+03 | 2.223E+03 | 2.629E+03 | 2.522E+03 | 2.702E+03 | 2.700E+03 | 3.070E+03 | 3.179E+03 | 8.993E+04 | 3.550E+03 |
| HBA | 3.947E+06 | 6.407E+02 | 3.003E+02 | 4.184E+02 | 5.202E+02 | 6.035E+02 | 7.001E+02 | 8.110E+02 | 9.205E+02 | 1.395E+03 | 1.778E+03 | 1.200E+03 | 1.300E+03 | 1.400E+03 | 1.501E+03 | 1.603E+03 | 2.792E+03 | 3.049E+03 | 1.903E+03 | 2.093E+03 | 2.715E+03 | 2.265E+03 | 2.500E+03 | 2.539E+03 | 2.686E+03 | 2.700E+03 | 2.915E+03 | 3.183E+03 | 1.334E+06 | 5.257E+03 |
| DMOA | 5.713E+05 | 5.011E+03 | 3.876E+03 | **4.032E+02** | 5.203E+02 | 6.058E+02 | 7.001E+02 | 8.044E+02 | 9.182E+02 | 1.846E+03 | 2.627E+03 | 1.201E+03 | 1.300E+03 | 1.400E+03 | 1.502E+03 | 1.603E+03 | 8.149E+03 | 6.970E+03 | 1.902E+03 | 3.259E+03 | 3.129E+03 | 2.225E+03 | 2.629E+03 | 2.522E+03 | 2.686E+03 | 2.700E+03 | 3.123E+03 | 3.106E+03 | **3.107E+03** | **3.296E+03** |
| RUN | 6.749E+03 | 1.684E+03 | 4.148E+02 | 4.161E+02 | 5.200E+02 | 6.052E+02 | 7.003E+02 | 8.205E+02 | 9.366E+02 | 1.169E+03 | 1.550E+03 | 1.200E+03 | 1.300E+03 | 1.400E+03 | 1.503E+03 | 1.603E+03 | 3.046E+03 | 5.748E+03 | 1.903E+03 | 2.091E+03 | 3.491E+03 | 2.267E+03 | 2.500E+03 | 2.574E+03 | 2.700E+03 | 2.700E+03 | 2.855E+03 | **3.000E+03** | 3.555E+03 | 4.279E+03 |
| PEOA | 1.560E+05 | 1.221E+03 | 3.106E+02 | 4.291E+02 | 5.190E+02 | 6.037E+02 | 7.006E+02 | 8.213E+02 | 9.232E+02 | 1.625E+03 | 1.755E+03 | 1.201E+03 | 1.300E+03 | 1.400E+03 | 1.502E+03 | 1.603E+03 | 3.594E+03 | 8.713E+03 | 1.903E+03 | 2.110E+03 | 4.106E+03 | 2.249E+03 | 2.500E+03 | 2.530E+03 | 2.652E+03 | 2.700E+03 | 2.711E+03 | **3.000E+03** | 4.030E+03 | 5.184E+03 |
| HGS | 1.116E+05 | 8.349E+03 | 1.808E+03 | 4.137E+02 | 5.200E+02 | 6.044E+02 | 7.002E+02 | 8.001E+02 | 9.187E+02 | 1.052E+03 | 1.602E+03 | 1.200E+03 | 1.300E+03 | 1.400E+03 | 1.502E+03 | 1.603E+03 | 7.258E+03 | 1.569E+04 | 1.902E+03 | 6.909E+03 | 7.167E+03 | 2.250E+03 | 2.500E+03 | 2.564E+03 | 2.696E+03 | 2.700E+03 | 2.893E+03 | **3.000E+03** | 3.131E+03 | 3.316E+03 |
| FLA | 1.197E+02 | 2.004E+02 | 3.000E+02 | 4.258E+02 | 5.194E+02 | 6.055E+02 | 7.002E+02 | 8.226E+02 | 9.276E+02 | 1.483E+03 | 1.918E+03 | 1.200E+03 | 1.300E+03 | 1.400E+03 | 1.501E+03 | 1.603E+03 | 2.459E+03 | 2.415E+03 | 1.904E+03 | 2.095E+03 | 2.544E+03 | 2.296E+03 | 2.629E+03 | 2.553E+03 | 2.700E+03 | 2.700E+03 | 3.055E+03 | 3.191E+03 | 5.214E+03 | 4.048E+03 |
| BFO | 3.203E+04 | 3.250E+03 | 1.469E+03 | 4.247E+02 | 5.193E+02 | 6.034E+02 | 7.001E+02 | 8.213E+02 | 9.213E+02 | 1.713E+03 | 1.982E+03 | 1.200E+03 | 1.300E+03 | 1.400E+03 | 1.502E+03 | 1.603E+03 | 6.278E+03 | 6.279E+03 | 1.903E+03 | 2.937E+03 | 2.851E+03 | 2.259E+03 | 2.629E+03 | 2.535E+03 | 2.673E+03 | 2.700E+03 | 2.705E+03 | 3.274E+03 | 3.372E+03 | 4.229E+03 |
| CTCM | 3.574E+04 | 1.578E+03 | 2.530E+03 | 4.179E+02 | 5.201E+02 | 6.042E+02 | 7.002E+02 | 8.258E+02 | 9.235E+02 | 1.403E+03 | 1.712E+03 | 1.200E+03 | 1.300E+03 | 1.400E+03 | 1.502E+03 | 1.603E+03 | 4.440E+03 | 5.262E+03 | 1.903E+03 | 3.449E+03 | 3.109E+03 | 2.285E+03 | 2.625E+03 | 2.564E+03 | 2.691E+03 | 2.707E+03 | 3.031E+03 | 3.531E+03 | 1.332E+05 | 4.752E+03 |
| SSA | 3.745E+04 | 2.099E+03 | 3.115E+02 | 4.208E+02 | 5.200E+02 | 6.048E+02 | 7.002E+02 | 8.207E+02 | 9.414E+02 | 1.564E+03 | 1.952E+03 | 1.200E+03 | 1.300E+03 | 1.400E+03 | 1.502E+03 | 1.603E+03 | 4.973E+03 | 9.703E+03 | 1.903E+03 | 2.156E+03 | 2.472E+03 | 2.294E+03 | 2.500E+03 | 2.596E+03 | 2.700E+03 | 2.700E+03 | 2.900E+03 | **3.000E+03** | 3.161E+03 | 3.520E+03 |
| AVOA | 1.178E+05 | 3.212E+03 | 9.871E+02 | 4.240E+02 | 5.200E+02 | 6.057E+02 | 7.004E+02 | 8.087E+02 | 9.327E+02 | 1.118E+03 | 1.915E+03 | 1.200E+03 | 1.300E+03 | 1.400E+03 | 1.503E+03 | 1.603E+03 | 6.176E+03 | 1.023E+04 | 1.903E+03 | 4.806E+03 | 7.027E+03 | 2.244E+03 | 2.500E+03 | 2.582E+03 | 2.699E+03 | 2.700E+03 | 2.881E+03 | **3.000E+03** | 3.565E+03 | 4.217E+03 |
| ESOA | 1.230E+07 | 5.564E+08 | 5.195E+03 | 4.786E+02 | 5.200E+02 | 6.056E+02 | 7.216E+02 | 8.184E+02 | 9.187E+02 | 1.457E+03 | 1.621E+03 | 1.200E+03 | 1.300E+03 | 1.402E+03 | 1.506E+03 | 1.603E+03 | 1.771E+04 | 5.327E+03 | 1.905E+03 | 3.511E+03 | 7.746E+03 | 2.237E+03 | 2.523E+03 | 2.543E+03 | 2.692E+03 | 2.700E+03 | 2.813E+03 | 3.250E+03 | 3.292E+03 | 5.547E+03 |
| FATA | 1.215E+06 | 3.101E+04 | 3.127E+03 | 4.214E+02 | 5.204E+02 | 6.063E+02 | 7.004E+02 | 8.141E+02 | 9.338E+02 | 1.325E+03 | 1.921E+03 | 1.200E+03 | 1.300E+03 | 1.400E+03 | 1.503E+03 | 1.603E+03 | 4.963E+03 | 1.041E+04 | 1.903E+03 | 3.289E+03 | 6.672E+03 | 2.308E+03 | 2.500E+03 | 2.600E+03 | 2.700E+03 | 2.704E+03 | 2.893E+03 | **3.000E+03** | 3.544E+03 | 4.330E+03 |
| BKA | 5.520E+05 | 1.150E+08 | 1.582E+03 | 4.335E+02 | 5.202E+02 | 6.060E+02 | 7.081E+02 | 8.298E+02 | 9.303E+02 | 1.600E+03 | 1.863E+03 | 1.200E+03 | 1.300E+03 | 1.401E+03 | 1.506E+03 | 1.603E+03 | 3.222E+03 | 1.925E+03 | 1.904E+03 | 2.164E+03 | 2.656E+03 | 2.283E+03 | 2.500E+03 | 2.578E+03 | 2.694E+03 | 2.700E+03 | 2.855E+03 | **3.000E+03** | 3.441E+03 | 4.757E+03 |
| DBO | 1.098E+06 | 6.286E+03 | 1.153E+03 | 4.551E+02 | 5.203E+02 | 6.052E+02 | 7.003E+02 | 8.215E+02 | 9.329E+02 | 1.377E+03 | 1.961E+03 | 1.200E+03 | 1.300E+03 | 1.401E+03 | 1.502E+03 | 1.603E+03 | 1.041E+04 | 7.985E+03 | 1.904E+03 | 2.701E+03 | 8.988E+03 | 2.258E+03 | 2.617E+03 | 2.546E+03 | 2.676E+03 | 2.700E+03 | 2.839E+03 | 3.260E+03 | 6.127E+04 | 4.106E+03 |
| GPC | 3.836E+06 | 3.523E+08 | 1.610E+03 | 4.267E+02 | 5.200E+02 | 6.047E+02 | 7.063E+02 | 8.208E+02 | 9.325E+02 | 1.084E+03 | 1.588E+03 | 1.200E+03 | 1.300E+03 | 1.401E+03 | 1.503E+03 | 1.603E+03 | 2.990E+04 | 1.307E+04 | 1.902E+03 | 6.259E+03 | 1.622E+04 | 2.223E+03 | 2.500E+03 | 2.594E+03 | 2.700E+03 | 2.700E+03 | 2.900E+03 | 3.007E+03 | 3.468E+03 | 3.691E+03 |
| AO | 7.761E+06 | 2.915E+05 | 3.451E+03 | 4.365E+02 | 5.203E+02 | 6.044E+02 | 7.007E+02 | 8.236E+02 | 9.245E+02 | 1.475E+03 | 1.759E+03 | 1.201E+03 | 1.300E+03 | 1.400E+03 | 1.504E+03 | 1.603E+03 | 2.117E+04 | 1.339E+04 | 1.904E+03 | 6.171E+03 | 7.766E+03 | 2.269E+03 | 2.500E+03 | 2.600E+03 | 2.699E+03 | 2.700E+03 | 2.888E+03 | **3.000E+03** | 4.149E+03 | 4.129E+03 |

**Table 15**

Numerical results of ECO and competing metaheuristic algorithms on CEC-2017.

| MH | *F*1 | *F*3 | *F*4 | *F*5 | *F*6 | *F*7 | *F*8 | *F*9 | *F*10 | *F*11 | *F*12 | *F*13 | *F*14 | *F*15 | *F*16 | *F*17 | *F*18 | *F*19 | *F*20 | *F*21 | *F*22 | *F*23 | *F*24 | *F*25 | *F*26 | *F*27 | *F*28 | *F*29 | *F*30 |
|---|---|---|---|---|---|---|---|---|---|---|---|---|---|---|---|---|---|---|---|---|---|---|---|---|---|---|---|---|---|
| **ECO** | 1.535E+02 | **3.000E+02** | 4.003E+02 | 5.091E+02 | 6.000E+02 | 7.197E+02 | 8.123E+02 | **9.000E+02** | **1.283E+03** | **1.103E+03** | 3.432E+03 | 1.444E+03 | 1.428E+03 | 1.520E+03 | **1.601E+03** | **1.717E+03** | 1.900E+03 | 1.910E+03 | **2.002E+03** | **2.200E+03** | 2.296E+03 | 2.610E+03 | 2.649E+03 | **2.902E+03** | **2.867E+03** | 3.092E+03 | **3.090E+03** | 3.157E+03 | 3.827E+03 |
| ARO | 8.199E+02 | **3.000E+02** | 4.030E+02 | 5.159E+02 | 6.000E+02 | 7.312E+02 | 8.157E+02 | 9.040E+02 | 1.458E+03 | 1.109E+03 | 9.730E+03 | 1.575E+03 | 1.414E+03 | **1.505E+03** | 1.697E+03 | 1.721E+03 | 2.178E+03 | **1.901E+03** | 2.013E+03 | 2.271E+03 | **2.293E+03** | **2.607E+03** | 2.727E+03 | 2.920E+03 | 2.908E+03 | 3.100E+03 | 3.187E+03 | 3.183E+03 | 5.719E+03 |
| CFOA | 1.609E+03 | **3.000E+02** | 4.011E+02 | **5.087E+02** | 6.013E+02 | **7.168E+02** | **8.073E+02** | 9.003E+02 | 1.455E+03 | 1.120E+03 | 8.295E+03 | 1.987E+03 | 1.447E+03 | 1.622E+03 | 1.629E+03 | 1.740E+03 | 2.633E+03 | 1.937E+03 | 2.027E+03 | 2.261E+03 | 2.297E+03 | 2.609E+03 | 2.712E+03 | 2.919E+03 | 2.905E+03 | 3.093E+03 | 3.227E+03 | **3.156E+03** | 3.532E+04 |
| CSA | 5.366E+02 | **3.000E+02** | 4.048E+02 | 5.097E+02 | 6.014E+02 | 7.218E+02 | 8.087E+02 | 9.030E+02 | 1.529E+03 | 1.113E+03 | 5.601E+03 | 1.685E+03 | 1.437E+03 | 1.562E+03 | 1.634E+03 | 1.737E+03 | 2.020E+03 | 1.939E+03 | 2.027E+03 | 2.292E+03 | 2.297E+03 | 2.617E+03 | 2.735E+03 | 2.922E+03 | 2.891E+03 | 3.093E+03 | 3.273E+03 | 3.158E+03 | 1.010E+05 |
| WSO | 9.302E+02 | 3.031E+02 | 4.018E+02 | 5.157E+02 | 6.004E+02 | 7.202E+02 | 8.090E+02 | 9.093E+02 | 1.384E+03 | 1.125E+03 | 3.510E+03 | 1.418E+03 | 1.430E+03 | 1.519E+03 | 1.669E+03 | 1.728E+03 | 1.827E+03 | 1.905E+03 | 2.026E+03 | 2.279E+03 | 2.308E+03 | 2.630E+03 | 2.678E+03 | 2.919E+03 | 2.936E+03 | 3.115E+03 | 3.189E+03 | 3.171E+03 | 1.144E+05 |
| INFO | **1.000E+02** | **3.000E+02** | **4.000E+02** | 5.194E+02 | 6.007E+02 | 7.376E+02 | 8.209E+02 | 9.132E+02 | 1.712E+03 | 1.135E+03 | **1.646E+03** | 1.396E+03 | 1.444E+03 | 1.540E+03 | 1.725E+03 | 1.755E+03 | 1.842E+03 | 1.913E+03 | 2.031E+03 | 2.319E+03 | 2.302E+03 | 2.625E+03 | 2.744E+03 | 2.924E+03 | 3.047E+03 | 3.100E+03 | 3.299E+03 | 3.234E+03 | 2.919E+05 |
| RIME | 3.365E+03 | 3.000E+02 | 4.083E+02 | 5.111E+02 | 6.000E+02 | 7.198E+02 | 8.112E+02 | 9.000E+02 | 1.327E+03 | 1.107E+03 | 1.441E+04 | 1.195E+04 | 1.453E+03 | 1.751E+03 | 1.749E+03 | 1.756E+03 | 5.052E+03 | 1.991E+03 | 2.026E+03 | 2.305E+03 | 2.300E+03 | 2.616E+03 | 2.724E+03 | 2.927E+03 | 3.014E+03 | 3.096E+03 | 3.284E+03 | 3.174E+03 | 3.370E+05 |
| TJO | 1.409E+03 | 3.001E+02 | 4.046E+02 | 5.151E+02 | 6.020E+02 | 7.247E+02 | 8.140E+02 | 9.000E+02 | 1.763E+03 | 1.125E+03 | 1.039E+04 | 2.104E+03 | 1.452E+03 | 1.734E+03 | 1.737E+03 | 1.753E+03 | 2.272E+03 | 1.930E+03 | 2.081E+03 | 2.206E+03 | 2.302E+03 | 2.615E+03 | 2.582E+03 | 2.933E+03 | 2.869E+03 | 3.094E+03 | 3.107E+03 | 3.195E+03 | 9.864E+04 |
| RDCO | 9.436E+02 | **3.000E+02** | 4.004E+02 | 5.156E+02 | 6.003E+02 | 7.282E+02 | 8.173E+02 | **9.000E+02** | 2.063E+03 | 1.126E+03 | 1.712E+04 | 7.016E+03 | 1.488E+03 | 1.747E+03 | 1.612E+03 | 1.747E+03 | 5.801E+03 | 1.989E+03 | 2.042E+03 | 2.212E+03 | 2.301E+03 | 2.611E+03 | 2.734E+03 | 2.944E+03 | 2.900E+03 | 3.097E+03 | 3.100E+03 | 3.181E+03 | 1.232E+04 |
| SMA | 7.810E+03 | 3.000E+02 | 4.087E+02 | 5.133E+02 | 6.000E+02 | 7.221E+02 | 8.155E+02 | 9.000E+02 | 1.538E+03 | 1.111E+03 | 4.178E+04 | 1.179E+04 | 1.437E+03 | 1.530E+03 | 1.729E+03 | 1.736E+03 | 2.701E+04 | 2.637E+03 | 2.020E+03 | 2.298E+03 | 2.339E+03 | 2.618E+03 | 2.756E+03 | 2.927E+03 | 3.152E+03 | **3.090E+03** | 3.292E+03 | 3.179E+03 | 3.448E+04 |
| IAO | 1.909E+08 | 9.999E+02 | 4.155E+02 | 5.250E+02 | 6.118E+02 | 7.347E+02 | 8.183E+02 | 9.775E+02 | 1.715E+03 | 1.111E+03 | 2.000E+03 | **1.337E+03** | **1.412E+03** | 1.512E+03 | 1.649E+03 | 1.737E+03 | **1.825E+03** | 1.903E+03 | 2.033E+03 | 2.228E+03 | 2.314E+03 | 2.618E+03 | 2.641E+03 | 2.931E+03 | 3.006E+03 | 3.093E+03 | 3.216E+03 | 3.167E+03 | 2.600E+05 |
| SO | 2.152E+03 | **3.000E+02** | 4.030E+02 | 5.165E+02 | 6.004E+02 | 7.331E+02 | 8.167E+02 | 9.045E+02 | 1.637E+03 | 1.113E+03 | 1.617E+04 | 3.580E+03 | 1.471E+03 | 1.646E+03 | 1.701E+03 | 1.747E+03 | 5.096E+03 | 1.970E+03 | 2.038E+03 | 2.320E+03 | 2.302E+03 | 2.624E+03 | 2.752E+03 | 2.915E+03 | 3.231E+03 | 3.106E+03 | 3.279E+03 | 3.198E+03 | 1.105E+05 |
| SAO | 2.545E+03 | **3.000E+02** | 4.001E+02 | 5.148E+02 | 6.001E+02 | 7.207E+02 | 8.123E+02 | 9.004E+02 | 1.580E+03 | 1.107E+03 | 9.513E+03 | 1.172E+04 | 7.239E+03 | 4.389E+03 | 1.678E+03 | 1.743E+03 | 2.220E+04 | 1.322E+04 | 2.068E+03 | 2.304E+03 | 2.301E+03 | 2.617E+03 | 2.744E+03 | 2.930E+03 | 3.015E+03 | 3.095E+03 | 3.354E+03 | 3.197E+03 | 3.926E+05 |
| PKO | 4.375E+03 | 3.000E+02 | 4.052E+02 | 5.186E+02 | 6.000E+02 | 7.257E+02 | 8.148E+02 | **9.000E+02** | 1.884E+03 | 1.106E+03 | 5.301E+04 | 2.555E+03 | 1.504E+03 | 1.935E+03 | 1.663E+03 | 1.729E+03 | 1.327E+04 | 4.425E+03 | 2.032E+03 | 2.304E+03 | 2.301E+03 | 2.619E+03 | 2.749E+03 | 2.924E+03 | 2.979E+03 | 3.090E+03 | 3.188E+03 | 3.178E+03 | 1.385E+05 |
| HBA | 2.607E+03 | **3.000E+02** | 4.023E+02 | 5.183E+02 | 6.003E+02 | 7.343E+02 | 8.157E+02 | 9.014E+02 | 1.800E+03 | 1.132E+03 | 1.812E+04 | 2.845E+03 | 1.501E+03 | 1.565E+03 | 1.701E+03 | 1.749E+03 | 6.735E+03 | 2.044E+03 | 2.071E+03 | 2.294E+03 | 2.300E+03 | 2.636E+03 | 2.725E+03 | 2.928E+03 | 3.072E+03 | 3.120E+03 | 3.377E+03 | 3.244E+03 | 2.762E+06 |
| DMOA | 2.055E+03 | 3.006E+02 | 4.038E+02 | 5.178E+02 | **6.000E+02** | 7.325E+02 | 8.168E+02 | **9.000E+02** | 2.440E+03 | 1.103E+03 | 6.074E+04 | 5.043E+03 | 1.478E+03 | 2.038E+03 | 1.604E+03 | 1.737E+03 | 1.346E+04 | 2.134E+03 | 2.021E+03 | 2.284E+03 | 2.301E+03 | 2.628E+03 | 2.756E+03 | 2.913E+03 | 2.997E+03 | 3.139E+03 | 3.280E+03 | 3.220E+03 | **3.777E+03** |
| RUN | 3.883E+03 | 3.000E+02 | 4.004E+02 | 5.292E+02 | 6.137E+02 | 7.510E+02 | 8.259E+02 | 1.005E+03 | 1.577E+03 | 1.122E+03 | 9.527E+03 | 6.058E+03 | 1.505E+03 | 1.525E+03 | 1.759E+03 | 1.758E+03 | 1.714E+04 | 5.360E+03 | 2.074E+03 | 2.213E+03 | 2.304E+03 | 2.622E+03 | 2.737E+03 | 2.915E+03 | 2.999E+03 | 3.094E+03 | 3.329E+03 | 3.200E+03 | 5.130E+03 |
| PEOA | 2.170E+03 | 3.000E+02 | 4.098E+02 | 5.261E+02 | 6.120E+02 | 7.380E+02 | 8.183E+02 | 9.076E+02 | 1.744E+03 | 1.136E+03 | 1.192E+05 | 1.101E+04 | 1.471E+03 | 1.627E+03 | 1.690E+03 | 1.754E+03 | 1.297E+04 | 2.307E+03 | 2.055E+03 | 2.201E+03 | 2.307E+03 | 2.620E+03 | **2.542E+03** | 2.924E+03 | 2.908E+03 | 3.095E+03 | 3.182E+03 | 3.203E+03 | 4.592E+05 |
| HGS | 6.559E+03 | 3.000E+02 | 4.039E+02 | 5.208E+02 | 6.004E+02 | 7.323E+02 | 8.187E+02 | 9.000E+02 | 1.516E+03 | 1.120E+03 | 2.522E+04 | 9.923E+03 | 1.690E+03 | 3.145E+03 | 1.730E+03 | 1.745E+03 | 2.226E+04 | 7.430E+03 | 2.008E+03 | 2.309E+03 | 2.416E+03 | 2.626E+03 | 2.741E+03 | 2.942E+03 | 3.297E+03 | 3.104E+03 | 3.357E+03 | 3.208E+03 | 2.576E+05 |
| FLA | 3.060E+03 | **3.000E+02** | 4.000E+02 | 5.326E+02 | 6.076E+02 | 7.361E+02 | 8.224E+02 | 9.491E+02 | 1.989E+03 | 1.161E+03 | 2.030E+04 | 2.171E+03 | 1.498E+03 | 1.609E+03 | 1.770E+03 | 1.792E+03 | 2.037E+03 | 1.989E+03 | 2.114E+03 | 2.252E+03 | 2.300E+03 | 2.633E+03 | 2.751E+03 | 2.945E+03 | 3.119E+03 | 3.099E+03 | 3.282E+03 | 3.282E+03 | 7.575E+05 |
| BFO | 4.346E+03 | **3.000E+02** | 4.032E+02 | 5.240E+02 | 6.028E+02 | 7.402E+02 | 8.230E+02 | 1.059E+03 | 1.997E+03 | 1.161E+03 | 1.476E+04 | 8.022E+03 | 1.488E+03 | 1.767E+03 | 1.728E+03 | 1.756E+03 | 1.108E+04 | 2.595E+03 | 2.086E+03 | 2.202E+03 | 2.301E+03 | 2.620E+03 | 2.657E+03 | 2.943E+03 | 3.053E+03 | 3.104E+03 | 3.296E+03 | 3.220E+03 | 3.698E+05 |
| CTCM | 2.188E+03 | **3.000E+02** | 4.049E+02 | 5.382E+02 | 6.136E+02 | 7.368E+02 | 8.185E+02 | 1.070E+03 | 1.639E+03 | 1.132E+03 | 1.326E+04 | 5.351E+03 | 2.202E+03 | 2.474E+03 | 1.792E+03 | 1.757E+03 | 6.483E+03 | 5.072E+03 | 2.096E+03 | 2.301E+03 | 2.301E+03 | 2.694E+03 | 2.734E+03 | 2.910E+03 | 3.189E+03 | 3.152E+03 | 3.253E+03 | 3.236E+03 | 7.635E+04 |
| SSA | 4.028E+03 | **3.000E+02** | 4.011E+02 | 5.330E+02 | 6.045E+02 | 7.684E+02 | 8.296E+02 | 1.084E+03 | 1.832E+03 | 1.160E+03 | 1.450E+04 | 8.748E+03 | 1.488E+03 | 1.632E+03 | 1.754E+03 | 1.791E+03 | 7.723E+03 | 1.986E+03 | 2.064E+03 | 2.321E+03 | 2.340E+03 | 2.636E+03 | 2.757E+03 | 2.940E+03 | 3.319E+03 | 3.113E+03 | 3.393E+03 | 3.254E+03 | 3.741E+05 |
| AVOA | 3.631E+03 | **3.000E+02** | 4.065E+02 | 5.319E+02 | 6.061E+02 | 7.592E+02 | 8.252E+02 | 1.025E+03 | 1.707E+03 | 1.136E+03 | 1.033E+05 | 1.092E+04 | 1.523E+03 | 2.399E+03 | 1.761E+03 | 1.779E+03 | 1.158E+04 | 6.289E+03 | 2.072E+03 | 2.252E+03 | 2.307E+03 | 2.638E+03 | 2.741E+03 | 2.934E+03 | 3.018E+03 | 3.100E+03 | 3.315E+03 | 3.231E+03 | 4.375E+04 |
| ESOA | 1.002E+09 | 4.882E+03 | 4.760E+02 | 5.210E+02 | 6.174E+02 | 7.302E+02 | 8.159E+02 | 9.721E+02 | 1.736E+03 | 1.223E+03 | 9.164E+05 | 4.365E+03 | 1.806E+03 | 2.026E+03 | 1.709E+03 | 1.747E+03 | 2.816E+03 | 2.745E+03 | 2.067E+03 | 2.211E+03 | 2.380E+03 | 2.651E+03 | 2.633E+03 | 2.954E+03 | 3.107E+03 | 3.117E+03 | 3.358E+03 | 3.201E+03 | 1.912E+05 |
| FATA | 1.665E+05 | 4.609E+02 | 4.060E+02 | 5.334E+02 | 6.089E+02 | 7.509E+02 | 8.229E+02 | 9.147E+02 | 1.888E+03 | 1.133E+03 | 5.428E+05 | 8.072E+03 | 1.457E+03 | 2.192E+03 | 1.837E+03 | 1.768E+03 | 2.283E+04 | 2.252E+03 | 2.067E+03 | 2.269E+03 | 2.312E+03 | 2.629E+03 | 2.722E+03 | 2.927E+03 | 2.903E+03 | 3.108E+03 | 3.343E+03 | 3.212E+03 | 2.180E+05 |
| BKA | 1.566E+06 | 1.774E+03 | 4.155E+02 | 5.378E+02 | 6.267E+02 | 7.634E+02 | 8.225E+02 | 1.165E+03 | 1.905E+03 | 1.140E+03 | 1.683E+04 | 2.344E+03 | 1.473E+03 | 1.596E+03 | 1.734E+03 | 1.764E+03 | 4.603E+03 | 1.955E+03 | 2.082E+03 | 2.295E+03 | 2.362E+03 | 2.646E+03 | 2.747E+03 | 2.925E+03 | 3.152E+03 | 3.108E+03 | 3.237E+03 | 3.226E+03 | 8.674E+05 |
| DBO | 7.591E+03 | **3.000E+02** | 4.229E+02 | 5.342E+02 | 6.069E+02 | 7.508E+02 | 8.291E+02 | 9.328E+02 | 1.777E+03 | 1.160E+03 | 1.545E+06 | 1.260E+04 | 1.507E+03 | 1.847E+03 | 1.759E+03 | 1.758E+03 | 1.930E+04 | 2.122E+03 | 2.064E+03 | 2.209E+03 | 2.306E+03 | 2.638E+03 | 2.602E+03 | 2.926E+03 | 3.040E+03 | 3.100E+03 | 3.310E+03 | 3.239E+03 | 4.845E+05 |
| GPC | 2.072E+08 | 1.576E+03 | 4.485E+02 | 5.295E+02 | 6.152E+02 | 7.613E+02 | 8.222E+02 | 1.089E+03 | 1.549E+03 | 1.131E+03 | 9.204E+04 | 8.706E+03 | 3.480E+03 | 2.570E+03 | 1.763E+03 | 1.740E+03 | 2.250E+04 | 8.134E+03 | 2.053E+03 | 2.283E+03 | 2.357E+03 | 2.640E+03 | 2.762E+03 | 2.935E+03 | 3.056E+03 | 3.097E+03 | 3.244E+03 | 3.183E+03 | 3.751E+05 |
| AO | 3.652E+05 | 3.565E+02 | 4.064E+02 | 5.268E+02 | 6.136E+02 | 7.518E+02 | 8.220E+02 | 1.014E+03 | 1.883E+03 | 1.171E+03 | 2.977E+06 | 1.733E+04 | 1.846E+03 | 4.740E+03 | 1.753E+03 | 1.767E+03 | 2.681E+04 | 6.866E+03 | 2.108E+03 | 2.279E+03 | 2.305E+03 | 2.635E+03 | 2.762E+03 | 2.924E+03 | 3.015E+03 | 3.099E+03 | 3.407E+03 | 3.228E+03 | 6.036E+05 |

# Appendix B. Experimental results on CEC-2020

**Table 16**
Numerical results of ECO and competing metaheuristic algorithms on CEC-2020.

| MH | Dim | *F*1 | *F*2 | *F*3 | *F*4 | *F*5 | *F*6 | *F*7 | *F*8 | *F*9 | *F*10 |
|---|---|---|---|---|---|---|---|---|---|---|---|
| **ECO** | 10 | 1.69888E+02 | 1.42045E+03 | 7.19530E+02 | 1.90975E+03 | **1.94903E+03** | **1.60129E+03** | 2.13286E+03 | **2.29502E+03** | **2.61573E+03** | **2.89884E+03** |
| | 30 | 3.91671E+02 | **3.66733E+03** | **7.97826E+02** | 2.63556E+03 | 7.55108E+03 | **2.09889E+03** | 6.67077E+03 | **2.30000E+03** | 2.88873E+03 | **2.88612E+03** |
| | 50 | **4.94529E+02** | 6.65836E+03 | **9.39196E+02** | **1.08577E+04** | **3.36816E+04** | **2.77739E+03** | **2.08574E+04** | **5.16050E+03** | 3.06083E+03 | **3.04034E+03** |
| | 100 | **1.18207E+03** | **1.36097E+04** | **1.47342E+03** | **2.85145E+03** | **2.68771E+05** | **4.66695E+03** | **1.01203E+05** | **1.69576E+04** | 3.98784E+03 | **3.25929E+03** |
| ARO | 10 | 3.42848E+02 | 1.49398E+03 | 7.29484E+02 | **1.90113E+03** | 3.38028E+03 | 1.73383E+03 | **2.12820E+03** | 2.29942E+03 | 2.68156E+03 | 2.92546E+03 |
| | 30 | 3.94139E+03 | 3.93538E+03 | 9.29419E+02 | 6.21110E+03 | 1.52616E+05 | 2.71890E+03 | 7.50830E+04 | 2.40883E+03 | 2.94995E+03 | 2.90654E+03 |
| | 50 | 3.03010E+03 | 7.01877E+03 | 1.29413E+03 | 1.56479E+04 | 2.40186E+05 | 3.32166E+03 | 2.25225E+05 | 8.47836E+03 | 3.25668E+03 | 3.07747E+03 |
| | 100 | 7.02664E+03 | 1.52491E+04 | 2.60114E+03 | 4.64283E+03 | 1.28709E+06 | 5.47491E+03 | 8.05184E+05 | 1.79007E+04 | 4.50041E+03 | 3.31934E+03 |
| CFOA | 10 | 1.12471E+03 | 1.40630E+03 | **7.16901E+02** | 1.92728E+03 | 2.14414E+03 | 1.64285E+03 | 2.35023E+03 | 2.30345E+03 | 2.71921E+03 | 2.91564E+03 |
| | 30 | 2.75871E+03 | 4.05992E+03 | 8.43367E+02 | 1.34002E+04 | 8.75759E+03 | 2.28803E+03 | 8.71131E+03 | 2.30091E+03 | 2.91081E+03 | 2.93369E+03 |
| | 50 | 2.11318E+03 | 6.82116E+03 | 1.08323E+03 | 4.96019E+04 | 4.78035E+04 | 2.93099E+03 | 2.55479E+04 | 5.93024E+03 | 3.15252E+03 | 3.08426E+03 |
| | 100 | 4.94413E+03 | 1.44001E+04 | 1.97587E+03 | 5.67482E+05 | 6.36463E+05 | 5.61387E+03 | 2.12387E+05 | 1.73370E+04 | 4.16077E+03 | 3.31593E+03 |
| CSA | 10 | 4.62711E+02 | 1.67352E+03 | 7.21313E+02 | 1.93845E+03 | 2.03602E+03 | 1.62964E+03 | 2.24498E+03 | 2.30104E+03 | 2.73399E+03 | 2.91734E+03 |
| | 30 | 2.17745E+03 | 4.87468E+03 | 8.37848E+02 | 2.64744E+03 | 1.91104E+04 | 2.19487E+03 | 1.55280E+04 | 3.61051E+03 | **2.88251E+03** | 2.91991E+03 |
| | 50 | 2.46879E+03 | 8.74254E+03 | 1.03142E+03 | 1.10791E+04 | 6.83825E+04 | 2.91324E+03 | 5.41499E+04 | 9.28727E+03 | **3.04084E+03** | 3.08196E+03 |
| | 100 | 4.63556E+03 | 1.93136E+04 | 1.75849E+03 | 9.21852E+05 | 7.00463E+05 | 5.30729E+03 | 2.61920E+05 | 2.14430E+04 | **3.94856E+03** | 3.35186E+03 |
| WSO | 10 | 1.75987E+03 | **1.35203E+03** | 7.20344E+02 | 1.90305E+03 | 1.95305E+03 | 1.68288E+03 | 2.15446E+03 | 2.30934E+03 | 2.66714E+03 | 2.92416E+03 |
| | 30 | 4.08451E+03 | 3.81710E+03 | 1.03409E+03 | 5.85880E+03 | 1.22908E+05 | 2.44882E+03 | 1.06668E+05 | 2.65454E+03 | 3.36959E+03 | 2.90142E+03 |
| | 50 | 6.46388E+05 | **6.61789E+03** | 1.50047E+03 | 1.93992E+04 | 2.44446E+05 | 3.34518E+03 | 2.30928E+05 | 9.00932E+03 | 4.08150E+03 | 3.09319E+03 |
| | 100 | 1.27077E+09 | 1.36704E+04 | 2.96418E+03 | 6.75223E+03 | 8.60074E+05 | 6.07244E+03 | 4.74302E+05 | 1.93469E+04 | 5.12064E+03 | 3.56305E+03 |
| INFO | 10 | **1.00000E+02** | 1.74049E+03 | 7.33594E+02 | 1.91262E+03 | 2.11628E+03 | 1.73302E+03 | 2.38399E+03 | 2.30221E+03 | 2.75011E+03 | 2.92850E+03 |
| | 30 | **1.00005E+02** | 4.93076E+03 | 9.78514E+02 | **2.08993E+03** | **6.93494E+03** | 2.68436E+03 | **4.54806E+03** | 4.87569E+03 | 2.97208E+03 | 2.89892E+03 |
| | 50 | 5.21487E+03 | 8.55075E+03 | 1.38576E+03 | 1.44744E+04 | 4.36648E+04 | 3.55656E+03 | 2.12228E+04 | 9.80709E+03 | 3.29871E+03 | 3.05075E+03 |
| | 100 | 6.18795E+03 | 1.66468E+04 | 2.79274E+03 | 8.35616E+03 | 3.39367E+05 | 6.03451E+03 | 1.01659E+05 | 1.87630E+04 | 4.75243E+03 | 3.29834E+03 |

**Table 17**
Running time (in seconds) of ECO and competing metaheuristic algorithms on CEC-2020.

| MH | Dim | *F*1 | *F*2 | *F*3 | *F*4 | *F*5 | *F*6 | *F*7 | *F*8 | *F*9 | *F*10 |
|---|---|---|---|---|---|---|---|---|---|---|---|
| **ECO** | 10 | 0.431 | 0.775 | 0.717 | 1.624 | 0.705 | 0.643 | 0.617 | 0.786 | **0.594** | 0.803 |
| | 30 | 2.071 | 3.527 | 3.075 | 11.180 | 3.016 | 2.720 | 2.470 | 4.526 | **3.796** | 4.711 |
| | 50 | 5.679 | 8.247 | 7.185 | 29.653 | 6.872 | 6.193 | 5.532 | 12.491 | **10.482** | 14.124 |
| | 100 | 22.728 | 32.687 | 28.880 | 117.047 | 25.412 | 22.898 | 21.947 | **55.634** | 79.282 | 84.743 |
| ARO | 10 | 0.435 | 0.789 | 0.909 | 1.851 | 0.623 | 0.639 | 0.643 | 0.861 | 0.901 | 0.804 |
| | 30 | 1.730 | 3.111 | 3.733 | 12.007 | 2.610 | 2.590 | 2.555 | 4.575 | 5.255 | 4.317 |
| | 50 | 3.613 | 8.451 | 8.401 | 31.243 | 5.912 | 5.794 | 5.812 | **12.100** | 14.488 | **12.427** |
| | 100 | 19.373 | 35.222 | 32.468 | **96.259** | 23.404 | 22.508 | 22.279 | 57.797 | **73.282** | 83.313 |
| CFOA | 10 | 0.586 | 1.175 | 1.109 | 2.057 | 1.003 | 0.987 | 0.982 | 1.254 | 1.263 | 1.158 |
| | 30 | 2.899 | 4.671 | 4.242 | 12.429 | 3.792 | 3.820 | 3.655 | 6.103 | 6.747 | 5.918 |
| | 50 | 5.735 | 10.117 | 9.104 | 30.664 | 8.185 | 8.184 | 7.690 | 15.667 | 18.343 | 16.622 |
| | 100 | 24.777 | 35.535 | 33.079 | 115.659 | 29.823 | 29.508 | 28.033 | 72.283 | 93.358 | 82.537 |
| CSA | 10 | **0.232** | **0.507** | **0.450** | **1.321** | 0.424 | **0.409** | **0.418** | **0.657** | 0.710 | **0.614** |
| | 30 | 1.306 | **2.588** | **2.141** | **9.819** | **1.972** | **1.929** | **1.899** | **4.264** | 5.071 | **4.092** |
| | 50 | **3.175** | **6.584** | **5.300** | **27.142** | **4.899** | **4.825** | **4.598** | 12.422 | 15.221 | 13.042 |
| | 100 | **17.346** | **28.317** | **24.552** | 105.974 | **22.375** | **22.053** | **21.250** | 64.183 | 81.915 | **81.508** |
| WSO | 10 | 0.243 | 0.520 | 0.488 | 1.386 | **0.421** | 0.424 | 0.437 | 0.670 | 0.721 | 0.671 |
| | 30 | **1.248** | 2.705 | 2.445 | 10.523 | 2.259 | 2.150 | 2.155 | 4.461 | 5.338 | 4.651 |
| | 50 | 4.248 | 6.892 | 6.340 | 28.455 | 5.862 | 5.517 | 5.485 | 12.970 | 16.330 | 15.071 |
| | 100 | 21.651 | 30.286 | 27.953 | 110.282 | 26.224 | 24.991 | 24.524 | 66.942 | 89.569 | 84.302 |
| INFO | 10 | 1.526 | 1.711 | 1.627 | 2.367 | 1.501 | 1.471 | 1.486 | 1.877 | 1.622 | 1.490 |
| | 30 | 5.265 | 7.045 | 6.089 | 13.684 | 5.948 | 5.878 | 5.802 | 8.748 | 8.500 | 7.304 |
| | 50 | 12.937 | 15.180 | 13.140 | 34.551 | 12.913 | 12.579 | 12.400 | 20.202 | 22.431 | 19.736 |
| | 100 | 42.410 | 52.772 | 45.719 | 130.895 | 44.132 | 44.035 | 45.080 | 83.379 | 95.709 | 89.215 |

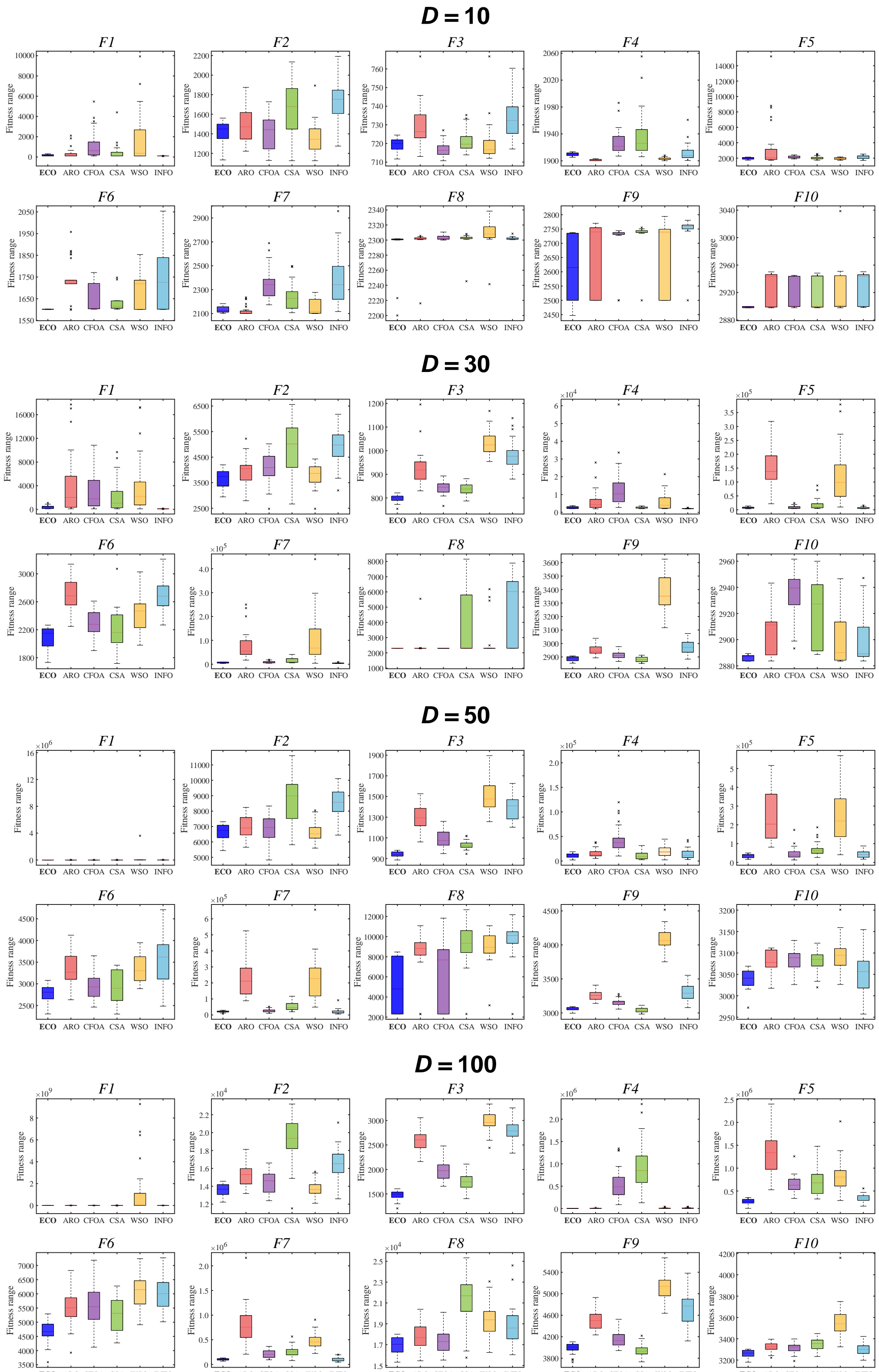

**Figure 12:** Range of numerical results of ECO and competing metaheuristic algorithms on CEC-2020.

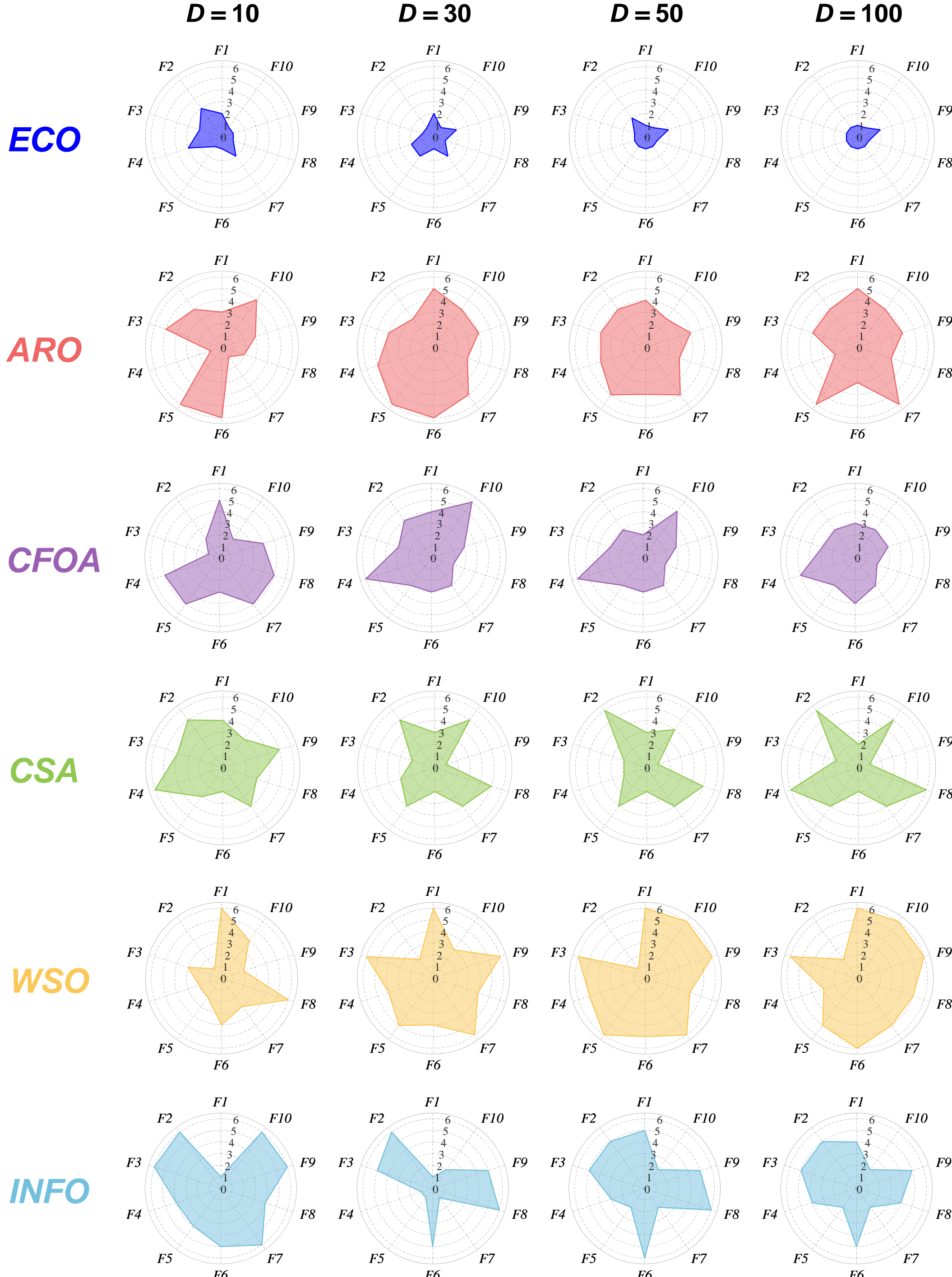

**Figure 13:** Rank of ECO and competing metaheuristic algorithms on CEC-2020.

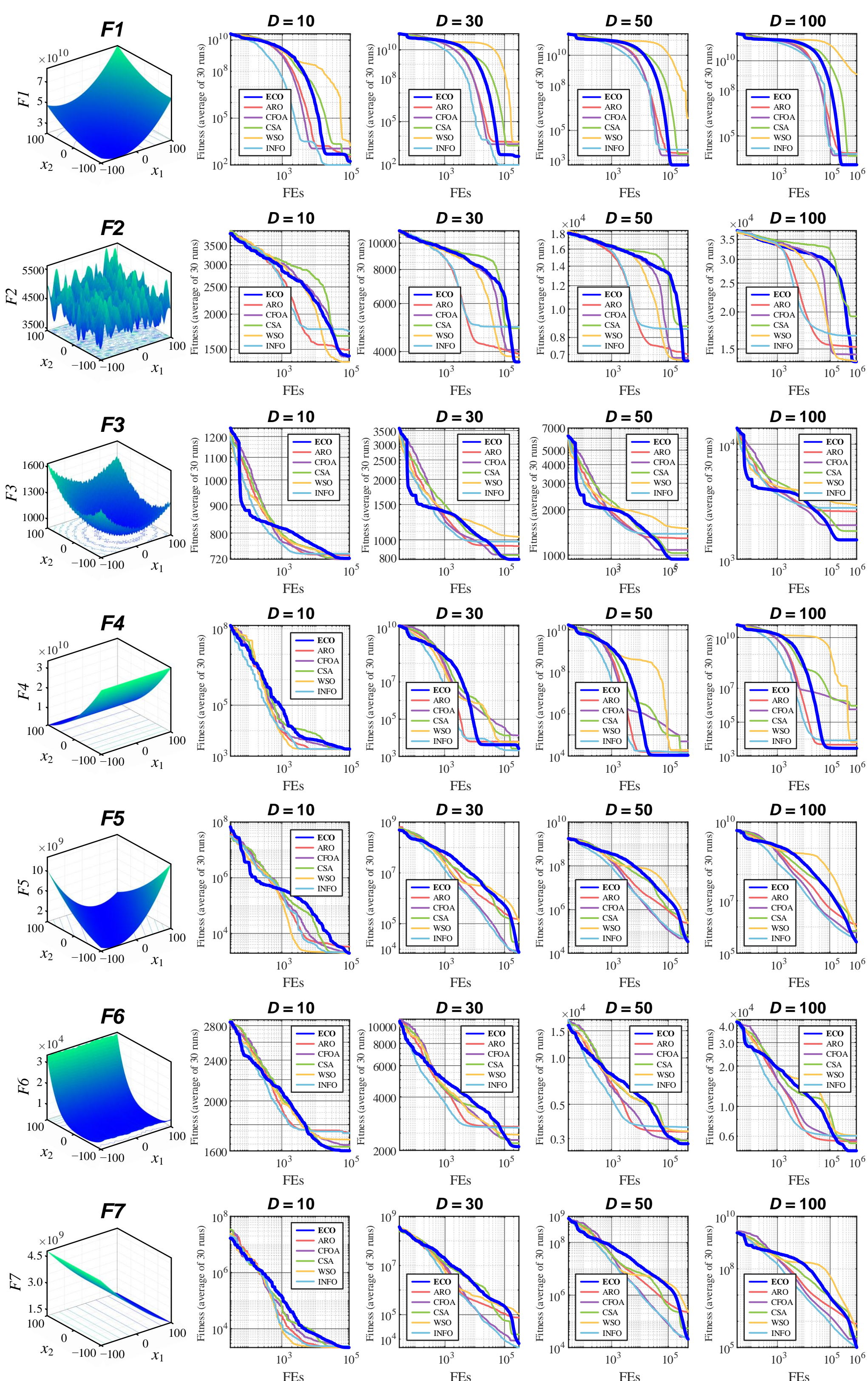

**Figure 14:** Mean convergence curve of ECO and competing metaheuristic algorithms on CEC-2020.

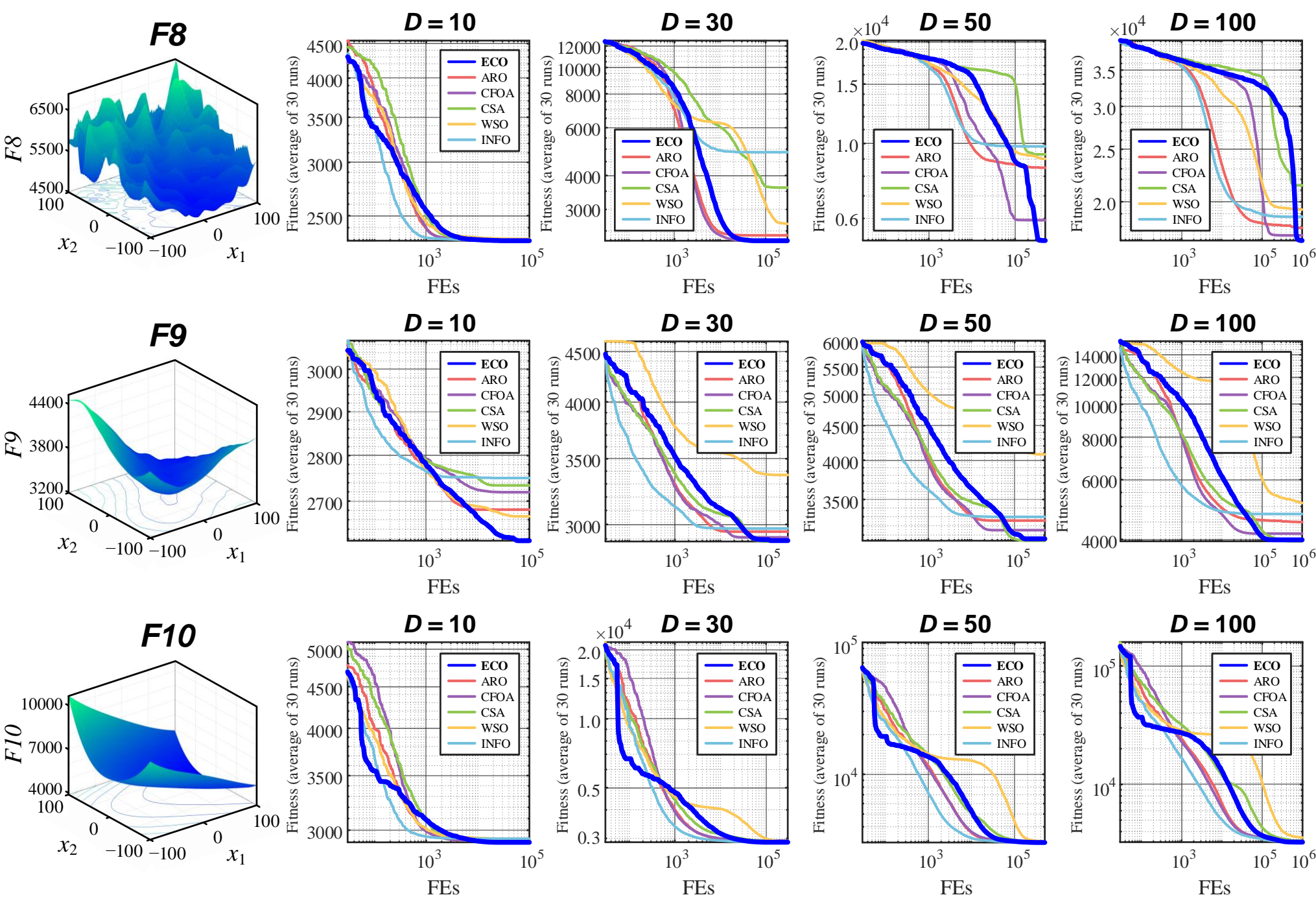

**Figure 14a:** Mean convergence curve of ECO and competing metaheuristic algorithms on CEC-2020 (continued).

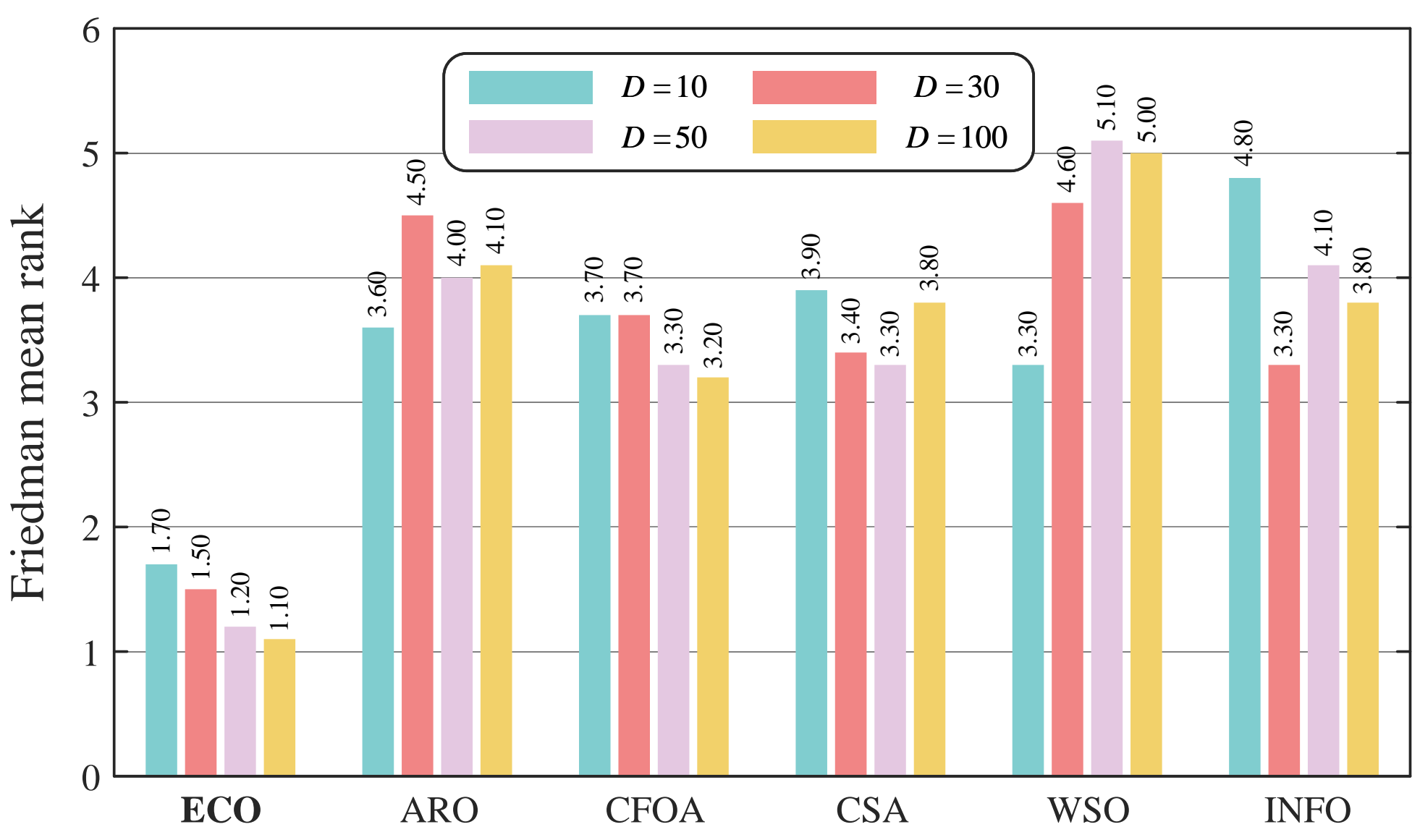

**Figure 15:** Scalability of ECO and competing metaheuristic algorithms on CEC-2020 ($D = 10/30/50/100$).

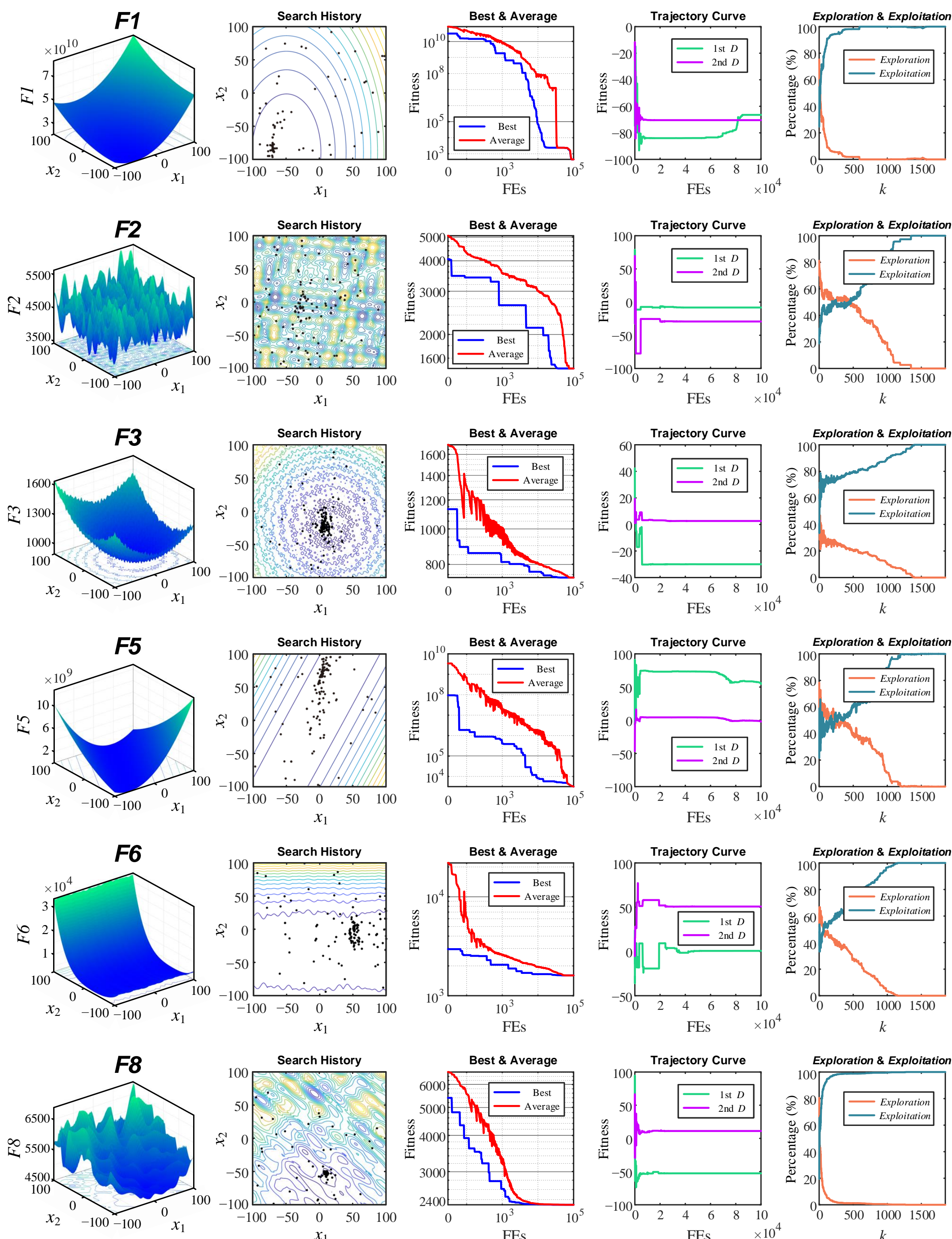

**Figure 16:** Qualitative analysis of ECO on 6 typical optimization functions of CEC-2020.